\documentclass{article} %
\usepackage{iclr2021_conference,times}

\usepackage{amsmath,amsfonts,bm}

\def\eqref#1{equation~\ref{#1}}

\def\1{\bm{1}}

\DeclareMathAlphabet{\mathsfit}{\encodingdefault}{\sfdefault}{m}{sl}
\SetMathAlphabet{\mathsfit}{bold}{\encodingdefault}{\sfdefault}{bx}{n}

\usepackage[pagebackref=false,breaklinks=true,colorlinks,bookmarks=false]{hyperref}
\hypersetup{colorlinks=true, linkcolor=blue, anchorcolor=blue, citecolor=blue, filecolor=blue, urlcolor=blue}

\usepackage{booktabs} 
\usepackage{url}
\usepackage{graphicx}
\usepackage{amsmath}
\usepackage{kotex} %
\usepackage{amssymb}
\usepackage{amsthm}
\usepackage{xcolor}   
\usepackage{adjustbox}
\usepackage{comment}
\usepackage[ruled]{algorithm2e}
\usepackage{bm}
\usepackage{mathtools}
\usepackage{multirow}
\usepackage{wrapfig}
\usepackage{kotex}
\usepackage{subcaption}
\usepackage{rotating}
\title{Learning Features with Parameter-Free \\ Layers}

\author{Dongyoon Han$^{1}$, YoungJoon Yoo$^{1,2}$, Beomyoung Kim$^{2}$, Byeongho Heo$^{1}$ \\ 
$^{1}$NAVER AI Lab, $^{2}$NAVER CLOVA
}

\iclrfinalcopy %

\newcommand\red[1]{{\textcolor{black}{#1}}}

\def\be#1\ee{\begin{align}#1\end{align}}
\def\bea#1\eea{\begin{eqnarray}#1\end{eqnarray}}
\def\ba#1\ea{\begin{align*}#1\end{align*}}
\def\bs#1\es{\begin{equation}\begin{split}#1\end{split}\end{equation}}

\definecolor{darkergreen}{RGB}{21, 152, 56}
\newcommand\greenp[1]{\textcolor{darkergreen}{(#1)}}
\definecolor{red2}{RGB}{252, 54, 65}
\newcommand\redp[1]{\textcolor{red2}{(#1)}}
\newcommand\greenpscript[1]{\scriptsize\greenp{#1}}
\newcommand\redpscript[1]{\scriptsize\redp{#1}}

\begin{document}

\maketitle

\begin{abstract}
Trainable layers such as convolutional building blocks are the standard network design choices by learning parameters to capture the global context through successive spatial operations. When designing an efficient network, trainable layers such as the depthwise convolution is the source of efficiency in the number of parameters and FLOPs, but there was little improvement to the model speed in practice. This paper argues that simple built-in parameter-free operations can be a favorable alternative to the efficient trainable layers replacing spatial operations in a network architecture. We aim to break the stereotype of organizing the spatial operations of building blocks into trainable layers. 
Extensive experimental analyses based on layer-level studies with fully-trained models and neural architecture searches are provided to investigate whether parameter-free operations such as the max-pool are functional. The studies eventually give us a simple yet effective idea for redesigning network architectures, where the parameter-free operations are heavily used as the main building block without sacrificing the model accuracy as much. Experimental results on the ImageNet dataset demonstrate that the network architectures with parameter-free operations could enjoy the advantages of further efficiency in terms of model speed, the number of the parameters, and FLOPs. Code and ImageNet pretrained models are available at \url{https://github.com/naver-ai/PfLayer}.
\end{abstract}

\section{Introduction}
\vspace{-1.5mm}
Image classification has been advanced with deep convolutional neural networks~\citep{vgg,densenet,resnet} with the common design paradigm of the network building blocks with trainable spatial convolutions inside. Such trainable layers with learnable parameters effectively grasp attentive signals to distinguish input but are computationally heavy. Rather than applying pruning or distillation techniques to reduce the computational cost, developing new efficient operations has been another underlying strategy. For example, a variant of the regular convolution, such as depthwise convolution~\citep{mobilenetv1} has been proposed to bring more efficiency by reducing the inter-channel computations. The operation has benefits in the computational budgets, including the number of parameters and FLOPs. However, the networks heavily using the depthwise convolution~\citep{mobilenetv1,mobilenetv2,efficientnet} have an inherent downside of the latency, which generally do not reach the speed of the regular convolution.

In a line of study of efficient operations, there have been many works~\citep{wang2018versatile,wu2018shift, han2020ghostnet,tan2021efficientnetv2} based on the regular convolution and the depthwise convolution. Most methods utilize the depthwise convolution's efficiency or target FLOPs-efficiency but are slow in the computation. Meanwhile, parameter-free operations were proposed; a representative work is the Shift operation~\citep{wu2018shift}. Its efficiency stems from the novel operation without learning spatial parameters by letting the feature mixing convolutions learn from the shifted features. However, the implementation does not bring about the actual speed-up as expected. This is because the operation-level optimization is still demanding compared to the regular convolution with highly optimized performance. Another parameter-free operation feature shuffling~\citep{shufflenetv1} is a seminal operation that reduces the computational cost of the feature mixing layer. However, it hardly plays the role of a spatial operation.

In this paper, we focus on efficient parameter-free operations that actually replace trainable layers for network design. We revisit the popular parameter-free operations, the max-pool and the avg-pool operations ($i.e.,$ layers), which are used in many deep neural networks~\citep{vgg, densenet} restricted to perform downsampling~\citep{vgg,mobilenetv1,mobilenetv2,resnet}. 
Can those simple parameter-free operations be used as the main network building block? If so, one can reduce a large portion of the parameters and the overall computational budget required during training and inference. To answer the question, the max-pool and the avg-pool operations are chosen to be demonstrated as representative simple parameter-free operations. %
We first conduct comprehensive studies on the layer replacements of the regular convolutions inside networks searched upon the baseline models with model training. Additionally, we incorporate a neural architecture search~\citep{darts} to explore effective architectures based on the operation list with the parameter-free operations and convolutional operations. Based on the investigations, we provide a simple rule of thumb to design an efficient architecture using the parameter-free operations upon primitive network architectures. The design guide is applied to popular heavy networks and validated by the performance trained on ImageNet~\citep{ImageNet}. It turns out that our models have apparent benefits to the computational costs, particularly the faster model speeds. In addition, ImageNet-C~\citep{hendrycks2019benchmarking} and ImageNet-O~\citep{hendrycks2021natural} results show our models are less prone to be overfitting. We further propose a novel deformable parameter-free operation based on the max-pool and the avg-pool to demonstrate the future of a parameter-free operation. Finally, we show that the parameter-free operations can successfully replace the self-attention layer~\citep{transformer}, thus attaining further efficiency and speed up. We summarize our contributions as follows:
\vspace{-2.5mm}
\begin{itemize}
    \item We study whether parameter-free operations can replace trainable layers as a network building block. To our knowledge, this is the first attempt to investigate a simple, built-in parameter-free layer as a building block for further efficiency (\S\ref{sec:study}).
    \item We provide a rule of thumb for designing a deep neural network including convolutional neural networks and vision transformers with parameter-free layers (\S\ref{sec:design}).
    \item Experimental results show that our efficient models outperform the previous efficient models and yield faster model speeds with further robustness  (\S\ref{sec:exp}).
\end{itemize}

\section{Preliminaries}
\label{sec:preliminary}
\vspace{-2mm}
A network building block where trainable layers are inside is a fundamental element of modularized networks~\citep{resnext,mobilenetv2,efficientnet}. We start a discussion with the elementary building block and then explore more efficient ones.

\vspace{-2mm}
\subsection{Basic Building Blocks}

\vspace{-2mm}
\paragraph{Convolution Layer.}
We recall the regular convolutional operation by formulating with matrix multiplication first. Let $f\in\mathcal{R}^{c_{in}\times H\times W}$ as the input feature,  %
the regular convolution of the feature $f$ with kernel size $k$ and stride $r$ is given as
\begin{equation}
y_{o, i,j} = \sigma\left(\sum^{\lfloor k/2 \rfloor}_{h, w=-\lfloor k/2 \rfloor}\sum^{c_{in}}_{u=1}{W_{o, u, h,w}}\cdot f_{u, r*i+h, r*j+w}\right),
\label{eq:convolution}
\end{equation}
where $W$ denotes the weight matrix, and the function $\sigma(\cdot)$ denotes an activation function such as ReLU~\citep{relu} with or without batch normalization (BN)~\citep{BN}. The convolution itself has been a building block due to the expressiveness and design flexibility. VGG~\citep{vgg} network was designed by  accumulating the 3${\times}$3 convolution to substitute the convolutions with larger kernel sizes but still have high computational costs.

\vspace{-2mm}
\paragraph{Bottleneck Block.}
We now recall the bottleneck block, which primarily aimed at efficiency. We represent the bottleneck block by the matrix multiplication as \eqref{eq:convolution}:
\begin{equation}
y_{o, i,j} = \sigma\left(\sum^{\rho c_{in}}_{v=1}P_{o, v} \cdot \sigma\left(\sum^{\lfloor k/2  \rfloor}_{h, w=-\lfloor k/2 \rfloor}\sum^{\rho c_{in}}_{u=1}{W_{v, u, h,w}}\cdot g_{v, r*i+h, r*j+w}\right)\right),
\label{eq:bottleneck}
\end{equation}
where $g_{o,i,j} = \sigma(\sum^{c_{in}}_{u=1} Q_{o, u} \cdot f_{u, i, j})$, and the matrix $P$ and $Q$ denote the weights of 1${\times}$1 convolutions with the inner dimensions $\rho c_{in}$. 
This design regime is efficient in terms of the computational budgets and even proven to be effective in the generalization ability when stacking up the bottleneck blocks compared with the basic blocks. 
Albeit a 3${\times}$3 convolution is replaced with two feature mixing layers ($i.e.,$ 1${\times}$1 convolutions), the expressiveness is still high enough with a low channel expansion ratio $\rho$ ($i.e.,$ $\rho=1/4$ in ResNet~\citep{resnet,resnext}). However, due to the presence of the regular 3${\times}$3 convolution, only adjusting $\rho$  hardly achieves further efficiency. 

\vspace{-2mm}
\subsection{Efficient Building Blocks}
\vspace{-2mm}
\paragraph{Inverted Bottleneck.}
The grouped operations, including the group convolution~\citep{resnext} and the depthwise convolution~\citep{mobilenetv1}) have emerged as more efficient building blocks. Using the depthwise convolution inside a bottleneck~\citep{mobilenetv2} is called the inverted bottleneck, which is represented as
\begin{equation}
y_{o, i,j} = \sigma\left(\sum^{\rho c_{in}}_{v=1}P_{o, v} \cdot \sigma\left(\sum^{\lfloor k/2 \rfloor}_{h, w=-\lfloor k/2 \rfloor}{W_{v, h,w}}\cdot g_{v, r*i+h, r*j+w}\right)\right),
\label{eq:invbottleneck}
\end{equation}
where the summation over the channels in \eqref{eq:bottleneck} has vanished. This slight modification takes advantage of further generalization ability, and therefore stacking of the blocks leads to outperform ResNet with a better trade-off between accuracy and efficiency as proven in many architectures~\citep{efficientnet,han2021rethinking}. The feature mixing operation in \eqref{eq:bottleneck} with the following point-wise activation function ($i.e.,$ ReLU) may offer a sufficient rank of the feature with the efficient operation. However, the inverted bottleneck and the variants below usually need a large expansion ratio $\rho>1$ to secure the expressiveness~\citep{mobilenetv2,mobilenetv3,efficientnet}, so the actual speed is hampered by the grouped operation that requires more optimization on GPU~\citep{gibson2020optimizing,lu2021optimizing}.
\vspace{-2mm}

\paragraph{Variants of Inverted Bottleneck.}
More refinements on the bottleneck block could bring further efficiency; the prior works~\citep{wang2018versatile,han2020ghostnet,tan2021efficientnetv2,wu2018shift} fully or partially redesigned the layers in the bottleneck with new operations, and therefore theoretical computational costs have been decreased. VersatileNet~\citep{wang2018versatile} replaced the convolutions with the proposed filters consisting of multiple convolutional filters; GhostNet~\citep{han2020ghostnet} similarly replaced the layers with the proposed module that concatenates a preceding regular convolution and additional depthwise convolution. Rather than involving new operations, a simple replacement by simplifying the operations has been proposed instead. EfficientNetV2~\citep{tan2021efficientnetv2} improved the training speed by fusing the pointwise and the depthwise convolutions to a single regular convolution. This takes advantage of using the highly-optimized operation rather than using new unoptimized operations. %
ShiftNet~\citep{wu2018shift} simplify the depthwise convolution to the shift operation, where the formulation is very similar to \eqref{eq:invbottleneck}: 
\begin{equation}
y_{o, i,j} = \sigma\left(\sum^{\rho c_{in}}_{v=1}P_{o, v} \cdot \sigma\left(\sum^{\lfloor k/2 \rfloor}_{h, w=-\lfloor k/2 \rfloor}{W_{v, h,w}}\cdot g_{v, r*i+h, r*j+w}\right)\right),
\label{eq:shift}
\end{equation}
where $W_{v,:,:}$ is simplified to have one 1 and 0 for the rest all $h,w$ for each $v$. This change involves a new operation, so-called Shift, that looks like shifting the features after the computation, and the preceding and the following pointwise convolutions mix the shifted output. The shift operation is actually classified as a parameter-free operation, but a large expansion ratio is yet needed to hold the expressiveness. Furthermore, the implementation is not as readily optimized as regular operations, even in CUDA implementation~\citep{he2019addressnet,chen2019all}.

\section{Efficient Building Block with Parameter-free Operations}
\label{sec:study}
\vspace{-2mm}
In this section, we extend an efficient building block by incorporating parameter-free operations. We empirically study whether the parameter-free operations can be used as the main building block in deep neural networks like regular convolution.
\begin{figure*}[t]
\small
\centering
\begin{subfigure}[ht!]{0.14\linewidth}
\quad
\end{subfigure} 
\begin{subfigure}[ht!]{0.27\linewidth}
\quad\quad\quad {\fontsize{8.0}{9.0}\selectfont SGD}
\end{subfigure} \ 
\begin{subfigure}[ht!]{0.27\linewidth}
\quad\quad \quad {\fontsize{8.0}{9.0}\selectfont AdamW}
\end{subfigure} \ 
\begin{subfigure}[ht!]{0.27\linewidth}
\quad\quad\quad {\fontsize{8.0}{9.0}\selectfont AdamP}
\end{subfigure} 
\\
\begin{subfigure}[ht!]{0.055\linewidth}
\rotatebox{90}{\parbox{2cm}{{\fontsize{8.0}{9.0}\selectfont Channel \\ Widths: 32}}} \vspace{9mm}
\end{subfigure} 
\begin{subfigure}[ht!]{0.27\linewidth}
\includegraphics[page=1, trim = 0mm 0mm 0mm 0mm, clip, width=1.0\linewidth]{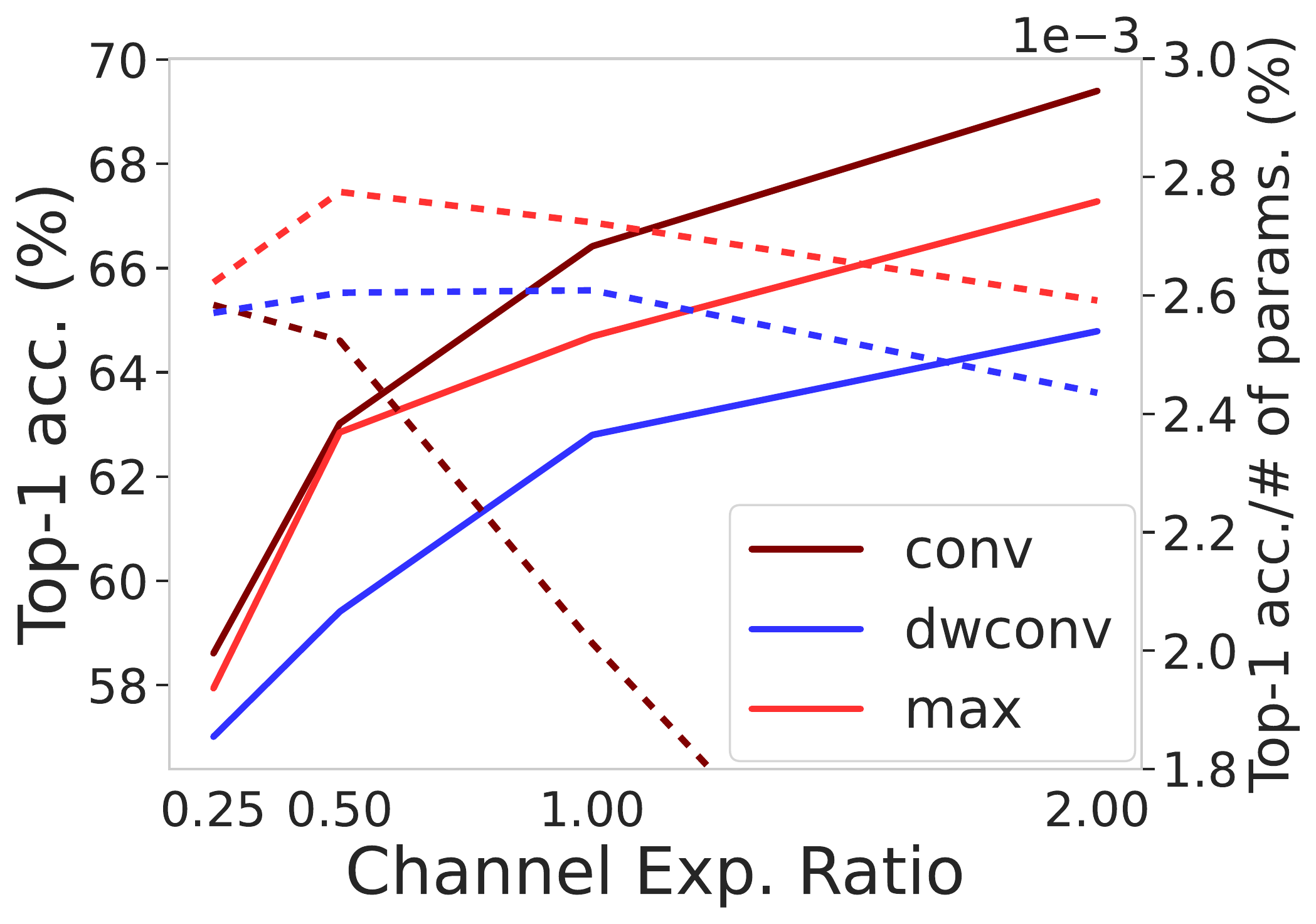}
\end{subfigure} \ \
\begin{subfigure}[ht!]{0.27\linewidth}
\includegraphics[page=1, trim = 0mm 0mm 0mm 0mm, clip, width=1.0\linewidth]{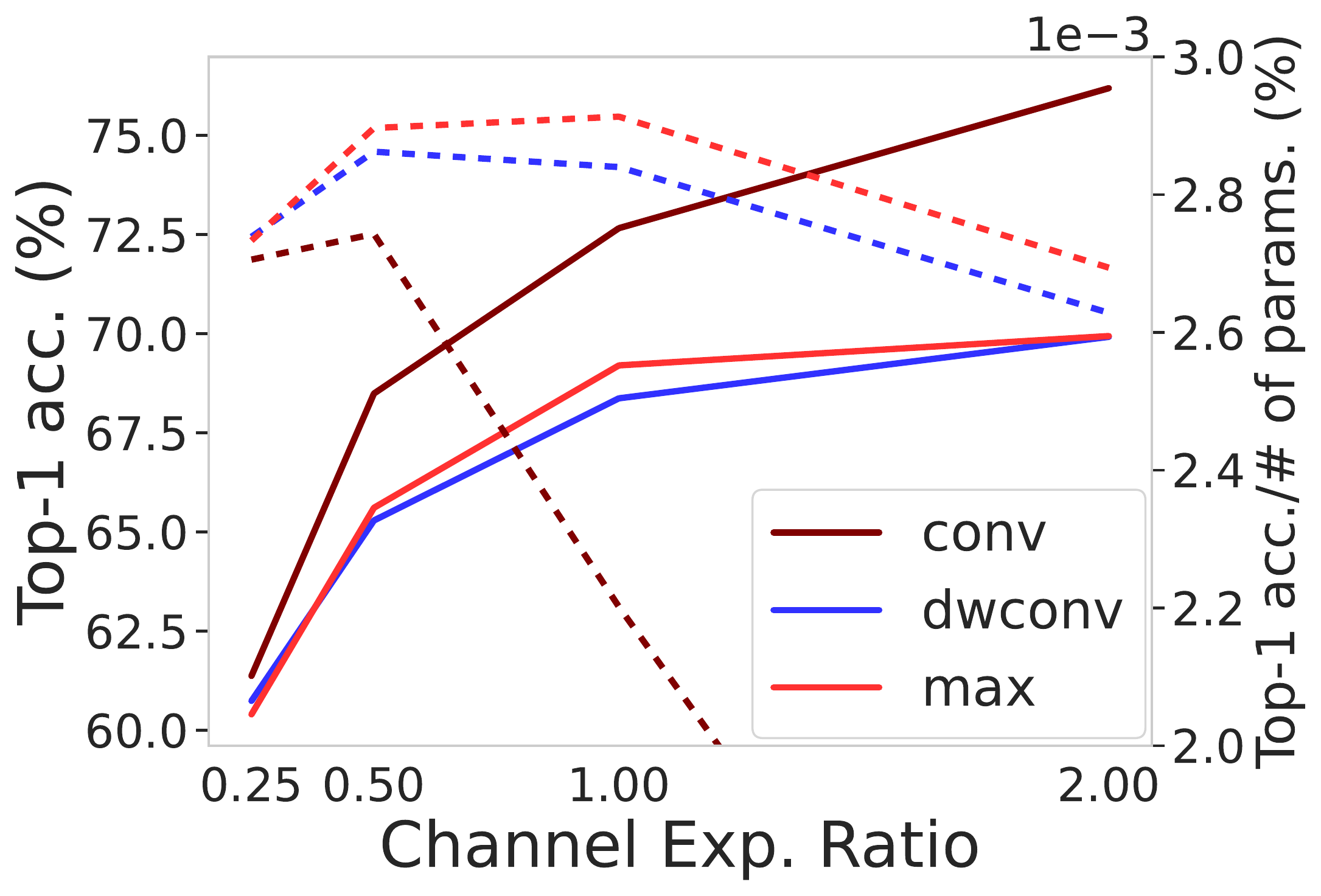}  
\end{subfigure} \ \
\begin{subfigure}[ht!]{0.27\linewidth}
\includegraphics[page=1, trim = 0mm 0mm 0mm 0mm, clip, width=1.0\linewidth]{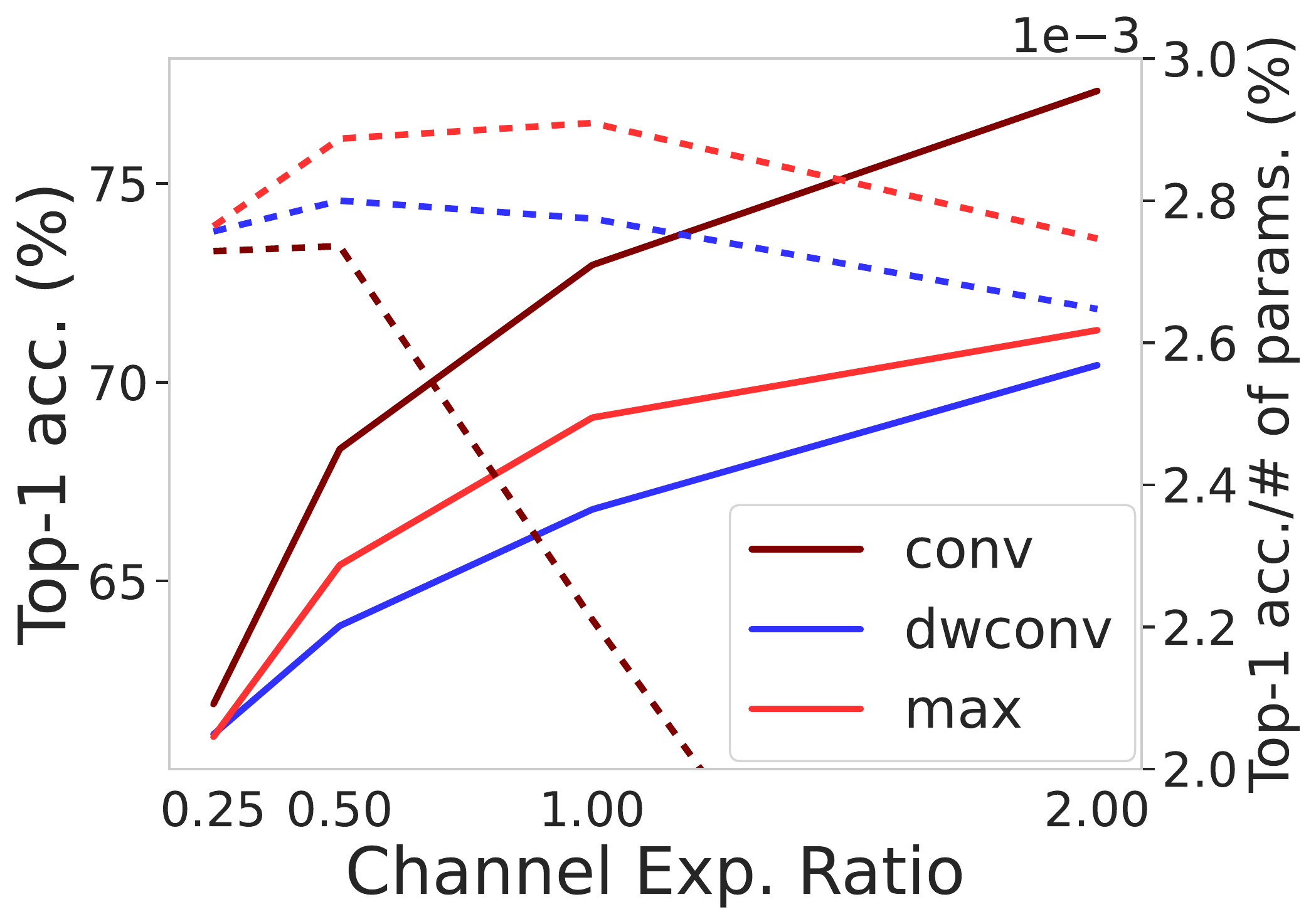}  
\end{subfigure} 
\\
\vspace{-2mm}
\begin{subfigure}[ht!]{0.055\linewidth}
\rotatebox{90}{\parbox{2cm}{{\fontsize{8.0}{9.0}\selectfont Channel \\ Widths: 64}}} \vspace{9mm}
\end{subfigure} 
\begin{subfigure}[ht!]{0.27\linewidth}
\includegraphics[page=1, trim = 0mm 0mm 0mm 0mm, clip, width=1.0\linewidth]{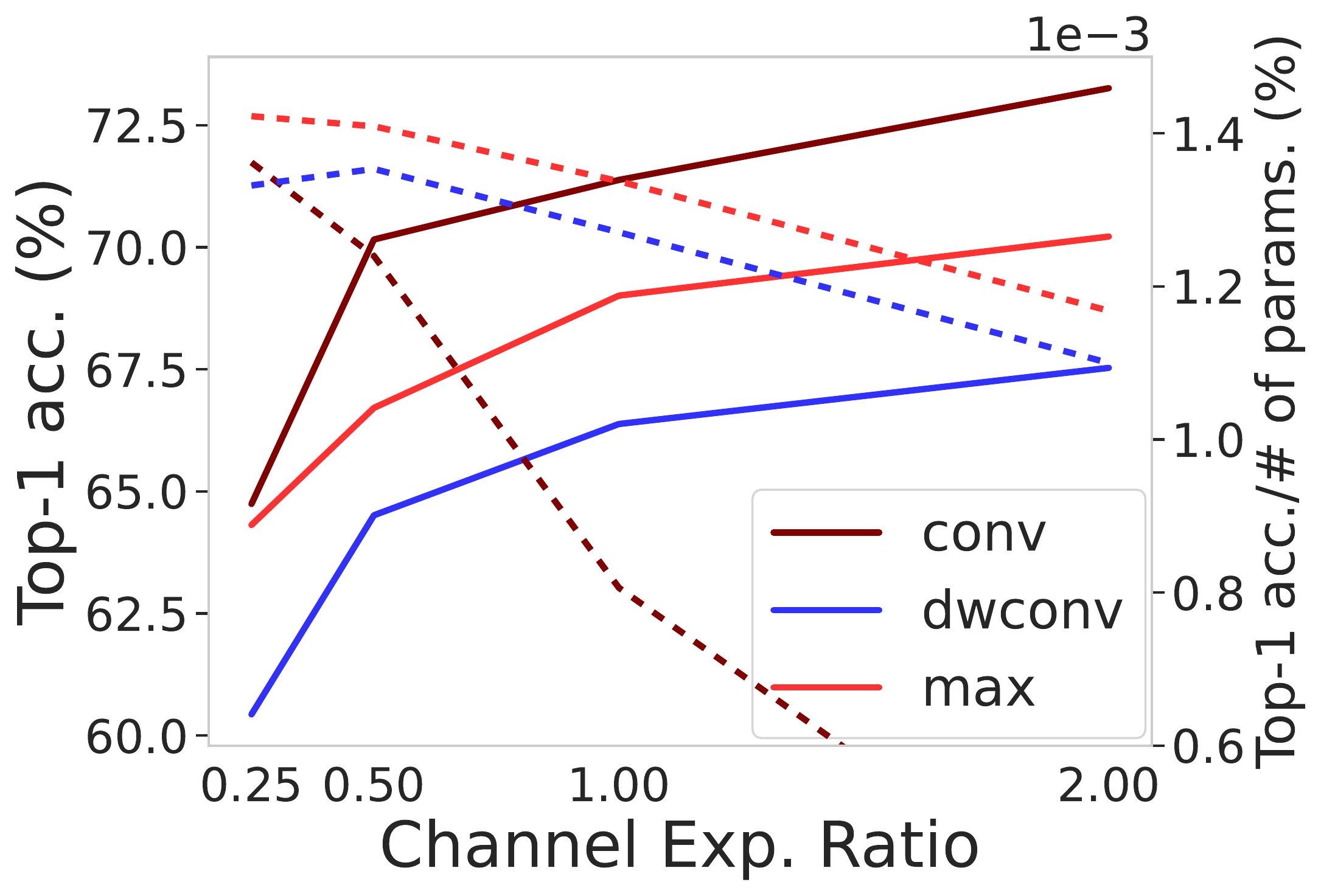}
\end{subfigure} \ \
\begin{subfigure}[ht!]{0.27\linewidth}
\includegraphics[page=1, trim = 0mm 0mm 0mm 0mm, clip, width=1.0\linewidth]{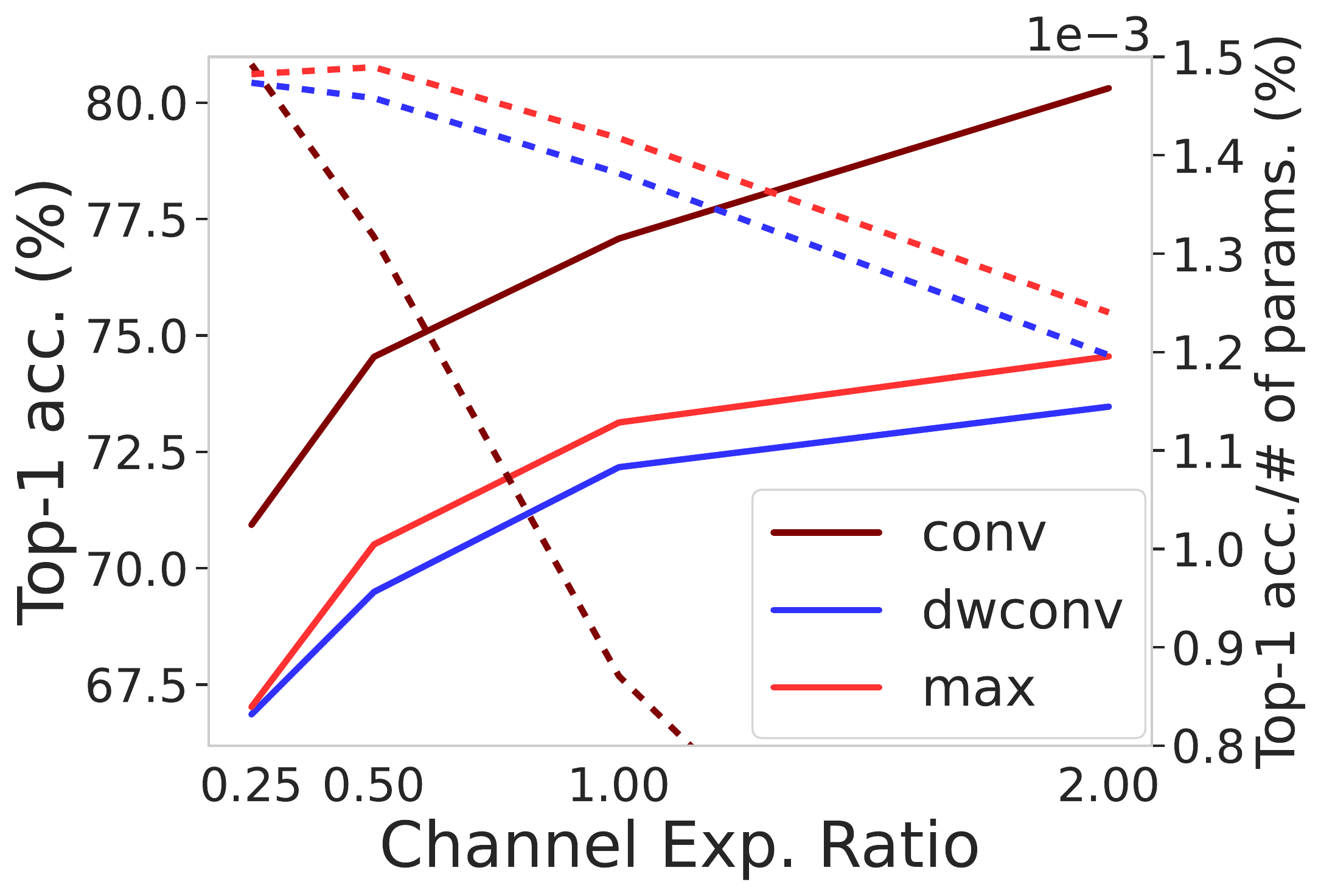}  
\end{subfigure} \ \
\begin{subfigure}[ht!]{0.27\linewidth}
\includegraphics[page=1, trim = 0mm 0mm 0mm 0mm, clip, width=1.0\linewidth]{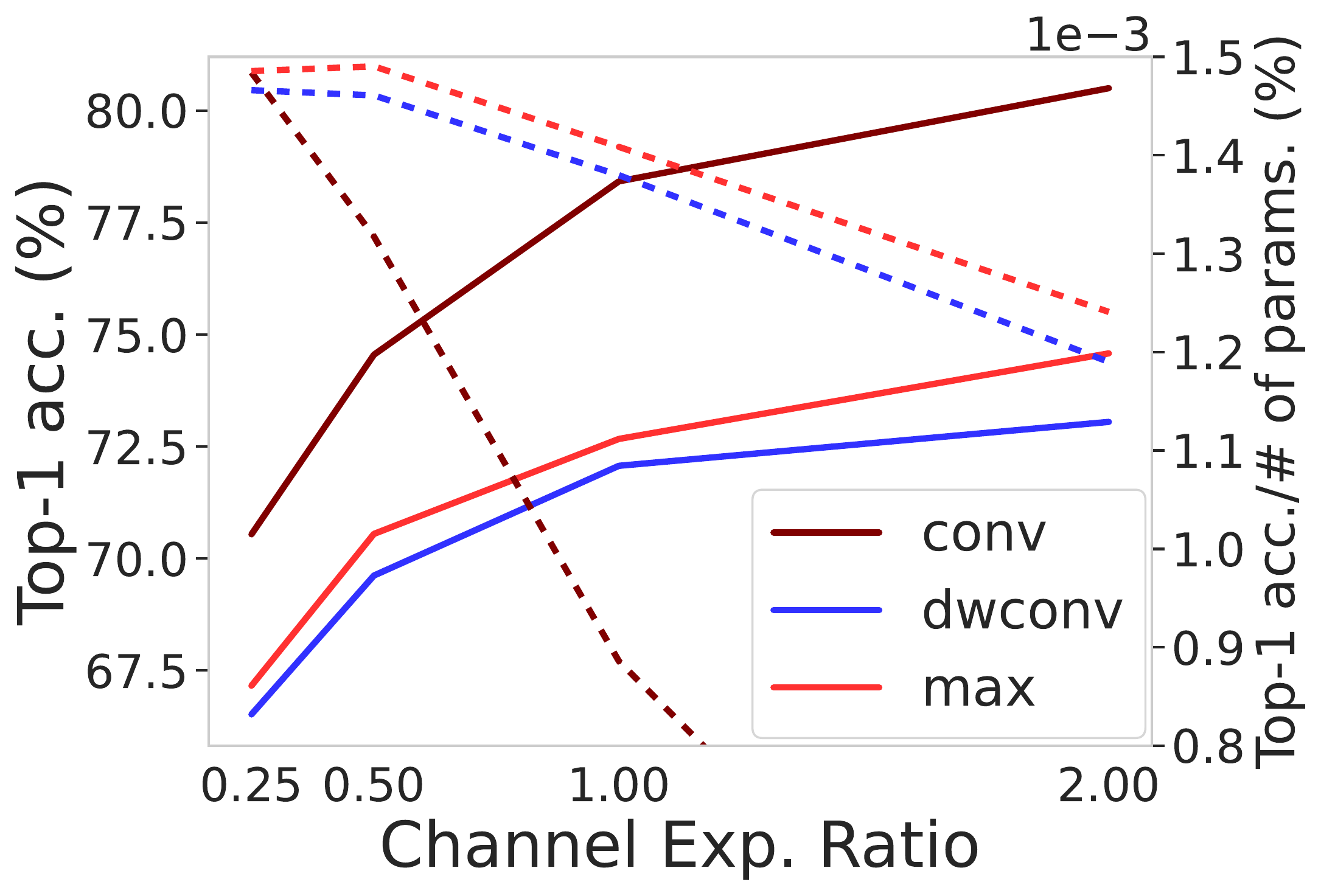}  
\end{subfigure} 
\vspace{-4mm}
\caption{\small {\bf Single bottleneck study.} We visualize top-1 accuracy ({\bf solid lines}) trained with the different setups including 1) varying channel expansion ratios inside a bottleneck; 2) the different channel widths: 32 ({\bf upper row}) and 64 ({\bf lower row}); 3) diverse optimizers: SGD ({\bf left}), AdamW ({\bf middle}), and AdamP ({\bf right}); We further plot accuracy per \# parameters ({\bf dashed lines}) to show the parameter-efficiency of the operations. We observe the regular convolutions work well but be replaceable at a low channel expansion ratio; the alternative operations are highly efficient; the max-pool consistently beats the depthwise convolution. } 
\label{fig:fig_cifar10_expansion_toy}
\vspace{-4mm}
\end{figure*}

\vspace{-2mm}
\subsection{Motivation}
\vspace{-2mm}
The learning mechanism of ResNets~\citep{resnet,preresnet} has been analyzed as the iterative unrolled estimation~\citep{greff2016highway,jastrzkebski2017residual} which iteratively refine the features owing to the presence of skip connections. This indicates that some layers do not contribute much to learning, which has also been revealed in layer-drop experiments~\citep{veit2016residual,stochasticdepth}. In this light, we argue some building blocks in a residual network can be replaced with parameter-free operations. We investigate the replacement of the spatial operation in popular networks with parameter-free operations to rethink the common practice of network design.%

\vspace{-2mm}
\subsection{Rethinking Parameter-Free Operations}
\label{sec_sub:rethinking}
\vspace{-2mm}
Our goal is to design an efficient building block that boosts model speed in practice while maintaining model accuracy. Based on \eqref{eq:bottleneck}, we remove the inter-channel operation similarly done in \eqref{eq:invbottleneck} and \eqref{eq:shift}. Then, instead of assigning $W_{v,h,w}$ to be a single value, as in \eqref{eq:shift}, we let $W$ have a dependency on the feature $g$ ($i.e.,$ $W_{v,h,w}{=}s(g_{v, r*i+h, r*j+w})$) by introducing a function $s(\cdot)$. Then, the layer would have different learning dynamics interacting with $P$ and $Q$. 

Among many candidates to formalize $s(\cdot)$, we allocate the function with a simple one that does not have trainable parameters. We let $s(\cdot)$ pick some large values of $g$ in the range of all $h,w$ per each $v$ like the impulse function. %
Here, we simplify the function $s(\cdot)$ again to use only the largest values of $g$, where we have the representation of $W_{v,h^*,w^*}{=}1,  (h^*,w^*){=}\text{argmax}_{(h,w)}g_{v,r*i+h,r*j+w}$ and other $W_{v,h,w}$ to be 0 in \eqref{eq:invbottleneck}. 
One may come up with another simple option: let $s(\cdot)$ compute the moments such as mean or variance of $g$. 
In practice, those operations can be replaced with built-in spatial pooling operations, which are highly optimized at the operation level. 
In the following section, we empirically study the effectivness of efficient parameter-free operations in network designs.
\vspace{-2mm}

\subsection{Empirical Studies}
\label{sec_sub:empirical_studies}
\vspace{-2mm}
\paragraph{On a Single Bottleneck.} Our study begins with a single bottleneck block of ResNet~\citep{resnet}. We identify such a parameter-free operation can replace the depthwise convolution which is a strong alternative of the regular convolution. %
We train a large number of different architectures on CIFAR10 and observe the accuracy trend of trained models. The models consist of a single bottleneck, but the channel expansion ratio inside a bottleneck varies from 0.25 to 2, and there are two options for the base-channel width (32 and 64). Training is done with three different optimizers of SGD, AdamW~\citep{loshchilov2017decoupled}, and AdamP~\citep{heo2020adamp}. Finally, we have total 4${\times}$2${\times}$3${\times}$3=72 models for the study. Fig.\ref{fig:fig_cifar10_expansion_toy} exhibit that the parameter-free operation consistently outperforms the depthwise convolution in a bottleneck\footnote{The bottleneck with the max-pool operation here is designed by following the formulation in \S\ref{sec_sub:rethinking}; we call it the efficient bottleneck. Note that the max-pool operation needs to be adopted due to the performance failures of using the avg-pool operation (see Appendix~\ref{supp_sec:pool_comparison}).}. The regular convolutions clearly achieved higher absolute accuracy, but the efficiency (the accuracy over the number of parameters) is getting worse for larger expansion ratios. We observe that the accuracy gaps between the regular convolution and the parameter-free operation are low for small channel expansion ratios (especially at the ratio=1/4 where the original bottleneck~\citep{resnet,resnext} adopts). It means that the regular convolution in the original bottleneck block with the expansion ratio of 1/4 can be replaced by the parameter-free operation.
\vspace{-2mm}

\begin{figure*}[t]
\small
\centering
\hspace{-1mm}
\begin{subfigure}[ht!]{0.245\linewidth}
\includegraphics[page=1, trim = 0mm 0mm 0mm 0mm, clip, width=1.0\linewidth]{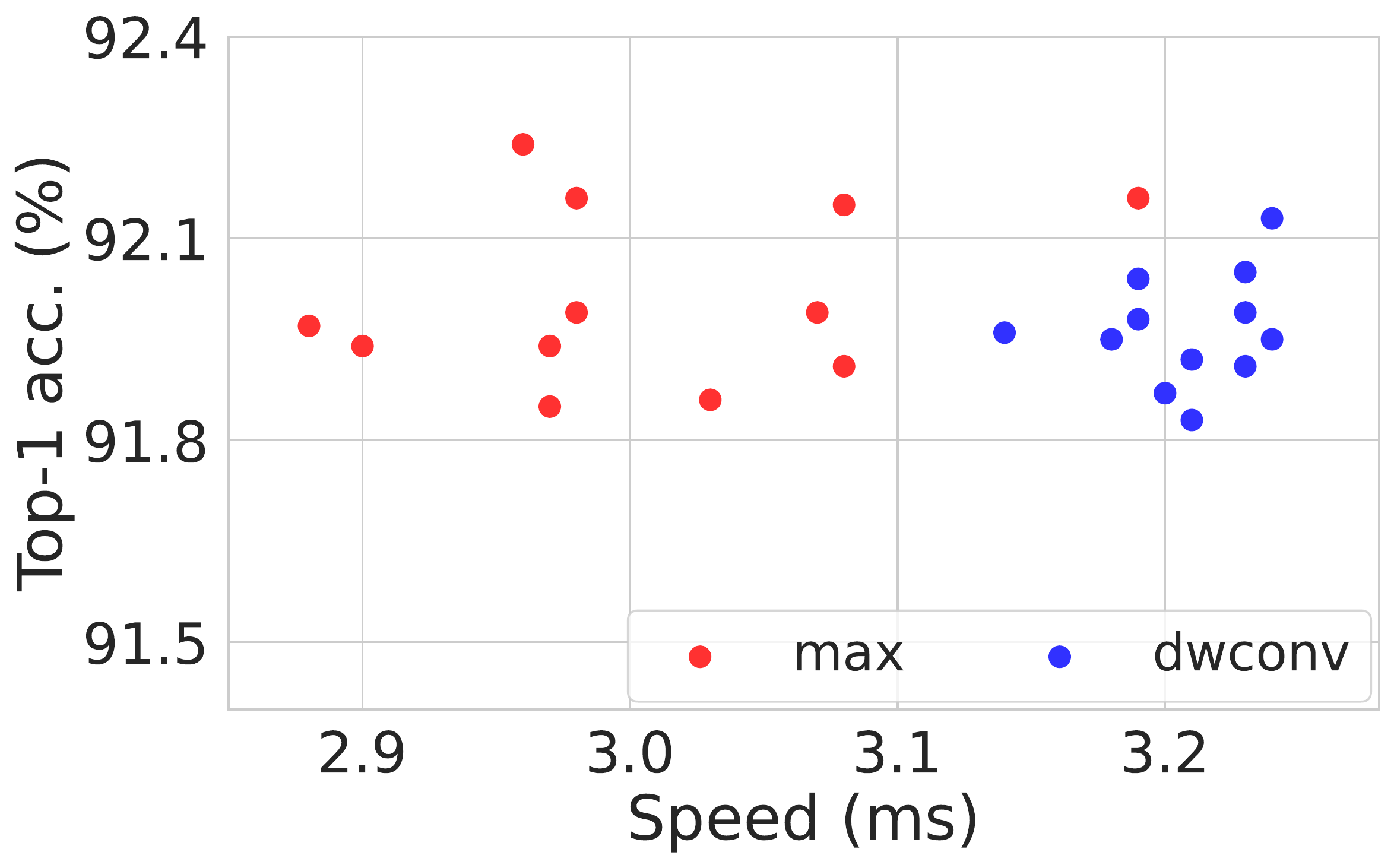}
\caption{\small Speed (1-image)}
\end{subfigure}  
\begin{subfigure}[ht!]{0.245\linewidth}
\includegraphics[page=1, trim = 0mm 0mm 0mm 0mm, clip, width=1.0\linewidth]{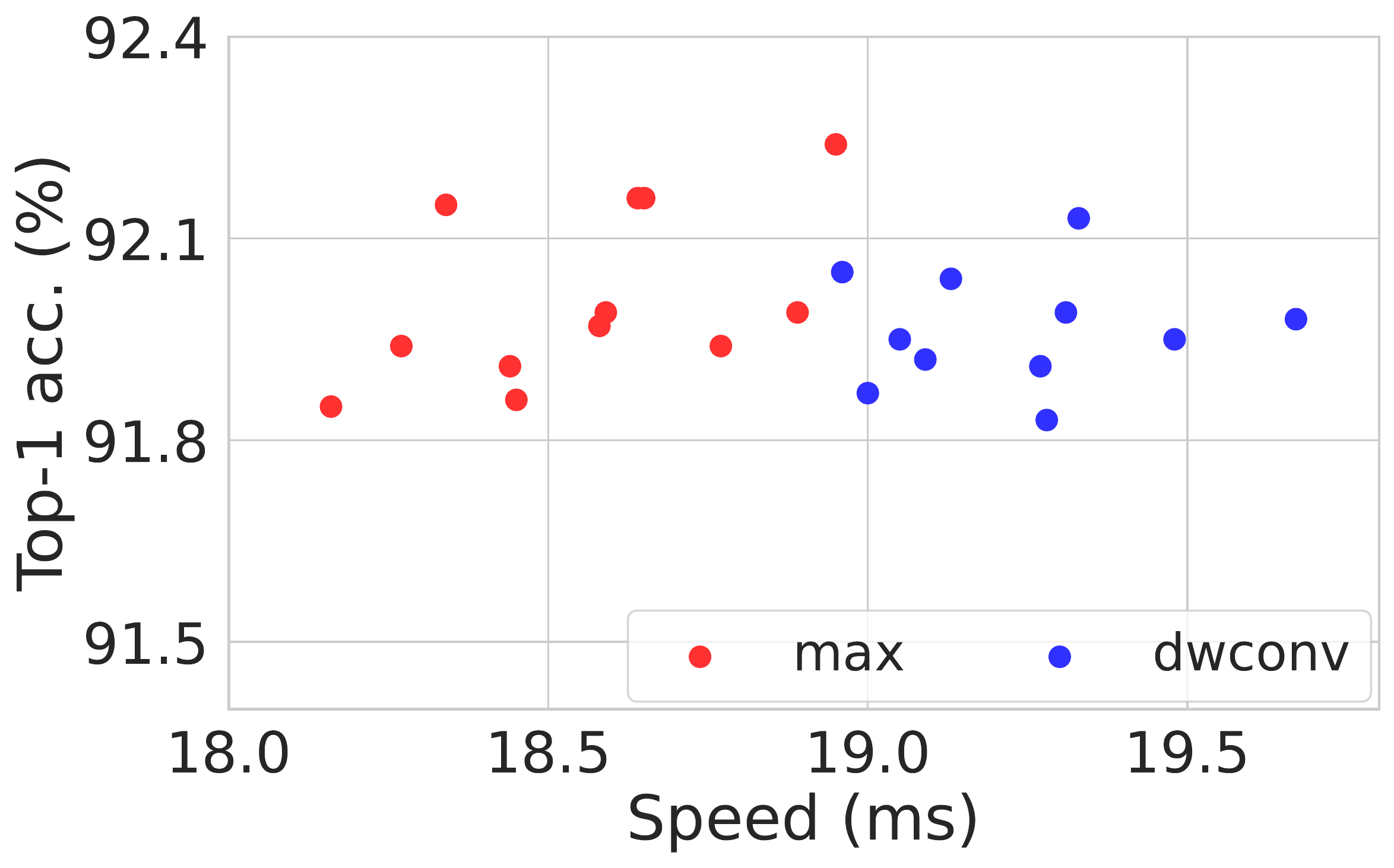}
\caption{\small Speed (256-images)}
\end{subfigure} 
\begin{subfigure}[ht!]{0.245\linewidth}
\includegraphics[page=1, trim = 0mm 0mm 0mm 0mm, clip, width=1.0\linewidth]{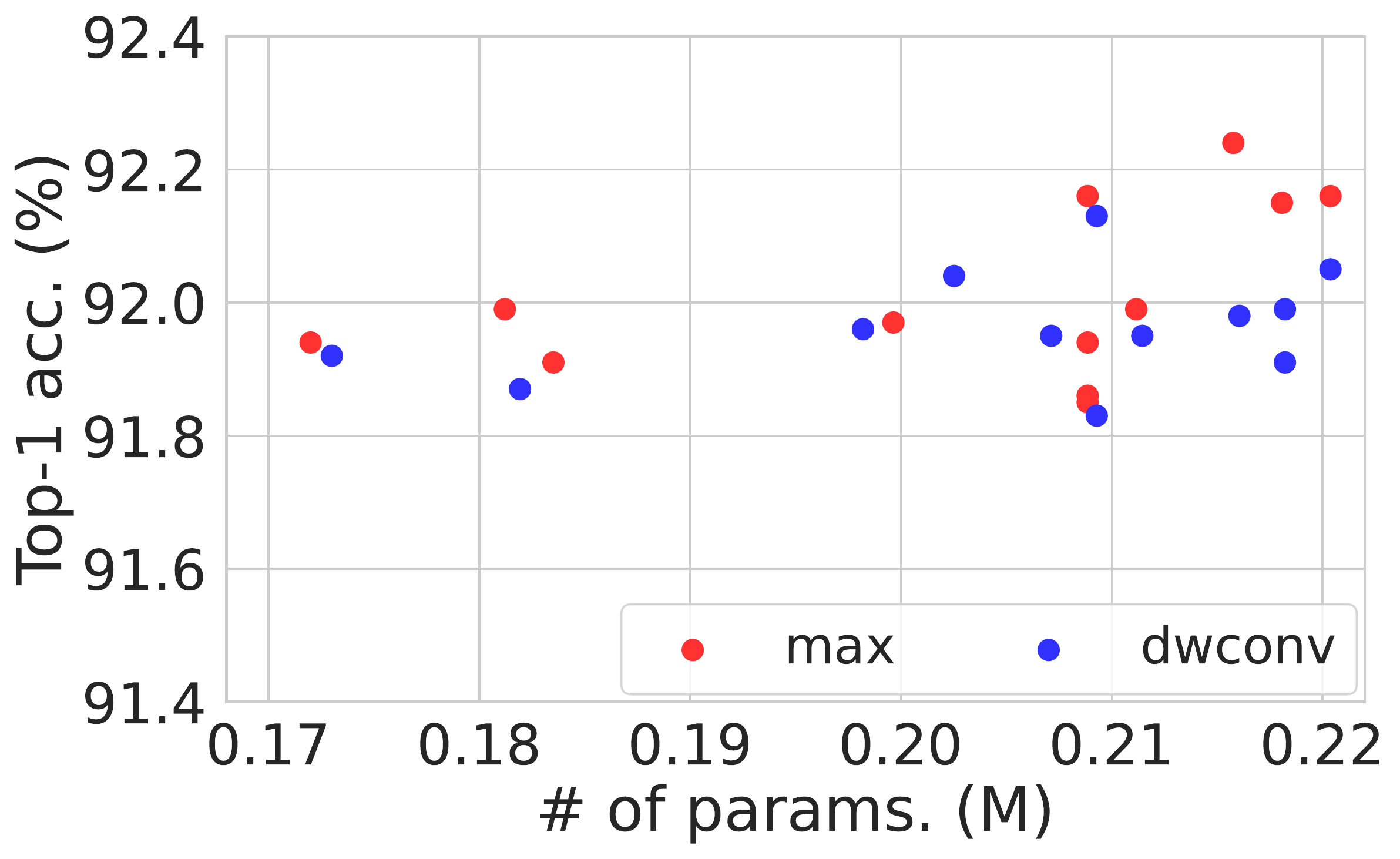}
\caption{\small \# parameters}
\end{subfigure} 
\begin{subfigure}[ht!]{0.245\linewidth}
\includegraphics[page=1, trim = 0mm 0mm 0mm 0mm, clip, width=1.0\linewidth]{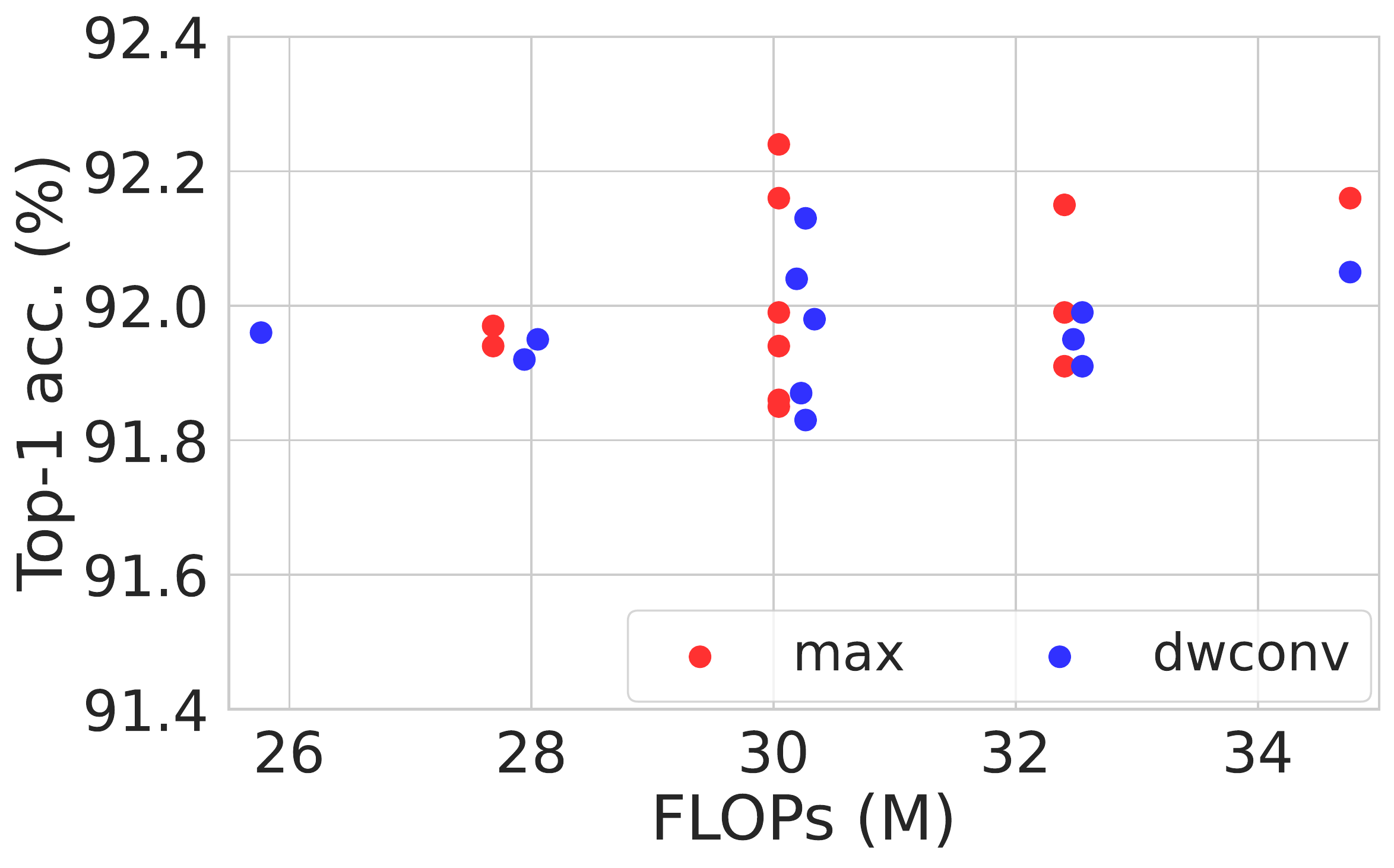}
\caption{\small FLOPs}
\end{subfigure} 
\vspace{-2mm}
\caption{\small {\bf Multiple bottlenecks study}. We respectively pick the 20\% best-performing models and visualize in the comparison graphs: (a) Accuracy vs. speed with the batch size of 1; (b) Accuracy vs. speed with the batch size of 256; (c) Accuracy vs. \# parameters; (d) Accuracy vs. FLOPs (use FLOPs to mean multiply-adds). The depthwise convolution ({\color{blue} blue dots}) is an efficient operation in \# parameters and FLOPs, but the parameter-free operation ({\color{red} red dots}) has a clear benefit in the model speed in practice. A particular parameter-free operation can be an alternative to the depthwise convolution when replacing the regular convolutions.}
\label{fig:fig_multilayer}
\vspace{-4mm}
\end{figure*}

\vspace{-1mm}
\paragraph{On Multiple Bottlenecks.}  We extend the single bottleneck study to deeper networks having multiple bottleneck blocks. We similarly study the bottleneck with different spatial operations, including the depthwise convolution and parameter-free operations, replacing the regular convolution. We choose the max-pool operation again as the parameter-free operation and compare it with the depthwise convolution. To study the impact of repeatedly using parameter-free operations in-depth, we exhaustively perform the layer replacements at layer-level upon the baseline network. We report the trade-off between accuracy and speed by replacing the regular convolutions with the max-pool operation and the depthwise convolution. We use the baseline network, which has eight bottleneck blocks in a standard ResNet architecture (i.e., ResNet-26), and the study has done by replacing each block have efficient operations (i.e., the efficient bottleneck and the bottleneck with the depthwise convolution). We train all the networks with large epochs (300 epochs) due to the depthwise convolution. Fig.\ref{fig:fig_multilayer} illustrates the trade-offs of top-1 accuracy and the actual model speed; we observe the models with the parameter-free operations are faster than those with the depthwise convolution as the model accuracies are comparable.
\vspace{-2mm}

\begin{table*}[h]
\fontsize{7.4}{8.4}\selectfont
\centering
\tabcolsep=0.10cm
\caption{\small {\bf Neural architecture search with individual cell searches.} We investigate how parameter-free operations are chosen with convolutional operations. Six different normal cells are searched individually, and two reduction cells are searched by a unified single cell. Surprisingly, the parameter-free operations are consistently found regardless of the settings. The parameter-free operations in each cell are represented as max @[({\color{blue} $d$}, $x_1$;$\dots ;x_i$)] and avg @[({\color{blue} $d$}, $x_1$;$\dots ;x_i$)] which denote the max-pool\_3${\times}$3 and the avg-pool\_3${\times}$3 appear at every edge towards $x_1 \dots x_i$-th nodes in {\color{blue}$d$}-th (from the input) cell, respectively. }
\label{table:darts_concat}
\vspace{-2.5mm}
\subfloat{\begin{tabular}{c|c|l|l|c}
\toprule
\# of Nodes & Seed No. & Normal Cells & Reduction Cell &  Prec-1 \\
\midrule
4 {\fontsize{5.5}{6.5}\selectfont (unified cells)}& 1 & {\color{red} no parameter-free ops.} & avg @[({\color{blue} 3}, 2), ({\color{blue} 6}, 2)] & 87.84\% \\ 
\midrule
1 & 1 &  max @[({\color{blue} 5}, 1)] & max @[({\color{blue} 3}, 1), ({\color{blue} 6}, 1)] & 86.15\% \\ 
1 & 2 &  max @[({\color{blue} 2}, 1)]  &  max @[({\color{blue} 3}, 1), ({\color{blue} 6}, 1)] & 86.27\% \\
1 & 3 &  max @[({\color{blue} 2}, 1)]  &  max @[({\color{blue} 3}, 1), ({\color{blue} 6}, 1)] & 85.42\% \\
\midrule
2 & 1 & max @[({\color{blue} 2}, 2), ({\color{blue} 5}, 1;2)] & {\color{red} no parameter-free ops.} & 87.64\% \\ 
2 & 2 & max @[({\color{blue} 4}, 2), ({\color{blue} 5}, 1;2), ({\color{blue} 7}, 2)], avg @[({\color{blue} 7}, 1)] & {\color{red} no parameter-free ops.} & 87.70\% \\
2 & 3 & max @[({\color{blue} 2}, 1), ({\color{blue} 5}, 1;2)] & {\color{red} no parameter-free ops.} & 87.65\% \\
\midrule
3 & 1 & max @[({\color{blue} 5}, 1;2;3), ({\color{blue} 7}, 1;2;3] & max @[({\color{blue} 3}, 3), ({\color{blue} 6}, 3)] & 88.09\% \\ 
3 & 2 & max @[({\color{blue} 4}, 2), ({\color{blue} 5}, 2), ({\color{blue} 7}, 2)], avg @[({\color{blue} 7}, 1)] & max @[({\color{blue} 3}, 3), ({\color{blue} 6}, 3)] & 88.06\% \\
3 & 3 & max @[({\color{blue} 4}, 1;3), ({\color{blue} 5}, 3)], avg @[({\color{blue} 5}, 2)] & {\color{red} no parameter-free ops.} & 88.03\% \\
\midrule
4 & 1 &  max @[({\color{blue} 2}, 3;4), ({\color{blue} 5}, 1;2;3;4), ({\color{blue} 7}, 1)], avg @[({\color{blue} 8}, 1;2)] & avg @[({\color{blue} 3}, 1;2), ({\color{blue} 6}, 1;2)] & 88.12\% \\ 
4 & 2 & max @[({\color{blue} 4}, 1;3;4), ({\color{blue} 5}, 2;3;4), ({\color{blue} 7}, 1;2;3;4)], avg @[({\color{blue} 5}, 1)] & {\color{red} no parameter-free ops.}  & 87.79\% \\
4 & 3 &  max @[({\color{blue} 2}, 3), ({\color{blue} 7}, 1;2)], avg @[({\color{blue} 4}, 1;2;3), ({\color{blue} 8}, 2)] &  avg @[({\color{blue} 3}, 2), ({\color{blue} 6}, 2)] & 87.54\% \\
\bottomrule
\end{tabular}}
\vspace{-4.5mm}
\end{table*}

\paragraph{On Neural Architecture Searches.} We further investigate whether parameter-free operations are likely to be chosen for model accuracy along with trainable operations in a neural architecture search (NAS). We choose DARTS~\citep{darts} with the default architecture of eight cells and perform searches on CIFAR-10~\citep{cifar}\footnote{We follow the original training settings of DARTS~\citep{darts}. Additionally, to avoid the collapse of DARTS due to the domination of skip connections in learned architectures, we insert the dropout at the \textsc{skip\_connect} operation proposed in the method~\citep{chen2019progressive}.}. Towards a more sophisticated study, we refine the search configuration as follows. First, we force the entire normal cells to be searched individually rather than following the default setting searching one unified normal cell. This is because when searching with the unified normal cell, it is more likely to be searched without the parameter-free operations to secure the expressiveness of the entire network (see the search result at the first row in Table~\ref{table:darts_concat}). Therefore we prevent this with the more natural configuration. Second, the operation list is changed to ensure fairness of the expressiveness among the operations; since the primitive DARTS has the default operation list with the separable convolutions (\textsc{sep\_conv\_3${\times}$3, sep\_conv\_5${\times}$5}) and the dilated separable convolutions (\textsc{dil\_conv\_3${\times}$3, dil\_conv\_5${\times}$5}) which consist of multiple convolutions and ReLU with BN. Thus, we need to simplify the primitive operations to \textsc{[max\_pool\_3${\times}$3, avg\_pool\_3${\times}$3, conv\_1${\times}$1, conv\_3${\times}$3, dw\_conv\_3${\times}$3, zero, skip\_connect]}. Finally, we further perform searches with different numbers of nodes inside each cell to confirm the trend of search results regarding the architectural constraint. All the searches are repeated with three different seed numbers.

\begin{wrapfigure}{r}{0.5\linewidth}
\small
\centering
\hspace{-4mm}
\begin{subfigure}[ht!]{0.5\linewidth}
\vspace{-2mm}
\includegraphics[page=1, trim = 0mm 0mm 0mm 0mm, clip, width=1.0\linewidth]{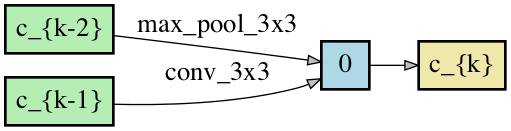}
\caption{\small 5-th normal cell}
\end{subfigure} 
\begin{subfigure}[ht!]{0.4\linewidth}
\vspace{-4mm}
\includegraphics[page=1, trim = 0mm 0mm 0mm 0mm, clip, width=1.0\linewidth]{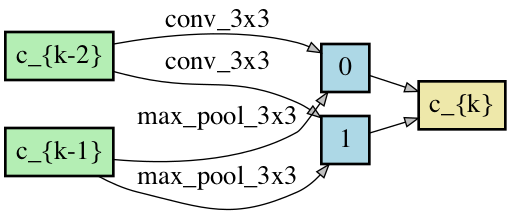}
\caption{\small 5-th normal cell}
\end{subfigure}  \\
\begin{subfigure}[ht!]{0.4\linewidth}
\includegraphics[page=1, trim = 0mm 0mm 0mm 0mm, clip, width=1.0\linewidth]{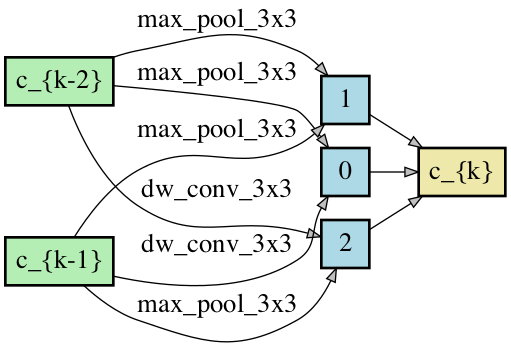}
\caption{\small 7-th normal cell}
\end{subfigure} 
\begin{subfigure}[ht!]{0.55\linewidth}
\vspace{7mm}
\includegraphics[page=1, trim = 0mm 0mm 0mm 0mm, clip, width=1.0\linewidth]{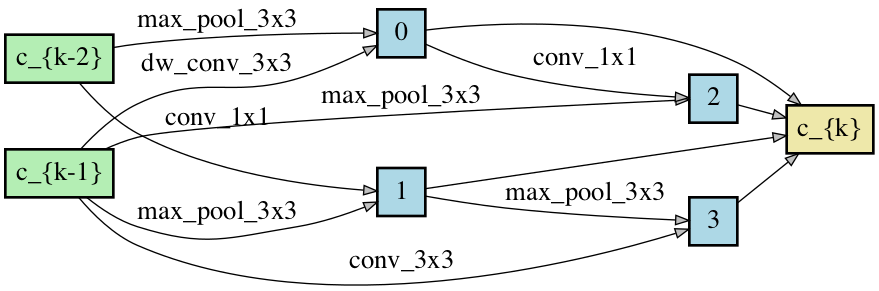}
\caption{\small 5-th normal cell}
\end{subfigure} 
\vspace{-3mm}
\caption{\small {\bf Visualization of the searched cells.} The example searched cells shown in Table~\ref{table:darts_concat} are visualized.}
\label{fig:fig_darts}
\vspace{-3mm}
\end{wrapfigure}

As shown in Table~\ref{table:darts_concat} and Fig.\ref{fig:fig_darts}, we observe the parameter-free operations are frequently searched in the normal cells, which is different from the original search result (at the first row) similarly shown in 
 the previous works~\citep{darts,chen2019progressive,liang2019darts+,chu2020fair}. Moreover, the accuracies are similar to the original setting with two unified cells, where the parameter-free operations are not searched in the normal cell (see the first and the last rows). Interestingly, the number of the picked parameter-free operations increases as the number of nodes increases; they do not dominate the reduction cells. As a result, it turns out that the accuracy objective in searches lets the parameter-free operations become chosen, and therefore the operations can be used at certain positions which looks similar to the Inception blocks~\citep{GoogleNet,BN,Inceptionv3}. Notice that similar results are produced with different settings and seeds.%

\section{Designing Efficient Deep Neural Networks}
\label{sec:design}
\vspace{-2mm}

Based on the studies above, parameter-free operations can be a building block of designing a network architecture. We apply the parameter-free operations to redesign a deeper neural network architecture, including a convoluption neural network (CNN) and a vision transformer (ViT).
\vspace{-2mm}
\subsection{Efficient CNN Architectures}

\vspace{-2mm}
\paragraph{Hybrid Architecture with Efficient Bottlenecks.}
We employ the proposed efficient bottleneck to design deeper efficient CNNs. The max-pool operation plays a role of the spatial operation in the efficient bottleneck;
specifically, the 3${\times}$3 convolution, BN, and ReLU triplet are replaced with a max-pool operation\footnote{We found that parameter-free operations such as the max-pool operation without following BN and ReLU do not degrade accuracy. Mathematically, using ReLU after the max-pool operation (without BN) does not give nonlinearity due to the preceding ReLU. Empirically, involving BN after a parameter-free operation could stabilize the training initially, but model accuracy was not improved in the end.}. To design the entire network architecture, fully utilizing the efficient bottlenecks would yield a much faster inference speed model, but a combination of the efficient bottleneck and the regular bottleneck has a better trade-off between accuracy and speed. We use ResNet50~\citep{resnet} as the baseline to equip a simple heuristic of building a hybrid architecture of the efficient bottleneck and the regular bottleneck. The observation from the NAS searches in \S\ref{sec_sub:empirical_studies}, the parameter-free operations are not explored much at the downsampling blocks but normal blocks. Thus, we similarly design a new efficient network architecture using the efficient bottlenecks as the fundamental building block except for the downsampling positions with the regular bottlenecks. We call this model {\it hybrid} architecture. The details of the architectures are illustrated in Appendix~\ref{supp_sec:architectures}. 

\vspace{-2mm}
\paragraph{Architectural Study.}
The architectural study is further conducted with the replacements of the regular spatial convolutions in the regular bottlenecks. We involve the efficient bottlenecks into the stages of ResNet50, replacing the existing regular bottlenecks exhaustively to cover the design space as much as possible. It is worth noting that performing a neural architecture search would yield a precise design guide, but a direct search on ImageNet is costly. We use the ImageNet dataset to provide a more convincing result of the investigation. We also report the performance of the competitive  \begin{wraptable}{r}{0.45\linewidth}
\small
\centering
\tabcolsep=0.15cm
\caption{\small {\bf Model study on ResNet50}. We report the performance of {\it hybrid} model, which is the most promising model compared with others, and the models with stage-level combinations of the regular bottleneck and the efficient bottleneck (denoted by $\bf B$ and $\bf{E}$, respectively).%
We further report the performance of two variant models by replacing each 3${\times}$3 convolution with 1) a 1${\times}$1 convolution; 2) a 3${\times}$3 depthwise convolution in all the regular bottlenecks. The two best trade-off models are in {\bf boldface}.}
\label{table:resnet_study}
\vspace{-2mm}
\begin{tabular}{@{}c|c|c@{}}
\toprule
Models & GPU Lat. (ms)& Top-1 acc.(\%) \\
\midrule
{\it Hybrid} & {\bf 7.3}  & {\bf 74.9} \\
{\bf E} / {\bf B} & {\bf 7.7} & {\bf 75.3} \\
\midrule
{\bf B}$\rightarrow${\bf B}$\rightarrow${\bf B}$\rightarrow${\bf B} & 8.7 & 76.2\\
{\bf E}$\rightarrow${\bf B}$\rightarrow${\bf B}$\rightarrow${\bf B} & 8.4 & 75.4\\
{\bf E}$\rightarrow${\bf E}$\rightarrow${\bf B}$\rightarrow${\bf B} & 8.0 & 75.1 \\
{\bf E}$\rightarrow${\bf E}$\rightarrow${\bf E}$\rightarrow${\bf B} & 7.3 & 73.1 \\
{\bf E}$\rightarrow${\bf E}$\rightarrow${\bf E}$\rightarrow${\bf E} & 6.8 & 72.0\\
{\bf B}$\rightarrow${\bf B}$\rightarrow${\bf B}$\rightarrow${\bf E} & 8.4 & 75.4\\
{\bf B}$\rightarrow${\bf B}$\rightarrow${\bf E}$\rightarrow${\bf E} & 7.7 & 74.6\\
{\bf B}$\rightarrow${\bf E}$\rightarrow${\bf E}$\rightarrow${\bf E} & 7.3 & 74.0 \\
\midrule
1${\times}$1 conv  & 6.2 & 30.1 \\
3${\times}$3 dwconv & 8.3 & 74.9 \\
\bottomrule
\end{tabular}
\vspace{-2mm}
\end{wraptable}
 architectures replacing the spatial operations ($i.e.,$ the 3${\times}$3 convolutions)  with 1) the 1${\times}$1 convolutions; 2) the 3${\times}$3 depthwise convolution in entire stages.  

Table~\ref{table:resnet_study} shows the trade-off between accuracy and speed of the different models mixing the regular and the efficient bottlenecks. {\bf B}$\rightarrow${\bf B}$\rightarrow${\bf B}$\rightarrow${\bf B} denotes the baseline ResNet50; {\bf E} / {\bf B} denotes the model using the regular bottleneck and the efficient block alternately. Among the different models, we highlight {\it hybrid} and {\bf E} / {\bf B} models due to the promising performance. This shows that only a simple module-based design regime can achieve improved performance over the baseline. {\bf E}$\rightarrow${\bf E}$\rightarrow${\bf E}$\rightarrow${\bf E} is the model using the efficient bottlenecks only and shows the fastest speed as we expected. The model with the 1${\times}$1 convolution replacements cannot reach the accuracy standard presumably due to the lack of the receptive field; this is why a sufficient number of spatial operations are required. We observe that the model with the depthwise convolutions shows a promising performance, but our models surpass it as we similarly observed in \S\ref{sec_sub:empirical_studies}. All the models are trained on ImageNet with the standard 90-epochs training setup\footnote{Trainings are done with the fixed image size 224${\times}$224 and the standard data augmentation~\citep{GoogleNet} with the random\_resized\_crop rate from 0.08 to 1.0. We use stochastic gradient descent (SGD) with Nesterov momentum~\citep{nag} with momentum of 0.9 and mini-batch size of 256, and learning rate is initially set to 0.4 by the linear scaling rule~\citep{goyal2017accurate} with step-decay learning rate scheduling; weight decay is set to 1e-4. The accuracy of the baseline ResNet50 has proven the correctness of the setting.}~\citep{preresnet} to report the performance.  
\vspace{-2mm}

\subsection{Efficient ViT Architectures}
\vspace{-2mm}
We provide a further use case of the parameter-free operations in a totally different architecture from CNN. We again choose the max-pool operation as the parameter-free operation and apply it to replace the self-attention layer~\citep{transformer} in vision transformer (ViT)~\citep{dosovitskiy2020image} to observe how the efficient operation can replace a complicated layer. We do not change the MLP in a transformer block but replace the self-attention module with the max-pool operation. Specifically, after the linear projection with the ratio of 3 ($i.e.,$ identical to the concatenation of the projection of query, key, and value in the self-attention layer in ViT), the projected 1d-features are reshaped to 2d-features for the input of the spatial operation. Consequently, the self-attention layer is replaced with a cheaper parameter-free operation and will bring clear efficiency. We use the global average pooling (GAP) layer instead of using the classification token of ViT since the classification token is hardly available without the self-attention layer. Many transformer-based ViTs can be a baseline; we additionally adopt a strong baseline Pooling-based Vision Transformer (PiT)~\citep{heo2021rethinking} to show applicability. More details are elaborated in Appendix~\ref{supp_sec:architectures}.

\section{Experiments}
\label{sec:exp}
\vspace{-2mm}
\subsection{ImageNet Classification}
\label{sec_sub:imagenet_exp}
\vspace{-2mm}
\paragraph{Efficient ResNets.}
We perform ImageNet~\citep{ImageNet} trainings to validate the model performance. We adopt the standard architecture ResNet50~\citep{resnet} as the baseline and train our models with the aforementioned standard 90-epochs training setting to fairly compare with the competitors~\citep{luo2017thinet, huang2018data, wu2018shift, wang2018versatile, han2020ghostnet, yu2019slimmable, qiu2021slimconv}, where the efficient operators were proposed or the networks were pruned. We report each averaged speed of the models with the publicly released codes on a V100 GPU. Table~\ref{table:imagenet_comparison} shows our networks acheieve faster inference speeds than those of the competitors with the comparable accuracies. The channel-width pruned models (Slimmable-R50 0.5${\times}$, 0.75${\times}$) and the models with new operations (Veratile-R50, GhostNet-R50, and SlimConv-R50) cannot reach the model speed to ours. %
Additionally, we report the improved model accuracy with further training tricks (see the details in Appendix~\ref{supp_sec:training_recipes}). This aims to show the maximal capacity of our model even using many parameter-free operations inside. The last rows in the table present our model can follow the baseline accuracy well, and we see the gap between the baseline and ours has diminished; this show a potential of using parameter-free operations for a network design.

\begin{table*}[t]
\small
\fontsize{8.5}{9.5}\selectfont
\centering
\tabcolsep=0.1cm
\caption{\small {\bf ImageNet performance comparison of efficient models.} We report the model performance including accuracy, the number of parameters, FLOPs, and the GPU latency measured on a V100 GPU. All the model speeds are measured by ourselves using the publicly released architectures. $^\dagger$: used further training recipes.}
\label{table:imagenet_comparison}
\vspace{-2mm}
\subfloat{\begin{tabular}{l|c|c|c|c|c} %
\toprule
Network Architecture & Params. (M) & FLOPs (G) & GPU (ms)  & Top-1 (\%) & Top-5 (\%) \\
\midrule
ResNet50 (R50)~\citep{resnet} & 25.6 & 4.1 & 8.7 & 76.2 & 93.8 \\
\midrule
Thinet70-R50~\citep{luo2017thinet} & 16.9 & 2.6  & - & 72.1 & 90.3 \\
SSS-R50~\citep{huang2018data} & 18.6 & 2.8  & - & 74.2 & 91.9 \\
Shift-R50~\citep{wu2018shift} & 22.x & N/A & - & 75.6 & 92.8 \\
Versatile-R50~\citep{wang2018versatile} & 17.1 & 1.8 & 18.7 & 75.5 & 92.4\\
Sllimable-R50 (0.5${\times}$)~\citep{yu2019slimmable} & 6.9 & 1.1 & 8.5 & 72.1 & N/A \\
Sllimable-R50 (0.75${\times}$)~\citep{yu2019slimmable} & 14.8 & 2.4 & 8.6 & 74.9 & N/A \\
GhostNet-R50~\citep{han2020ghostnet} & 14.0 & 2.1 & 20.3 & 75.0 & 92.3 \\
SlimConv-R50 ($k{=}8/3$)~\citep{qiu2021slimconv} & 12.1 & 1.9 & 24.5 & 75.5 & N/A \\
\midrule
Ours-R50 (max)  & 14.2 & 2.2 & 6.8 & 72.0 & 90.5  \\
Ours-R50 ({\it hybrid})  & 17.3 & 2.6 & 7.3 & 74.9 & 92.2 \\
Ours-R50 (deform\_max) & 18.0 & 2.9 & 10.3 & 75.3 & 92.5 \\
\midrule
Ours-R50 (max)$^\dagger$  & 14.2 & 2.2 & 6.8 & 74.3 & 92.0  \\
Ours-R50 ({\it hybrid})$^\dagger$  & 17.3 & 2.6 & 7.3 & 77.1 & 93.1 \\
Ours-R50 (deform\_max)$^\dagger$ & 18.0 & 2.9 & 10.3 & 78.3 & 93.9 \\
\bottomrule
\end{tabular}}
\vspace{-2mm}
\end{table*}

\vspace{-3mm}
\paragraph{Bigger Network Architectures.}
\begin{table*}[t]
\fontsize{8.0}{8.8}\selectfont
\centering
\tabcolsep=0.07cm
\caption{\small {\bf ImageNet performance of CNN models.} We report the ImageNet performance, mCE (ImageNet-C) and AUC (ImageNet-O) of the diverse CNN models. All the redesigned models experience massive reductions of the computational costs and the substantial gains on mCE and AUC but barely drop the accuracy (${<}$2.0\%).}
\label{table:imagenet_bigger_models}
\vspace{-2mm}
\begin{tabular}{l|c|c|c|c|c|c|c|c}
\toprule
Network & Params. (M)$\downarrow$ & FLOPs (G)$\downarrow$  & GPU (ms)$\downarrow$  & CPU (ms)$\downarrow$ & Top-1 (\%)$\uparrow$  & Top-5 (\%)$\uparrow$ & mCE (\%)$\downarrow$ & AUC (\%)$\uparrow$  \\
\midrule
ResNet50 & 25.6 & 4.1 & 8.7 & 45.4 & 78.5 & 94.2 & 63.8 & 51.7 \\
Ours-R50 & 17.3 \greenpscript{-33\%} & 2.6 \greenpscript{-37\%}& 7.3 \greenpscript{-17\%} & 39.8 \greenpscript{-12\%}& 77.1 \redpscript{-1.4}& 93.1 \redpscript{-1.1} & 57.5 \greenpscript{-6.3} & 52.9 \greenpscript{+1.2} \\
\midrule
ResNet50-SE & 28.1 & 4.1  & 13.9  &  98.0  & 79.5 & 94.7 & 69.6 & 58.9 \\
Ours-R50-SE & 19.9 \greenpscript{-29\%} & 2.6 \greenpscript{-37\%}& 12.5 \greenpscript{-10\%} & 92.6 \greenpscript{-9\%} & 78.2 \redpscript{-1.3}& 93.9 \redpscript{-0.8}& 63.8 \greenpscript{-5.8} & 59.2 \greenpscript{+0.3}\\
\midrule
ResNet101 & 44.6 & 7.8 & 16.7 & 80.5 & 80.1 & 94.9 & 69.7 & 51.7 \\
Ours-R101 & 26.3 \greenpscript{-41\%} & 4.3 \greenpscript{-45\%} & 13.5 \greenpscript{-19\%}& 65.7 \greenpscript{-18\%} & 78.2 \redpscript{-1.9}& 93.8 \redpscript{-1.1}& 62.0 \greenpscript{-7.7} & 54.6 \greenpscript{+2.9}\\
\midrule
WRN50-2 & 68.9 & 11.4 & 9.0  & 84.0 & 79.7 & 94.7 & 67.0 & 49.8 \\
Ours-WRN50-2 & 36.0 \greenpscript{-48\%} & 5.4 \greenpscript{-53\%} & 7.3 \greenpscript{-20\%} & 59.8 \greenpscript{-29\%} & 78.1 \redpscript{-1.6}& 93.8 \redpscript{-0.9}& 60.9 \greenpscript{-6.1} & 51.9 \greenpscript{+2.1}\\
\midrule
WRN101-2 & 126.9 & 22.8 & 16.9  & 146.2 & 80.9 & 95.3 & 73.2  & 50.6 \\
Ours-WRN101-2 & 53.9 \greenpscript{-58\%} & 8.9 \greenpscript{-61\%} & 13.7 \greenpscript{19\%} & 96.3 \greenpscript{-34\%} & 78.9 \redpscript{-2.0}& 94.2 \redpscript{-1.1} & 63.4 \greenpscript{-9.8} & 54.8 \greenpscript{+4.2}\\
\bottomrule
\end{tabular}
\vspace{-3mm}
\end{table*}
Our design regime is applied to complicated network architectures such as ResNet50-SE~\citep{SENet} and Wide ResNet-101-2 (WRN101-2)~\citep{wideresnet}. We report the ImageNet performance plus mean Corruption Error (mCE) on ImageNet-C~\citep{hendrycks2019benchmarking} and Area Under the precision-recall Curve (AUC) the measure of out-of-distribution detection
performance on ImageNet-O~\citep{hendrycks2021natural}. Table~\ref{table:imagenet_bigger_models} indicates the models redesigned with the parameter-free operations work well; bigger models are more significantly compressed in the computational costs. Additionally, we notice all the mCEs and AUC of our models are remarkably improved, overtaking the accuracy degradation, which means the models with parameter-free operations suffer less from overfitting. %

\vspace{-3mm}
\paragraph{Deformable Max-pool Operation.} We manifest a future direction of utilizing a parameter-free operation. We borrow a similar idea of the deformable convolution\footnote{We implement the operation upon the code: \url{ https://github.com/CharlesShang/DCNv2}.}~\citep{dai2017deformable,zhou2017}. Specifically, we identically involve a convolution to interpolate the features predicted by itself to perform a parameter-free operation on. The max-pool operation still covers the spatial operation; only the offset predictor has weights to predict the locations. We regard this operator as to how the performance of the max-pool operation can be stretched when involving few numbers of parameters. Only performing computation on predicted locations can improve the accuracy over the vanilla computation with a few extra costs, as shown in Table~\ref{table:imagenet_comparison}. Note that there is room for faster operation speed as the implementation can be further optimized.
\vspace{-2mm}

\subsection{COCO Object Detection}
\vspace{-2mm}

We verify the transferability of our efficient backbones on the COCO2017 dataset~\citep{coco2017}. We adopt the standard Faster RCNN~\citep{fasterrcnn} with FPN~\citep{fpn} without any bells and whistles to finetune the backbones following the original training settings~\citep{fasterrcnn,fpn} Table~\ref{table:coco_frcnn} shows our models achieve a better trade-off between the AP scores and the computational costs, including the model speed, and does not significantly degrade the AP scores even inside massive parameter-free operations. We further validate the backbone with the deformable max-pool operations in Table~\ref{table:coco_frcnn}. Strikingly, AP scores are improved over ResNet50 even with the low ImageNet accuracy; this shows the effectiveness of the operation in a localization task. %

\vspace{-2mm}
\begin{table*}[t]
\small
\centering
\fontsize{8.5}{9.5}\selectfont
\tabcolsep=0.08cm
\caption{\small {\bf COCO object detection results.} All the models are finetuned on \texttt{train2017} by ourselves using the ImageNet-pretrained backbones in Table~\ref{table:resnet_study}. We report box APs on \texttt{val2017}.}
\label{table:coco_frcnn}
\vspace{-2mm}
\begin{tabular}{l|c|c|ccc|c|c|c}
\toprule
\multirow{2}{*}{Backbone} & 
\multirow{2}{*}{IN Acc. (\%)} & 
\multirow{2}{*}{Input Size} & 
\multicolumn{3}{c|}{Bbox AP at IOU} &
\multirow{2}{*}{GPU (ms)} &
\multirow{2}{*}{Params. (M)} &
\multirow{2}{*}{FLOPs (G)}  \\
& & & AP & AP$_{\text{50}}$ & AP$_{\text{75}}$ & & \\
\midrule
ResNet50                & 76.2 & 1200${\times}$800 & 32.9 & 51.8 & 35.1 & 42.8 & 41.8 & 202.2  \\
ResNet50 ({\it hybrid})& 74.9 & 1200${\times}$800 & 31.9 \redpscript{-1.0} & 51.5 & 34.1 & 37.5 \greenpscript{-12\%}& 33.6  \greenpscript{-20\%} & 175.8 \greenpscript{-13\%}\\
ResNet50 (deform\_max)& 75.3 & 1200${\times}$800 & 33.2\greenpscript{+0.3} &  53.0 &  35.3 & 47.7\redpscript{+11\%} & 34.3 \greenpscript{-18\%}& 181.8 \greenpscript{-10\%}\\
\bottomrule
\end{tabular}
\vspace{-3mm}
\end{table*}

\begin{table}[t]
\small
\fontsize{8.5}{9.5}\selectfont
\centering
\caption{{\bf ImageNet performance of ViTs.} We report ViT models' performance trained on ImageNet with diverse training settings denoted by Vanilla, CutMix, and DeiT (with strong augmentations).}
\label{table:vit}
\vspace{-2mm}
\begin{tabular}{l|c|c|c|c|c}
\toprule
\multirow{2}{*}{Model} & \multicolumn{2}{c|}{Throughput (imgs/sec)} & \multirow{2}{*}{Vanilla} & \multirow{2}{*}{+CutMix} & \multirow{2}{*}{+DeiT} \\ \cmidrule(lr){2-3}
& 256-batch & 1-batch & & & \\
\midrule
ResNet50 & 962 & {\bf 112} & 76.2 & 77.6 & 78.8 \\
\midrule
ViT-S & 787 & 86 & 73.9 & 77.0 & 80.6\\
ViT-S (dwconv) & 571 \redpscript{-216} & 95 \greenpscript{+9} & 76.1 \greenpscript{+2.2} & 78.7 \greenpscript{+1.7} & 81.2 \greenpscript{+0.6} \\
ViT-S (max-pool) & 763 \redpscript{-24} & 96 \greenpscript{+10} & 74.2 \greenpscript{+0.3} & 77.3 \greenpscript{+0.3} & 80.0 \redpscript{-0.6} \\
\midrule
PiT-S & 952 & 57 & 75.5 & 78.7 & 81.1\\
PiT-S (dwconv) & 781 \redpscript{-171} & 90 \greenpscript{+33} & 76.1 \greenpscript{+0.6} & 78.6 \redpscript{-0.1} & 81.0  \redpscript{-0.1} \\
PiT-S (max-pool) & {\bf 1000} \greenpscript{+48} & 92 \greenpscript{+35} & 75.7 \greenpscript{+0.2} & 78.1 \redpscript{-0.6} & 80.8 \redpscript{-0.3}\\
\bottomrule
\end{tabular}
\vspace{-3mm}
\end{table}

\subsection{ImageNet Classification with Efficient Vision Transformers}
\vspace{-2mm}
We demonstrate using parameter-free operations in ViT~\citep{dosovitskiy2020image} in a novel way. We follow the aforementioned architectural modifications. Two vision transformer models ViT-S and PiT-S~\citep{heo2021rethinking} are trained on ImageNet with three different training settings: Vanilla~\citep{dosovitskiy2020image}, with CutMix~\citep{cutmix} and DeiT~\citep{deit} settings in \citet{heo2021rethinking}. We similarly apply the depthwise convolution into the self-attention layer, which can be a strong competitor to make a comparison. %
We report the performance of the models in Table~\ref{table:vit}; we report throughput (images/sec) as the speed measure following the ViT papers~\citep{dosovitskiy2020image,heo2021rethinking}, which is measured on 256 batch-size and a single batch-size both. The result demonstrates that ViT and PiT with the max-pool operation have faster throughput without significant accuracy degradation compared with the baselines; the depthwise convolution is a promising alternative, but the throughput is a matter compared with the max-pool operation for both architectures. Interestingly, PiT takes advantage of using the parameter-free operation, which presumably comes from larger features in early layers.

\vspace{-1mm}
\section{Conclusion}
\vspace{-2mm}
In this paper, we rethink parameter-free operations as the building block of learning spatial information to explore a novel way of designing network architecture. We have experimentally studied the applicability of the parameter-free operations in network design and rebuild network architectures, including convolutional neural networks and vision transformers, towards more efficient ones. Extensive results on a large-scale dataset including ImageNet and COCO with diverse network architectures have demonstrated the effectiveness over the existing efficient architectures and the use case of the parameter-free operations as the main building block. We believe our work highlighted a new design paradigm for future research beyond conventional efficient architecture designs.

\section*{Acknowledgements}
We would like to thank NAVER AI Lab members for valuable discussions. We also thank Seong Joon Oh, Sangdoo Yun, and Sungeun Hong for peer-reviews. NAVER Smart Machine Learning (NSML)~\citep{nsml} has been used for experiments.

\section*{Ethics Statement}
This paper studies a general topic in computer vision which is designing an efficient network architecture. Therefore, our work does not be expected to have any potential negative social impact but would contribute to the computer vision field by providing pretrained models.

\section*{Reproducibility Statement}
We provide detailed information of all the experiments in the paper. Furthermore, the details of our models with specific training hyper-paramaters are clearly announced for those who would like to design or train our proposed models. 
\bibliography{egbib}

\begin{thebibliography}{67}
\providecommand{\natexlab}[1]{#1}
\providecommand{\url}[1]{\texttt{#1}}
\expandafter\ifx\csname urlstyle\endcsname\relax
  \providecommand{\doi}[1]{doi: #1}\else
  \providecommand{\doi}{doi: \begingroup \urlstyle{rm}\Url}\fi

\bibitem[Bello et~al.(2021)Bello, Fedus, Du, Cubuk, Srinivas, Lin, Shlens, and
  Zoph]{bello2021revisiting}
Irwan Bello, William Fedus, Xianzhi Du, Ekin~D Cubuk, Aravind Srinivas,
  Tsung-Yi Lin, Jonathon Shlens, and Barret Zoph.
\newblock Revisiting resnets: Improved training and scaling strategies.
\newblock \emph{arXiv preprint arXiv:2103.07579}, 2021.

\bibitem[Chen et~al.(2019{\natexlab{a}})Chen, Xie, Zhang, and Pu]{chen2019all}
Weijie Chen, Di~Xie, Yuan Zhang, and Shiliang Pu.
\newblock All you need is a few shifts: Designing efficient convolutional
  neural networks for image classification.
\newblock In \emph{CVPR}, 2019{\natexlab{a}}.

\bibitem[Chen et~al.(2019{\natexlab{b}})Chen, Xie, Wu, and
  Tian]{chen2019progressive}
Xin Chen, Lingxi Xie, Jun Wu, and Qi~Tian.
\newblock Progressive differentiable architecture search: Bridging the depth
  gap between search and evaluation.
\newblock In \emph{ICCV}, 2019{\natexlab{b}}.

\bibitem[Chu et~al.(2020)Chu, Zhou, Zhang, and Li]{chu2020fair}
Xiangxiang Chu, Tianbao Zhou, Bo~Zhang, and Jixiang Li.
\newblock Fair darts: Eliminating unfair advantages in differentiable
  architecture search.
\newblock In \emph{ECCV}, 2020.

\bibitem[Cubuk et~al.(2019)Cubuk, Zoph, Shlens, and Le]{cubuk2019randaugment}
Ekin~D Cubuk, Barret Zoph, Jonathon Shlens, and Quoc~V Le.
\newblock Randaugment: Practical data augmentation with no separate search.
\newblock \emph{arXiv preprint arXiv:1909.13719}, 2019.

\bibitem[Dai et~al.(2017)Dai, Qi, Xiong, Li, Zhang, Hu, and
  Wei]{dai2017deformable}
Jifeng Dai, Haozhi Qi, Yuwen Xiong, Yi~Li, Guodong Zhang, Han Hu, and Yichen
  Wei.
\newblock Deformable convolutional networks.
\newblock In \emph{ICCV}, 2017.

\bibitem[Dosovitskiy et~al.(2021)Dosovitskiy, Beyer, Kolesnikov, Weissenborn,
  Zhai, Unterthiner, Dehghani, Minderer, Heigold, Gelly,
  et~al.]{dosovitskiy2020image}
Alexey Dosovitskiy, Lucas Beyer, Alexander Kolesnikov, Dirk Weissenborn,
  Xiaohua Zhai, Thomas Unterthiner, Mostafa Dehghani, Matthias Minderer, Georg
  Heigold, Sylvain Gelly, et~al.
\newblock An image is worth 16x16 words: Transformers for image recognition at
  scale.
\newblock In \emph{ICLR}, 2021.

\bibitem[Gibson et~al.(2020)Gibson, Cano, Turner, Crowley, O’Boyle, and
  Storkey]{gibson2020optimizing}
Perry Gibson, Jos{\'e} Cano, Jack Turner, Elliot~J Crowley, Michael O’Boyle,
  and Amos Storkey.
\newblock Optimizing grouped convolutions on edge devices.
\newblock In \emph{ASAP}, 2020.

\bibitem[Goodfellow et~al.(2013)Goodfellow, Warde-Farley, Mirza, Courville, and
  Bengio]{goodfellow2013maxout}
Ian Goodfellow, David Warde-Farley, Mehdi Mirza, Aaron Courville, and Yoshua
  Bengio.
\newblock Maxout networks.
\newblock In \emph{ICML}, 2013.

\bibitem[Goyal et~al.(2017)Goyal, Doll{\'a}r, Girshick, Noordhuis, Wesolowski,
  Kyrola, Tulloch, Jia, and He]{goyal2017accurate}
Priya Goyal, Piotr Doll{\'a}r, Ross Girshick, Pieter Noordhuis, Lukasz
  Wesolowski, Aapo Kyrola, Andrew Tulloch, Yangqing Jia, and Kaiming He.
\newblock Accurate, large minibatch sgd: Training imagenet in 1 hour.
\newblock \emph{arXiv preprint arXiv:1706.02677}, 2017.

\bibitem[Greff et~al.(2017)Greff, Srivastava, and
  Schmidhuber]{greff2016highway}
Klaus Greff, Rupesh~K Srivastava, and J{\"u}rgen Schmidhuber.
\newblock Highway and residual networks learn unrolled iterative estimation.
\newblock In \emph{ICLR}, 2017.

\bibitem[Han et~al.(2021)Han, Yun, Heo, and Yoo]{han2021rethinking}
Dongyoon Han, Sangdoo Yun, Byeongho Heo, and YoungJoon Yoo.
\newblock Rethinking channel dimensions for efficient model design.
\newblock In \emph{CVPR}, 2021.

\bibitem[Han et~al.(2020)Han, Wang, Tian, Guo, Xu, and Xu]{han2020ghostnet}
Kai Han, Yunhe Wang, Qi~Tian, Jianyuan Guo, Chunjing Xu, and Chang Xu.
\newblock Ghostnet: More features from cheap operations.
\newblock In \emph{CVPR}, 2020.

\bibitem[He et~al.(2016{\natexlab{a}})He, Zhang, Ren, and Sun]{preresnet}
Kaiming He, Xiangyu Zhang, Shaoqing Ren, and Jian Sun.
\newblock Identity mappings in deep residual networks.
\newblock In \emph{ECCV}, 2016{\natexlab{a}}.

\bibitem[He et~al.(2016{\natexlab{b}})He, Zhang, Ren, and Sun]{resnet}
Kaiming He, Xiangyu Zhang, Shaoqing Ren, and Jian Sun.
\newblock Deep residual learning for image recognition.
\newblock In \emph{CVPR}, 2016{\natexlab{b}}.

\bibitem[He et~al.(2019)He, Liu, Zhong, and Ma]{he2019addressnet}
Yihui He, Xianggen Liu, Huasong Zhong, and Yuchun Ma.
\newblock Addressnet: Shift-based primitives for efficient convolutional neural
  networks.
\newblock In \emph{WACV}, 2019.

\bibitem[Hendrycks \& Dietterich(2019)Hendrycks and
  Dietterich]{hendrycks2019benchmarking}
Dan Hendrycks and Thomas Dietterich.
\newblock Benchmarking neural network robustness to common corruptions and
  perturbations.
\newblock In \emph{ICLR}, 2019.

\bibitem[Hendrycks et~al.(2021)Hendrycks, Zhao, Basart, Steinhardt, and
  Song]{hendrycks2021natural}
Dan Hendrycks, Kevin Zhao, Steven Basart, Jacob Steinhardt, and Dawn Song.
\newblock Natural adversarial examples.
\newblock In \emph{CVPR}, 2021.

\bibitem[Heo et~al.(2021{\natexlab{a}})Heo, Chun, Oh, Han, Yun, Kim, Uh, and
  Ha]{heo2020adamp}
Byeongho Heo, Sanghyuk Chun, Seong~Joon Oh, Dongyoon Han, Sangdoo Yun, Gyuwan
  Kim, Youngjung Uh, and Jung-Woo Ha.
\newblock Adamp: Slowing down the slowdown for momentum optimizers on
  scale-invariant weights.
\newblock In \emph{ICLR}, 2021{\natexlab{a}}.

\bibitem[Heo et~al.(2021{\natexlab{b}})Heo, Yun, Han, Chun, Choe, and
  Oh]{heo2021rethinking}
Byeongho Heo, Sangdoo Yun, Dongyoon Han, Sanghyuk Chun, Junsuk Choe, and
  Seong~Joon Oh.
\newblock Rethinking spatial dimensions of vision transformers.
\newblock In \emph{ICCV}, 2021{\natexlab{b}}.

\bibitem[Hermans et~al.(2017)Hermans, Beyer, and Leibe]{randomerasing}
Alexander Hermans, Lucas Beyer, and Bastian Leibe.
\newblock In defense of the triplet loss for person re-identification.
\newblock \emph{arXiv preprint arXiv:1703.07737}, 2017.

\bibitem[Howard et~al.(2019)Howard, Sandler, Chu, Chen, Chen, Tan, Wang, Zhu,
  Pang, Vasudevan, et~al.]{mobilenetv3}
Andrew Howard, Mark Sandler, Grace Chu, Liang-Chieh Chen, Bo~Chen, Mingxing
  Tan, Weijun Wang, Yukun Zhu, Ruoming Pang, Vijay Vasudevan, et~al.
\newblock Searching for mobilenetv3.
\newblock In \emph{ICCV}, 2019.

\bibitem[Howard et~al.(2017)Howard, Zhu, Chen, Kalenichenko, Wang, Weyand,
  Andreetto, and Adam]{mobilenetv1}
Andrew~G Howard, Menglong Zhu, Bo~Chen, Dmitry Kalenichenko, Weijun Wang,
  Tobias Weyand, Marco Andreetto, and Hartwig Adam.
\newblock Mobilenets: Efficient convolutional neural networks for mobile vision
  applications.
\newblock \emph{arXiv preprint arXiv:1704.04861}, 2017.

\bibitem[Hu et~al.(2017)Hu, Shen, and Sun]{SENet}
Jie Hu, Li~Shen, and Gang Sun.
\newblock Squeeze-and-excitation networks.
\newblock In \emph{arXiv:1709.01507}, 2017.

\bibitem[Huang et~al.(2016)Huang, Sun, Liu, Sedra, and
  Weinberger]{stochasticdepth}
Gao Huang, Yu~Sun, Zhuang Liu, Daniel Sedra, and Kilian Weinberger.
\newblock Deep networks with stochastic depth.
\newblock In \emph{ECCV}, 2016.

\bibitem[Huang et~al.(2017)Huang, Liu, and Weinberger]{densenet}
Gao Huang, Zhuang Liu, and Kilian~Q Weinberger.
\newblock Densely connected convolutional networks.
\newblock In \emph{CVPR}, 2017.

\bibitem[Huang \& Wang(2018)Huang and Wang]{huang2018data}
Zehao Huang and Naiyan Wang.
\newblock Data-driven sparse structure selection for deep neural networks.
\newblock In \emph{ECCV}, 2018.

\bibitem[Ioffe \& Szegedy(2015)Ioffe and Szegedy]{BN}
Sergey Ioffe and Christian Szegedy.
\newblock Batch normalization: Accelerating deep network training by reducing
  internal covariate shift.
\newblock In \emph{ICML}, 2015.

\bibitem[Jastrzebski et~al.(2018)Jastrzebski, Arpit, Ballas, Verma, and
  Bengio]{jastrzkebski2017residual}
Stanis{\l}aw Jastrzebski, Devansh Arpit, Nicolas Ballas, Tong Verma, Vikas
  andc~Che, and Yoshua Bengio.
\newblock Residual connections encourage iterative inference.
\newblock In \emph{ICLR}, 2018.

\bibitem[Kim et~al.(2018)Kim, Kim, Seo, Kim, Park, Park, Jo, Kim, Yang, Kim,
  et~al.]{nsml}
Hanjoo Kim, Minkyu Kim, Dongjoo Seo, Jinwoong Kim, Heungseok Park, Soeun Park,
  Hyunwoo Jo, KyungHyun Kim, Youngil Yang, Youngkwan Kim, et~al.
\newblock {NSML}: Meet the {MLaaS} platform with a real-world case study.
\newblock \emph{arXiv preprint arXiv:1810.09957}, 2018.

\bibitem[Krizhevsky(2009)]{cifar}
A.~Krizhevsky.
\newblock Learning multiple layers of features from tiny images.
\newblock In \emph{Tech Report}, 2009.

\bibitem[Liang et~al.(2019)Liang, Zhang, Sun, He, Huang, Zhuang, and
  Li]{liang2019darts+}
Hanwen Liang, Shifeng Zhang, Jiacheng Sun, Xingqiu He, Weiran Huang, Kechen
  Zhuang, and Zhenguo Li.
\newblock Darts+: Improved differentiable architecture search with early
  stopping.
\newblock \emph{arXiv preprint arXiv:1909.06035}, 2019.

\bibitem[Lin et~al.(2014)Lin, Maire, Belongie, Hays, Perona, Ramanan,
  Doll{\'a}r, and Zitnick]{coco2017}
Tsung-Yi Lin, Michael Maire, Serge Belongie, James Hays, Pietro Perona, Deva
  Ramanan, Piotr Doll{\'a}r, and C~Lawrence Zitnick.
\newblock Microsoft coco: Common objects in context.
\newblock In \emph{ECCV}, 2014.

\bibitem[Lin et~al.(2017)Lin, Doll{\'a}r, Girshick, He, Hariharan, and
  Belongie]{fpn}
Tsung-Yi Lin, Piotr Doll{\'a}r, Ross Girshick, Kaiming He, Bharath Hariharan,
  and Serge Belongie.
\newblock Feature pyramid networks for object detection.
\newblock In \emph{ICCV}, 2017.

\bibitem[Liu et~al.(2019)Liu, Simonyan, and Yang]{darts}
Hanxiao Liu, Karen Simonyan, and Yiming Yang.
\newblock Darts: Differentiable architecture search.
\newblock In \emph{ICLR}, 2019.

\bibitem[Loshchilov \& Hutter(2017{\natexlab{a}})Loshchilov and
  Hutter]{loshchilov2016sgdr}
Ilya Loshchilov and Frank Hutter.
\newblock Sgdr: Stochastic gradient descent with warm restarts.
\newblock In \emph{ICLR}, 2017{\natexlab{a}}.

\bibitem[Loshchilov \& Hutter(2017{\natexlab{b}})Loshchilov and
  Hutter]{loshchilov2017decoupled}
Ilya Loshchilov and Frank Hutter.
\newblock Decoupled weight decay regularization.
\newblock \emph{arXiv preprint arXiv:1711.05101}, 2017{\natexlab{b}}.

\bibitem[Lu et~al.(2021)Lu, Zhang, and Wang]{lu2021optimizing}
Gangzhao Lu, Weizhe Zhang, and Zheng Wang.
\newblock Optimizing depthwise separable convolution operations on gpus.
\newblock \emph{IEEE Transactions on Parallel and Distributed Systems}, 2021.

\bibitem[Luo et~al.(2017)Luo, Wu, and Lin]{luo2017thinet}
Jian-Hao Luo, Jianxin Wu, and Weiyao Lin.
\newblock Thinet: A filter level pruning method for deep neural network
  compression.
\newblock In \emph{ICCV}, 2017.

\bibitem[Luo et~al.(2016)Luo, Li, Urtasun, and Zemel]{luo2016understanding}
Wenjie Luo, Yujia Li, Raquel Urtasun, and Richard Zemel.
\newblock Understanding the effective receptive field in deep convolutional
  neural networks.
\newblock In \emph{NeurIPS}, 2016.

\bibitem[Nair \& Hinton(2010)Nair and Hinton]{relu}
Vinod Nair and Geoffrey~E Hinton.
\newblock Rectified linear units improve restricted boltzmann machines.
\newblock In \emph{ICML}, 2010.

\bibitem[Nesterov(1983)]{nag}
Y.~E. Nesterov.
\newblock A method for solving the convex programming problem with convergence
  rate $\mathcal{O}(1/k^2)$.
\newblock \emph{Dokl. Akad. Nauk SSSR}, 269:\penalty0 543--547, 1983.

\bibitem[Qiu et~al.(2021)Qiu, Chen, Liu, Zhang, and Zeng]{qiu2021slimconv}
Jiaxiong Qiu, Cai Chen, Shuaicheng Liu, Heng-Yu Zhang, and Bing Zeng.
\newblock Slimconv: Reducing channel redundancy in convolutional neural
  networks by features recombining.
\newblock \emph{IEEE Transactions on Image Processing}, 30:\penalty0
  6434--6445, 2021.

\bibitem[Ren et~al.(2015)Ren, He, Girshick, and Sun]{fasterrcnn}
Shaoqing Ren, Kaiming He, Ross Girshick, and Jian Sun.
\newblock Faster r-cnn: Towards real-time object detection with region proposal
  networks.
\newblock In \emph{NeurIPS}, 2015.

\bibitem[Russakovsky et~al.(2015)Russakovsky, Deng, Su, Krause, Satheesh, Ma,
  Huang, Karpathy, Khosla, Bernstein, Berg, and Fei-Fei]{ImageNet}
Olga Russakovsky, Jia Deng, Hao Su, Jonathan Krause, Sanjeev Satheesh, Sean Ma,
  Zhiheng Huang, Andrej Karpathy, Aditya Khosla, Michael Bernstein,
  Alexander~C. Berg, and Li~Fei-Fei.
\newblock Imagenet large scale visual recognition challenge.
\newblock \emph{International Journal of Computer Vision}, 115\penalty0
  (3):\penalty0 211--252, 2015.

\bibitem[Sandler et~al.(2018)Sandler, Howard, Zhu, Zhmoginov, and
  Chen]{mobilenetv2}
Mark Sandler, Andrew Howard, Menglong Zhu, Andrey Zhmoginov, and Liang-Chieh
  Chen.
\newblock Mobilenetv2: Inverted residuals and linear bottlenecks.
\newblock In \emph{CVPR}, 2018.

\bibitem[Selvaraju et~al.(2017)Selvaraju, Cogswell, Das, Vedantam, Parikh, and
  Batra]{selvaraju2017grad}
Ramprasaath~R Selvaraju, Michael Cogswell, Abhishek Das, Ramakrishna Vedantam,
  Devi Parikh, and Dhruv Batra.
\newblock Grad-cam: Visual explanations from deep networks via gradient-based
  localization.
\newblock In \emph{ICCV}, 2017.

\bibitem[Seong et~al.(2018)Seong, Lee, Kee, Han, and Kim]{seong2018towards}
Sihyeon Seong, Yegang Lee, Youngwook Kee, Dongyoon Han, and Junmo Kim.
\newblock Towards flatter loss surface via nonmonotonic learning rate
  scheduling.
\newblock In \emph{UAI}, 2018.

\bibitem[Simonyan \& Zisserman(2015)Simonyan and Zisserman]{vgg}
Karen Simonyan and Andrew Zisserman.
\newblock Very deep convolutional networks for large-scale image recognition.
\newblock In \emph{ICLR}, 2015.

\bibitem[Szegedy et~al.(2015)Szegedy, Liu, Jia, Sermanet, Reed, Anguelov,
  Erhan, Vanhoucke, and Rabinovich]{GoogleNet}
Christian Szegedy, Wei Liu, Yangqing Jia, Pierre Sermanet, Scott Reed, Dragomir
  Anguelov, Dumitru Erhan, Vincent Vanhoucke, and Andrew Rabinovich.
\newblock Going deeper with convolutions.
\newblock In \emph{CVPR}, 2015.

\bibitem[Szegedy et~al.(2016)Szegedy, Vanhoucke, Ioffe, Shlens, and
  Wojna]{Inceptionv3}
Christian Szegedy, Vincent Vanhoucke, Sergey Ioffe, Jon Shlens, and Zbigniew
  Wojna.
\newblock Rethinking the inception architecture for computer vision.
\newblock In \emph{CVPR}, 2016.

\bibitem[Tan \& Le(2019)Tan and Le]{efficientnet}
Mingxing Tan and Quoc~V Le.
\newblock Efficientnet: Rethinking model scaling for convolutional neural
  networks.
\newblock \emph{arXiv preprint arXiv:1905.11946}, 2019.

\bibitem[Tan \& Le(2021)Tan and Le]{tan2021efficientnetv2}
Mingxing Tan and Quoc~V Le.
\newblock Efficientnetv2: Smaller models and faster training.
\newblock In \emph{ICML}, 2021.

\bibitem[Tarvainen \& Valpola(2017)Tarvainen and Valpola]{tarvainen2017mean}
Antti Tarvainen and Harri Valpola.
\newblock Mean teachers are better role models: Weight-averaged consistency
  targets improve semi-supervised deep learning results.
\newblock In \emph{NeurIPS}, 2017.

\bibitem[Touvron et~al.(2021)Touvron, Cord, Douze, Massa, Sablayrolles, and
  J{\'e}gou]{deit}
Hugo Touvron, Matthieu Cord, Matthijs Douze, Francisco Massa, Alexandre
  Sablayrolles, and Herv{\'e} J{\'e}gou.
\newblock Training data-efficient image transformers \& distillation through
  attention.
\newblock In \emph{ICML}, 2021.

\bibitem[Vaswani et~al.(2017)Vaswani, Shazeer, Parmar, Uszkoreit, Jones, Gomez,
  Kaiser, and Polosukhin]{transformer}
Ashish Vaswani, Noam Shazeer, Niki Parmar, Jakob Uszkoreit, Llion Jones,
  Aidan~N Gomez, {\L}ukasz Kaiser, and Illia Polosukhin.
\newblock Attention is all you need.
\newblock In \emph{Advances in neural information processing systems}, pp.\
  5998--6008, 2017.

\bibitem[Veit et~al.(2016)Veit, Wilber, and Belongie]{veit2016residual}
Andreas Veit, Michael~J Wilber, and Serge Belongie.
\newblock Residual networks behave like ensembles of relatively shallow
  networks.
\newblock In \emph{NeurIPS}, 2016.

\bibitem[Wang et~al.(2018)Wang, Xu, Xu, Xu, and Tao]{wang2018versatile}
Yunhe Wang, Chang Xu, Chunjing Xu, Chao Xu, and Dacheng Tao.
\newblock Learning versatile filters for efficient convolutional neural
  networks.
\newblock \emph{NeurIPS}, 2018.

\bibitem[Wu et~al.(2018)Wu, Wan, Yue, Jin, Zhao, Golmant, Gholaminejad,
  Gonzalez, and Keutzer]{wu2018shift}
Bichen Wu, Alvin Wan, Xiangyu Yue, Peter Jin, Sicheng Zhao, Noah Golmant, Amir
  Gholaminejad, Joseph Gonzalez, and Kurt Keutzer.
\newblock Shift: A zero flop, zero parameter alternative to spatial
  convolutions.
\newblock In \emph{CVPR}, 2018.

\bibitem[Xie et~al.(2017)Xie, Girshick, Doll{\'a}r, Tu, and He]{resnext}
Saining Xie, Ross Girshick, Piotr Doll{\'a}r, Zhuowen Tu, and Kaiming He.
\newblock Aggregated residual transformations for deep neural networks.
\newblock In \emph{CVPR}, 2017.

\bibitem[Yu et~al.(2019)Yu, Yang, Xu, Yang, and Huang]{yu2019slimmable}
Jiahui Yu, Linjie Yang, Ning Xu, Jianchao Yang, and Thomas Huang.
\newblock Slimmable neural networks.
\newblock In \emph{ICLR}, 2019.

\bibitem[Yun et~al.(2019)Yun, Han, Oh, Chun, Choe, and Yoo]{cutmix}
Sangdoo Yun, Dongyoon Han, Seong~Joon Oh, Sanghyuk Chun, Junsuk Choe, and
  Youngjoon Yoo.
\newblock Cutmix: Regularization strategy to train strong classifiers with
  localizable features.
\newblock In \emph{ICCV}, 2019.

\bibitem[Zagoruyko \& Komodakis(2016)Zagoruyko and Komodakis]{wideresnet}
Sergey Zagoruyko and Nikos Komodakis.
\newblock Wide residual networks.
\newblock In \emph{BMVC}, 2016.

\bibitem[Zhai et~al.(2021)Zhai, Kolesnikov, Houlsby, and
  Beyer]{zhai2021scaling}
Xiaohua Zhai, Alexander Kolesnikov, Neil Houlsby, and Lucas Beyer.
\newblock Scaling vision transformers.
\newblock \emph{arXiv preprint arXiv:2106.04560}, 2021.

\bibitem[Zhang et~al.(2018)Zhang, Zhou, Lin, and Sun]{shufflenetv1}
Xiangyu Zhang, Xinyu Zhou, Mengxiao Lin, and Jian Sun.
\newblock Shufflenet: An extremely efficient convolutional neural network for
  mobile devices.
\newblock In \emph{CVPR}, 2018.

\bibitem[Zhou \& Feng(2017)Zhou and Feng]{zhou2017}
Pan Zhou and Jiashi Feng.
\newblock The landscape of deep learning algorithms.
\newblock \emph{arXiv preprint arXiv:1705.07038}, 2017.

\bibitem[Zhu et~al.(2019)Zhu, Hu, Lin, and Dai]{zhu2019deformable}
Xizhou Zhu, Han Hu, Stephen Lin, and Jifeng Dai.
\newblock Deformable convnets v2: More deformable, better results.
\newblock In \emph{CVPR}, 2019.

\end{thebibliography}
\bibliographystyle{iclr2021_conference}

\clearpage
\appendix
\section*{Appendix}
\numberwithin{equation}{section}
\numberwithin{figure}{section}
\numberwithin{table}{section}

\section{Details of Efficient Architectures}
\label{supp_sec:architectures}
\begin{figure*}[h]
\small
\centering
\hspace{-2mm}
\begin{subfigure}[ht!]{0.49\linewidth}
\includegraphics[page=1, trim = 0mm 1mm 80mm 1mm, clip, width=1.0\linewidth]{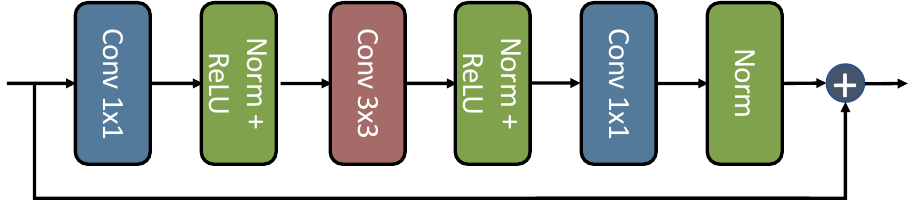}
\caption{Bottleneck~\citep{resnet}}
\label{fig_supps:bot}
\end{subfigure} \
\begin{subfigure}[ht!]{0.49\linewidth}
\includegraphics[page=1, trim = 92mm 1mm 0mm 1mm, clip, width=1.0\linewidth]{Figure/source/blocks/blocks2.pdf}  
\caption{Transformer~\citep{dosovitskiy2020image}}
\label{fig_supps:transformer}
\end{subfigure} \\
\begin{subfigure}[ht!]{0.43\linewidth}
\includegraphics[page=1, trim = 0mm 1mm 80mm 1mm, clip, width=1.0\linewidth]{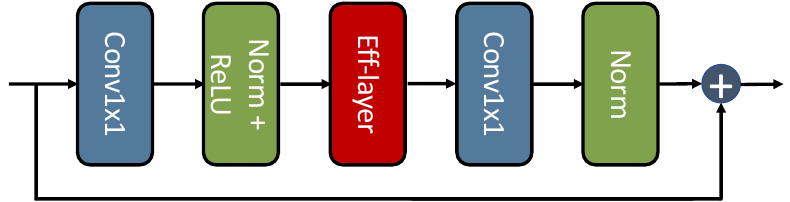}
\caption{Efficient Bottleneck}
\label{fig_supps:eff_bot}
\end{subfigure} \quad\quad\quad
\begin{subfigure}[ht!]{0.49\linewidth}
\includegraphics[page=1, trim = 80mm 1mm 0mm 1mm, clip, width=1.0\linewidth]{Figure/source/blocks/blocks1.pdf}  
\caption{Efficient Transformer}
\label{fig_supps:eff_transformer}
\end{subfigure}
\vspace{-2mm}
\caption{\small {\bf Schematic illustration of the efficient building blocks.} We visualize (a) the regular bottleneck in ResNets~\citep{resnet}; (b) the regular transformer in ViTs~\citep{dosovitskiy2020image}; (c) our efficient bottleneck; (d) our efficient transformer; the eff-layers in (c) and (d) denote the parameter-free operations.} 
\label{fig_supps:blocks}
\vspace{-3mm}
\end{figure*}

We elaborate on the efficient building blocks used in the experiments above. Fig.\ref{fig_supps:blocks} shows the schematic illustration of the proposed blocks compared with the original ones. We observe that the modification is simple and readily be applied to any network architecture.

\paragraph{Efficient Bottleneck.}
We replace the triplet of 3$\times$3 convolution, BN~\citep{BN}, and ReLU~\citep{relu} in the regular bottleneck with a single spatial parameter-free operation (denoted as eff-layer in Fig.\ref{fig_supps:eff_bot}). The fastest architecture in Table~\ref{table:resnet_study} and Table~\ref{table:imagenet_comparison} fully replace the regular bottlenecks, including the downsampling blocks with the efficient bottlenecks; therefore, the parameter-free operation (here the max-pool operation) plays a role of spatially aggregating the features by reducing the resolution. The {\it hybrid} architecture has an almost identical design regime with the most efficient architecture ($i.e.,$ {\bf E}$\rightarrow${\bf E}$\rightarrow${\bf E}$\rightarrow${\bf E} in Table~\ref{table:resnet_study}) except for the downsampling blocks. We remain each downsampling block with the regular bottleneck blocks and involve an efficient parameter-free operation before the spatial operation (we assign the avg-pool as the parameter-free operation). We apply the identical bottleneck configuration of the {\it hybrid} architecture to the diverse deep convolutional neural networks (CNNs) in Table~\ref{table:imagenet_bigger_models}. Note that we do not modify the depth and the width of network architecture, the stem of CNNs that is the first set of layers before the first bottleneck, and the output layer having a fully connected layer with 1000-d output dimension. Furthermore, we do not modify specific architectural elements such as the SE-block of ResNet50-SE~\citep{SENet} inside each regular bottleneck but use it in each efficient bottleneck at the same positions.

\paragraph{Efficient Transformer.}
Designing the efficient vision transformer (ViT) architecture is done by replacing half of the self-attention layers with the spatial parameter-free operation (denoted as eff-layer in Fig.\ref{fig_supps:eff_transformer}). In other words, the original transformer block and the efficient transformer block are used alternately. We use the 3${\times}$3 max-pool operation for eff-layer in the efficient transformer block and compare with the variant using the 5${\times}$5 depth-wise convolution for eff-layer. ReLU is added after eff-layer and the first linear layer of the efficient transformer block to give additional non-linearity. For Pooling-based Vision Transformer (PiT)~\citep{heo2020adamp} and ViT~\citep{dosovitskiy2020image}, we do not modify the architectural elements including 1) the patch size; 2) the stem that patchifies the input for the following transformer; 3) the number of transformers. We only change the classification head position from the classification token to the Global Average Pooling (GAP) at the last layer since the classification token is not compatible with convolutional operations. As reported in \citet{zhai2021scaling}, transformer with GAP shows comparable performance to ViT with the classification token. We also evaluate a more efficient network architecture which is fully equipped with the efficient transformer; it achieves faster speed, but slightly less accurate. We leave further improvements with a parameter-free operation inside the efficient transformer to reach the accuracy of the original transformer as future work.

\section{On Parameter-free Operations}
\label{supp_sec:pool_comparison}
\begin{figure*}[t]
\small
\centering
\begin{subfigure}[ht!]{0.28\linewidth}
\quad\quad\quad\quad\quad{\fontsize{8.0}{9.0}\selectfont SGD}
\end{subfigure} \ 
\begin{subfigure}[ht!]{0.28\linewidth}
\quad\quad\quad\quad\quad{\fontsize{8.0}{9.0}\selectfont AdamW}
\end{subfigure} \ 
\begin{subfigure}[ht!]{0.28\linewidth}
\quad\quad\quad\quad\quad{\fontsize{8.0}{9.0}\selectfont AdamP}
\end{subfigure} \\
\begin{subfigure}[ht!]{0.28\linewidth}
\includegraphics[page=1, trim = 0mm 0mm 0mm 0mm, clip, width=1.0\linewidth]{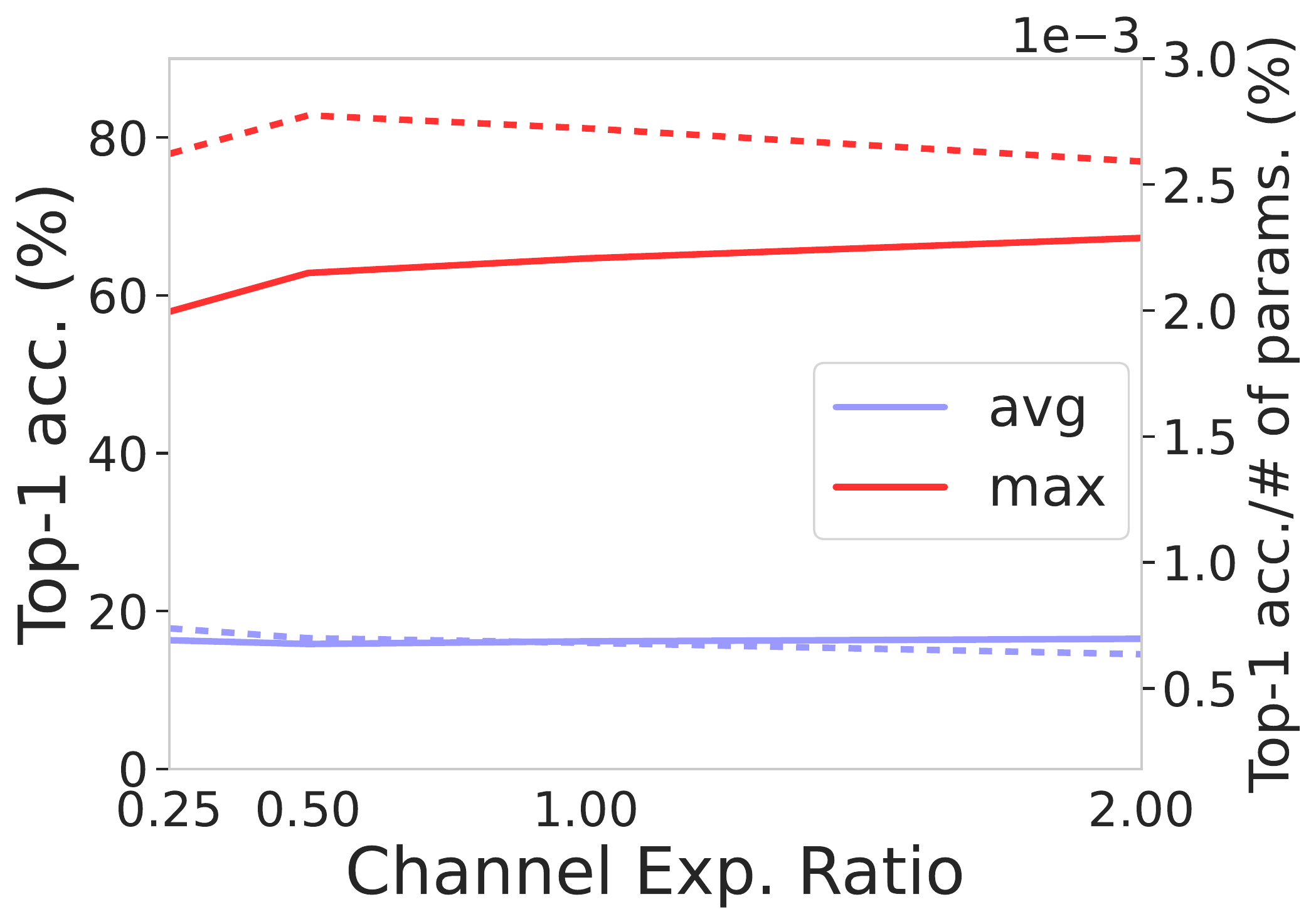}
\end{subfigure} \ \
\begin{subfigure}[ht!]{0.28\linewidth}
\includegraphics[page=1, trim = 0mm 0mm 0mm 0mm, clip, width=1.0\linewidth]{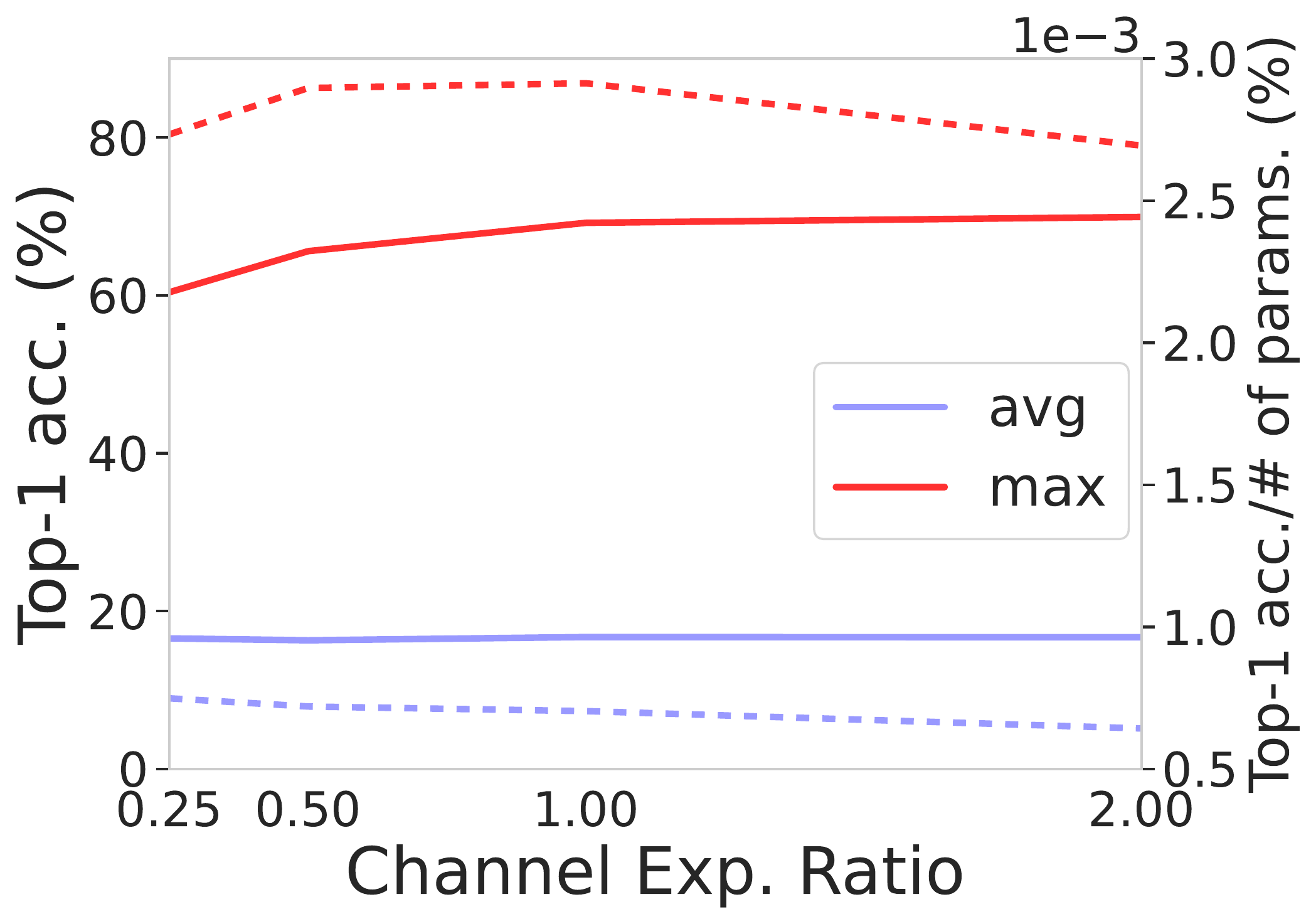}  
\end{subfigure} \ \
\begin{subfigure}[ht!]{0.28\linewidth}
\includegraphics[page=1, trim = 0mm 0mm 0mm 0mm, clip, width=1.0\linewidth]{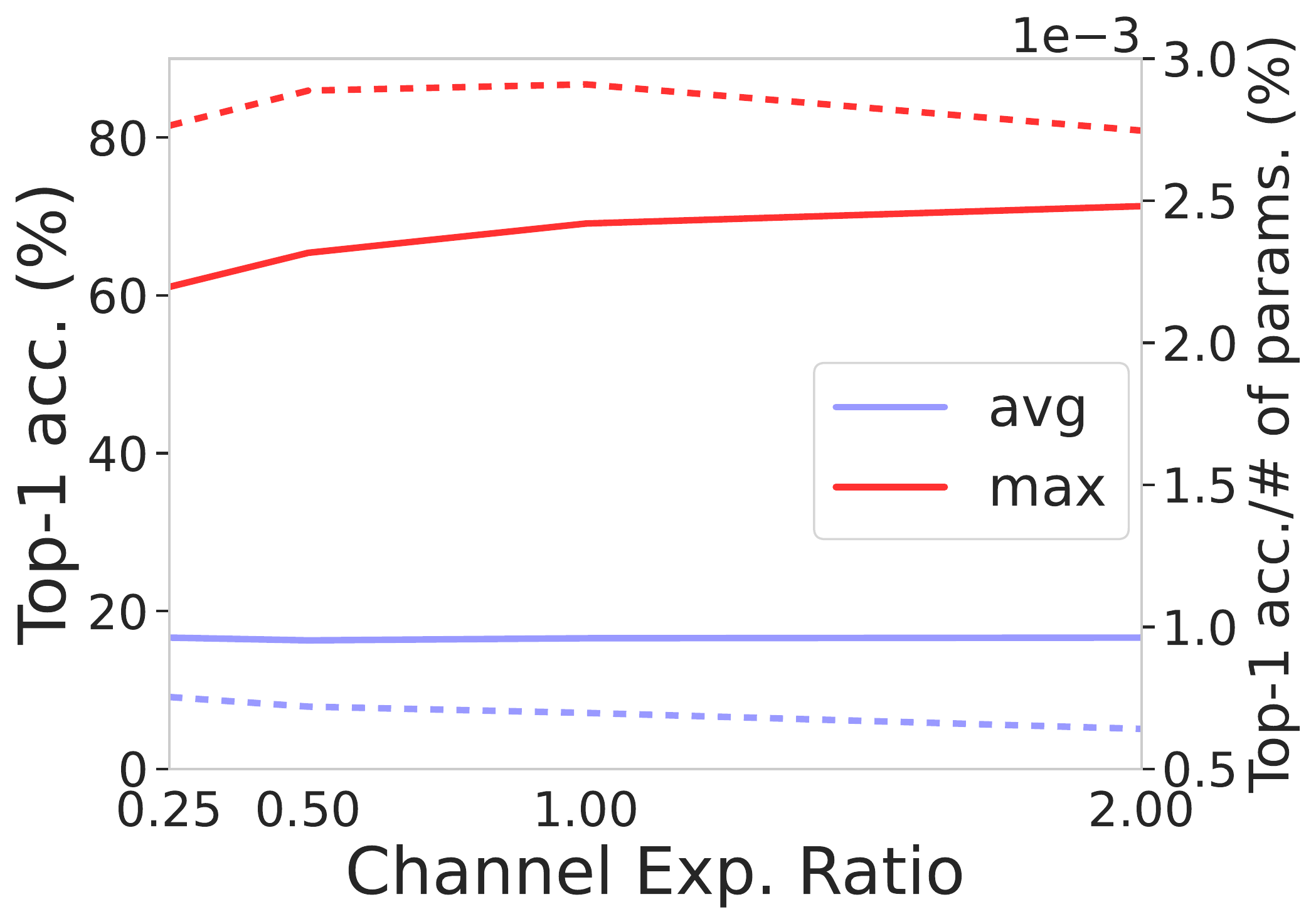}  
\end{subfigure} 
\vspace{-2mm}
\caption{\small {\bf Comparison of the parameter-free operations.} We visualize top-1 accuracy ({\bf solid lines}) with accuracy per \# parameters ({\bf dashed lines}) similar to Fig.\ref{fig:fig_cifar10_expansion_toy} with two parameter-free operations. All the settings are identical to the previous study, so we only plot the channel width of 32. We observe the max-pool operation consistently beat the avg-pool operation in a single bottleneck training. } 
\label{fig:fig_cifar10_avg_vs_max}
\vspace{-3mm}
\end{figure*}

\paragraph{Can We Use Avg-pool as the Spatial Parameter-free Operation?}
We mainly used the max-pool operation as the spatial parameter-free operation that replaces the regular convolutions and the self-attention layer in experiments. The max-pool operation is expected to have higher expressiveness compared with that of the avg-pool operation. Because the avg-pool operation is a conceptually smoothing operation, and the max-pool operation contains a nonlinearity similar to what ReLU has ($i.e.,$ $\text{max}(x)$ and $\text{max}(x, 0)$). Experimentally, we found low expressiveness of the avg-pool operation in the shallow network study. Fig.\ref{fig:fig_cifar10_avg_vs_max} shows the comparison of the max-pool and the avg-pool operations trained in a single bottleneck block. The setting is identical to the previous study in \S\ref{sec:study}, and we only visualize the case of the channel width of 32 because a similar trend with the channel width of 64 was observed. We observe the large accuracy gaps between the two operations. This may be a ground of the search results where few avg-pool operations are chosen in normal cells in the previous NAS experiments in \S\ref{sec_sub:empirical_studies}.

\paragraph{Deformable Avg-pool Operation.}
We here provide an interesting idea of using the avg-pool operation, which shows inferior outcomes than the max-pool operation in the operation expressiveness. We have proposed the deformable max-pool operation involving a few parameters to improve the parameter-free operation's discriminative power significantly. Using the same idea into the avg-pool operation is available; surprisingly, a downside of the low expressiveness of the avg-pool operation has been vanished, as shown in Table~\ref{table:imagenet_deform_avg}. This result exhibits that even a smoothing operation can be employed to learn discriminative features in a deep neural network.  %
\begin{table*}[h]
\small
\centering
\tabcolsep=0.1cm
\caption{\small {\bf ImageNet performance of deformable operations.} We report the model performance trained on ImageNet for the deformable operations. The new operation, dubbed deformable avg-pool, surprisingly reaches the accuracy of the deformable max-pool operation.}
\label{table:imagenet_deform_avg}
\vspace{-2mm}
\subfloat{\begin{tabular}{l|c|c|c|c|c} %
\toprule
Network Architecture & Params. (M) & FLOPs (G) & GPU (ms)  & Top-1 (\%) & Top-5 (\%) \\
\midrule
Ours-R50 (deform\_max) & 18.0 & 2.9 & 10.3 & 78.3 & 94.0 \\
Ours-R50 (deform\_avg) & 18.0 & 2.9 & 10.3 & 78.3 & 93.9 \\
\bottomrule
\end{tabular}}
\vspace{-3mm}
\end{table*}

\section{Detailed Training Recipes}
\label{supp_sec:training_recipes}
We use the following training recipes to maximize the accuracy of the ResNet-based models. We use the cosine learning rate scheduling~\citep{loshchilov2016sgdr} with the initial learning rate of 0.5 using four V100 GPUs with batch size of 512. Exponential moving average~\citep{tarvainen2017mean} over the network weights is used during training. We use the regularization techniques and data augmentations including label smoothing~\citep{Inceptionv3} (0.1), RandAug~\citep{cubuk2019randaugment} (magnitude of 9), Random Erasing~\citep{randomerasing} with pixels (0.2), lowered weight decay (1e-5), and a large training epochs (400 epochs)\footnote{When training for a larger epochs (600 epochs), the top-1 accuracy of Ours-R50 (max) in Table~\ref{table:imagenet_comparison} is improved to 75.5\%, which gets closer to the original ResNet's; furthermore,  Ours-R50 ({\it hybrid}), and Ours-R50 (deform\_max) reach 78.0 and 79.3, respectively.}. We use the code baseline in the renowned repository\footnote{\url{https://github.com/rwightman/pytorch-image-models/}} for our ImageNet training. 
The accuracy of the baseline models could reach the known improved accuracy reported in such a paper~\citep{bello2021revisiting} which presents the training recipes for highly improved models. 

\section{Understanding Max-pool Operation}
This section provides intuitive explanations how parameter-free operations such as the max-pool operation could work as replacing a trainable layers in a network.

\subsection{Connection with Maxout}
Maxout~\citep{goodfellow2013maxout} performs the max operation to a set of the inputs like an activation function with the multiple inputs. Maxout is designed to use after a bunch of linear or convolution layers to ensemble the outputs in a nonlinear manner, which leads to the increase of model capacity. The accuracy improvement in the original paper~\citep{goodfellow2013maxout} results from the claim that Maxout can approximate any operations, namely perform as a universal approximator. From an architectural point of view, the success of Maxout is probably due to the enhanced model capacity by the increased input dimension with the nonlinearity imposed by the max operation.

For the max-pool operation with the efficient bottleneck mainly used throughout the paper, the relationship between the Maxout operation and ours is worth discussing. Maxout outputs through the max operation of the outputs of multiple linear layers computed from a single input. Otherwise, the max-pool operation in the efficient bottleneck performs the max operation with multiple transformed inputs (i.e., transformed pixels spatially adjacent to each point) by the preceding 1${\times}$1 convolution, which acts as a single linear layer for a given channel. However, if we replace the small variations in neighboring pixels with the perturbations in the weights of the linear layer as claimed in \citet{seong2018towards}, then we may interpret the max-pool operation performs like Maxout. %
\begin{figure*}[t]
\fontsize{7.5}{8.5}\selectfont
\centering
\hspace{-3mm}
\begin{subfigure}[ht!]{0.018\linewidth}
\vspace{-13.5mm} \rotatebox{90}{\parbox{2cm}{\red{Input}}} %
\end{subfigure}
\begin{subfigure}[ht!]{0.105\linewidth}
\includegraphics[page=1, trim = 0mm 0mm 0mm 0mm, clip, width=1.0\linewidth]{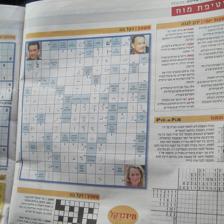}
\end{subfigure} 
\begin{subfigure}[ht!]{0.105\linewidth}
\includegraphics[page=1, trim = 0mm 0mm 0mm 0mm, clip, width=1.0\linewidth]{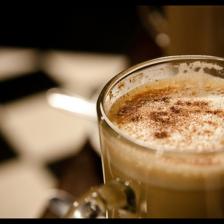}
\end{subfigure}
\begin{subfigure}[ht!]{0.105\linewidth}
\includegraphics[page=1, trim = 0mm 0mm 0mm 0mm, clip, width=1.0\linewidth]{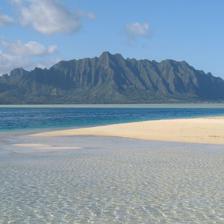}
\end{subfigure}
\begin{subfigure}[ht!]{0.105\linewidth}
\includegraphics[page=1, trim = 0mm 0mm 0mm 0mm, clip, width=1.0\linewidth]{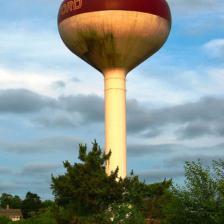}
\end{subfigure}
\begin{subfigure}[ht!]{0.105\linewidth}
\includegraphics[page=1, trim = 0mm 0mm 0mm 0mm, clip, width=1.0\linewidth]{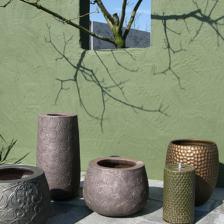}
\end{subfigure} 
\begin{subfigure}[ht!]{0.105\linewidth}
\includegraphics[page=1, trim = 0mm 0mm 0mm 0mm, clip, width=1.0\linewidth]{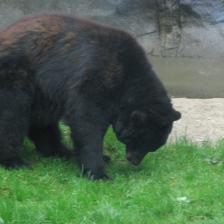}
\end{subfigure}
\begin{subfigure}[ht!]{0.105\linewidth}
\includegraphics[page=1, trim = 0mm 0mm 0mm 0mm, clip, width=1.0\linewidth]{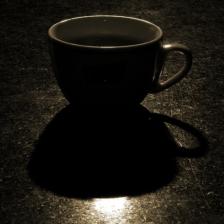}
\end{subfigure}
\begin{subfigure}[ht!]{0.105\linewidth}
\includegraphics[page=1, trim = 0mm 0mm 0mm 0mm, clip, width=1.0\linewidth]{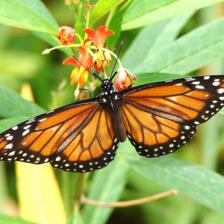}
\end{subfigure}
\begin{subfigure}[ht!]{0.105\linewidth}
\includegraphics[page=1, trim = 0mm 0mm 0mm 0mm, clip, width=1.0\linewidth]{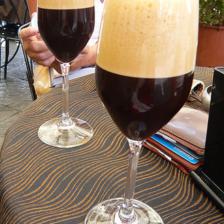}
\end{subfigure} \\
\hspace{-3mm} 
\begin{subfigure}[ht!]{0.018\linewidth}
\vspace{-7.5mm}\rotatebox{90}{\parbox{2cm}{\red{ResNet50}}}%
\end{subfigure}
\begin{subfigure}[ht!]{0.105\linewidth}
\includegraphics[page=1, trim = 0mm 0mm 0mm 0mm, clip, width=1.0\linewidth]{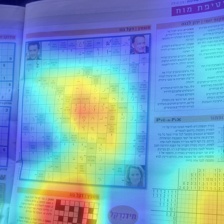}
\end{subfigure} 
\begin{subfigure}[ht!]{0.105\linewidth}
\includegraphics[page=1, trim = 0mm 0mm 0mm 0mm, clip, width=1.0\linewidth]{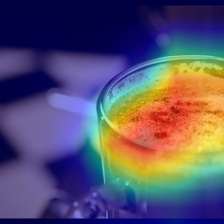}
\end{subfigure}
\begin{subfigure}[ht!]{0.105\linewidth}
\includegraphics[page=1, trim = 0mm 0mm 0mm 0mm, clip, width=1.0\linewidth]{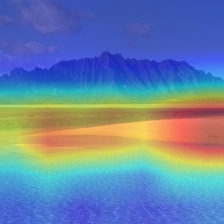}
\end{subfigure}
\begin{subfigure}[ht!]{0.105\linewidth}
\includegraphics[page=1, trim = 0mm 0mm 0mm 0mm, clip, width=1.0\linewidth]{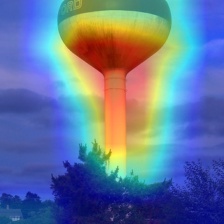}
\end{subfigure}
\begin{subfigure}[ht!]{0.105\linewidth}
\includegraphics[page=1, trim = 0mm 0mm 0mm 0mm, clip, width=1.0\linewidth]{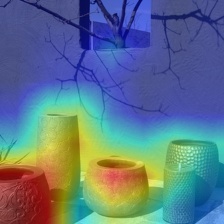}
\end{subfigure} 
\begin{subfigure}[ht!]{0.105\linewidth}
\includegraphics[page=1, trim = 0mm 0mm 0mm 0mm, clip, width=1.0\linewidth]{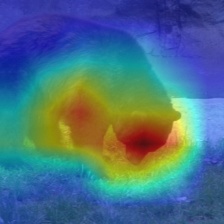}
\end{subfigure}
\begin{subfigure}[ht!]{0.105\linewidth}
\includegraphics[page=1, trim = 0mm 0mm 0mm 0mm, clip, width=1.0\linewidth]{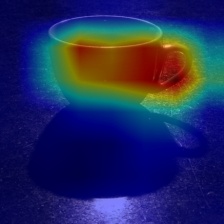}
\end{subfigure}
\begin{subfigure}[ht!]{0.105\linewidth}
\includegraphics[page=1, trim = 0mm 0mm 0mm 0mm, clip, width=1.0\linewidth]{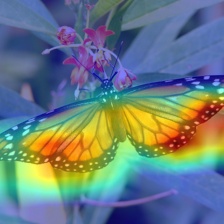}
\end{subfigure}
\begin{subfigure}[ht!]{0.105\linewidth}
\includegraphics[page=1, trim = 0mm 0mm 0mm 0mm, clip, width=1.0\linewidth]{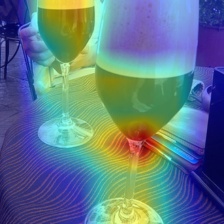}
\end{subfigure} \\
\hspace{-3mm}
\begin{subfigure}[ht!]{0.018\linewidth}
\vspace{-13.5mm} \rotatebox{90}{\parbox{2cm}{\red{Ours}}}
\end{subfigure}
\begin{subfigure}[ht!]{0.105\linewidth}
\includegraphics[page=1, trim = 0mm 0mm 0mm 0mm, clip, width=1.0\linewidth]{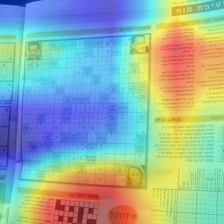}
\end{subfigure} 
\begin{subfigure}[ht!]{0.105\linewidth}
\includegraphics[page=1, trim = 0mm 0mm 0mm 0mm, clip, width=1.0\linewidth]{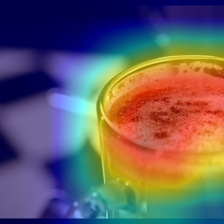}
\end{subfigure}
\begin{subfigure}[ht!]{0.105\linewidth}
\includegraphics[page=1, trim = 0mm 0mm 0mm 0mm, clip, width=1.0\linewidth]{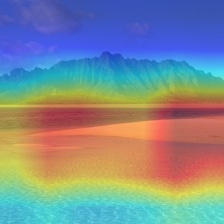}
\end{subfigure}
\begin{subfigure}[ht!]{0.105\linewidth}
\includegraphics[page=1, trim = 0mm 0mm 0mm 0mm, clip, width=1.0\linewidth]{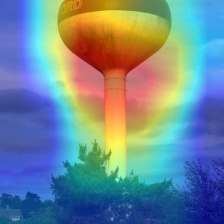}
\end{subfigure}
\begin{subfigure}[ht!]{0.105\linewidth}
\includegraphics[page=1, trim = 0mm 0mm 0mm 0mm, clip, width=1.0\linewidth]{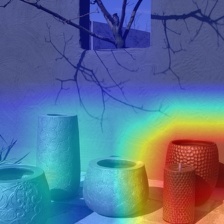}
\end{subfigure} 
\begin{subfigure}[ht!]{0.105\linewidth}
\includegraphics[page=1, trim = 0mm 0mm 0mm 0mm, clip, width=1.0\linewidth]{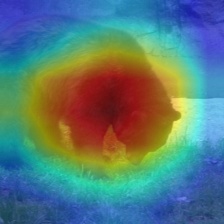}
\end{subfigure}
\begin{subfigure}[ht!]{0.105\linewidth}
\includegraphics[page=1, trim = 0mm 0mm 0mm 0mm, clip, width=1.0\linewidth]{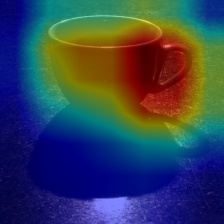}
\end{subfigure}
\begin{subfigure}[ht!]{0.105\linewidth}
\includegraphics[page=1, trim = 0mm 0mm 0mm 0mm, clip, width=1.0\linewidth]{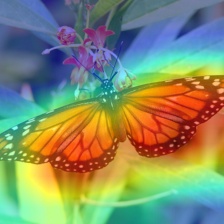}
\end{subfigure}
\begin{subfigure}[ht!]{0.105\linewidth}
\includegraphics[page=1, trim = 0mm 0mm 0mm 0mm, clip, width=1.0\linewidth]{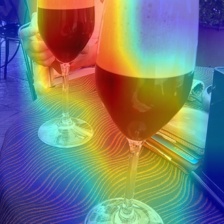}
\end{subfigure} \\
\vspace{-2mm}
\caption{\small \red{{\bf Grad-CAM visualization of the final features.} We visualize the highlighted features from the output of the final stage using Gradient-weighted Class Activation Mapping (Grad-CAM)~\citep{selvaraju2017grad}. The first row shows the original images, which are randomly picked from the ImageNet validation set; the second row shows the visualized features of the ImageNet-pretrained ResNet50; the last row shows the result of our ImageNet-pretrained ResNet50 ({\it hybrid}). Ours show similar outputs compared with ResNet50 but seem to capture the foreground with wider regions.}}
\label{fig:fig_feature_visualization}
\vspace{-4mm}
\end{figure*}

\subsection{Receptive Field Matter}
\begin{figure*}[t]
\fontsize{7.5}{8.5}\selectfont
\centering
\hspace{7mm}
\begin{subfigure}[ht!]{0.105\linewidth}
\quad\red{Input}
\end{subfigure} 
\begin{subfigure}[ht!]{0.105\linewidth}
\red{conv5\_2\_1}
\end{subfigure} 
\begin{subfigure}[ht!]{0.105\linewidth}
\red{conv5\_2\_2}
\end{subfigure} 
\begin{subfigure}[ht!]{0.105\linewidth}
\red{conv5\_2\_3}
\end{subfigure} 
\begin{subfigure}[ht!]{0.105\linewidth}
\red{conv5\_3\_1}
\end{subfigure} 
\begin{subfigure}[ht!]{0.105\linewidth}
\red{conv5\_3\_2}
\end{subfigure} 
\begin{subfigure}[ht!]{0.105\linewidth}
\red{conv5\_3\_3}
\end{subfigure}  \\
\vspace{0.8mm}
\begin{subfigure}[ht!]{0.018\linewidth}
\vspace{-10mm} \rotatebox{90}{\parbox{2cm}{\red{ResNet50}}} 
\end{subfigure}
\begin{subfigure}[ht!]{0.105\linewidth}
\includegraphics[page=1, trim = 0mm 0mm 0mm 0mm, clip, width=1.0\linewidth]{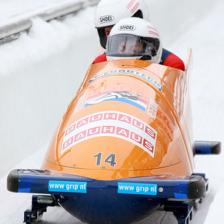}
\end{subfigure} 
\begin{subfigure}[ht!]{0.105\linewidth}
\includegraphics[page=1, trim = 0mm 0mm 0mm 0mm, clip, width=1.0\linewidth]{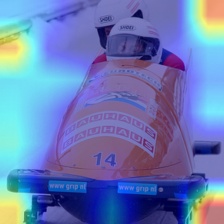}
\end{subfigure} 
\begin{subfigure}[ht!]{0.105\linewidth}
\includegraphics[page=1, trim = 0mm 0mm 0mm 0mm, clip, width=1.0\linewidth]{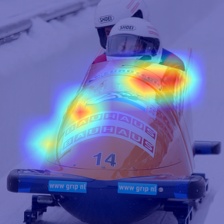}
\end{subfigure}
\begin{subfigure}[ht!]{0.105\linewidth}
\includegraphics[page=1, trim = 0mm 0mm 0mm 0mm, clip, width=1.0\linewidth]{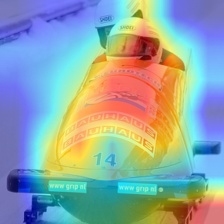}
\end{subfigure}
\begin{subfigure}[ht!]{0.105\linewidth}
\includegraphics[page=1, trim = 0mm 0mm 0mm 0mm, clip, width=1.0\linewidth]{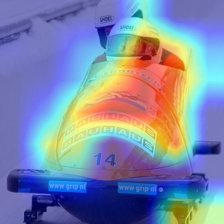}
\end{subfigure} 
\begin{subfigure}[ht!]{0.105\linewidth}
\includegraphics[page=1, trim = 0mm 0mm 0mm 0mm, clip, width=1.0\linewidth]{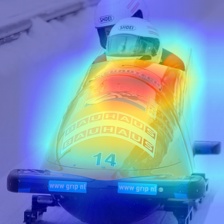}
\end{subfigure} 
\begin{subfigure}[ht!]{0.105\linewidth}
\includegraphics[page=1, trim = 0mm 0mm 0mm 0mm, clip, width=1.0\linewidth]{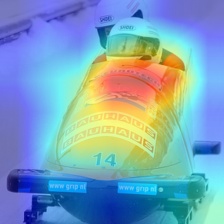}
\end{subfigure} \\
\begin{subfigure}[ht!]{0.018\linewidth}
\vspace{-14mm} \rotatebox{90}{\parbox{2cm}{\red{Ours}}}
\end{subfigure}
\begin{subfigure}[ht!]{0.105\linewidth}
\includegraphics[page=1, trim = 0mm 0mm 0mm 0mm, clip, width=1.0\linewidth]{Figure/source/IN_layerwise_visualization/ILSVRC2012_val_00031459.JPEG}
\end{subfigure} 
\begin{subfigure}[ht!]{0.105\linewidth}
\includegraphics[page=1, trim = 0mm 0mm 0mm 0mm, clip, width=1.0\linewidth]{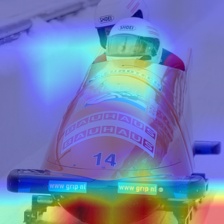}
\end{subfigure} 
\begin{subfigure}[ht!]{0.105\linewidth}
\includegraphics[page=1, trim = 0mm 0mm 0mm 0mm, clip, width=1.0\linewidth]{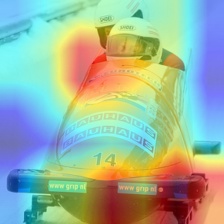}
\end{subfigure}
\begin{subfigure}[ht!]{0.105\linewidth}
\includegraphics[page=1, trim = 0mm 0mm 0mm 0mm, clip, width=1.0\linewidth]{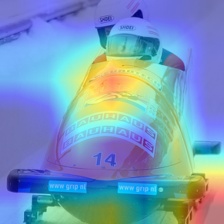}
\end{subfigure}
\begin{subfigure}[ht!]{0.105\linewidth}
\includegraphics[page=1, trim = 0mm 0mm 0mm 0mm, clip, width=1.0\linewidth]{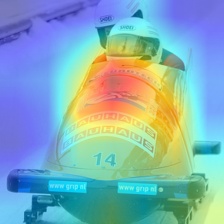}
\end{subfigure}
\begin{subfigure}[ht!]{0.105\linewidth}
\includegraphics[page=1, trim = 0mm 0mm 0mm 0mm, clip, width=1.0\linewidth]{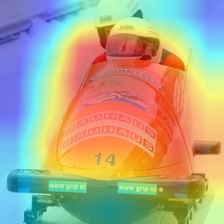}
\end{subfigure} 
\begin{subfigure}[ht!]{0.105\linewidth}
\includegraphics[page=1, trim = 0mm 0mm 0mm 0mm, clip, width=1.0\linewidth]{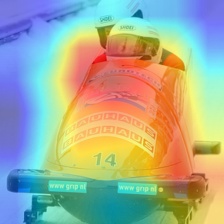}
\end{subfigure} \\ \vspace{1mm}
\begin{subfigure}[ht!]{0.018\linewidth}
\vspace{-10mm} \rotatebox{90}{\parbox{2cm}{\red{ResNet50}}} 
\end{subfigure} 
\begin{subfigure}[ht!]{0.105\linewidth}
\includegraphics[page=1, trim = 0mm 0mm 0mm 0mm, clip, width=1.0\linewidth]{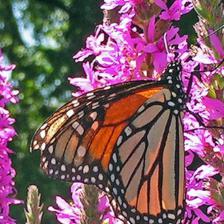}
\end{subfigure} 
\begin{subfigure}[ht!]{0.105\linewidth}
\includegraphics[page=1, trim = 0mm 0mm 0mm 0mm, clip, width=1.0\linewidth]{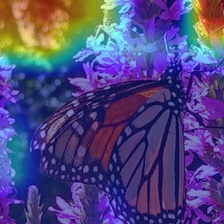}
\end{subfigure} 
\begin{subfigure}[ht!]{0.105\linewidth}
\includegraphics[page=1, trim = 0mm 0mm 0mm 0mm, clip, width=1.0\linewidth]{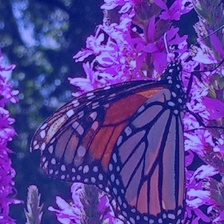}
\end{subfigure}
\begin{subfigure}[ht!]{0.105\linewidth}
\includegraphics[page=1, trim = 0mm 0mm 0mm 0mm, clip, width=1.0\linewidth]{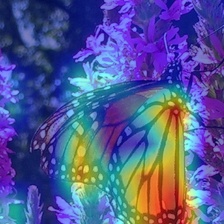}
\end{subfigure}
\begin{subfigure}[ht!]{0.105\linewidth}
\includegraphics[page=1, trim = 0mm 0mm 0mm 0mm, clip, width=1.0\linewidth]{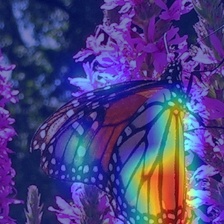}
\end{subfigure} 
\begin{subfigure}[ht!]{0.105\linewidth}
\includegraphics[page=1, trim = 0mm 0mm 0mm 0mm, clip, width=1.0\linewidth]{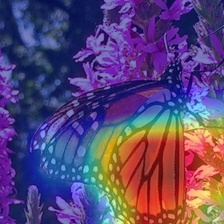}
\end{subfigure} 
\begin{subfigure}[ht!]{0.105\linewidth}
\includegraphics[page=1, trim = 0mm 0mm 0mm 0mm, clip, width=1.0\linewidth]{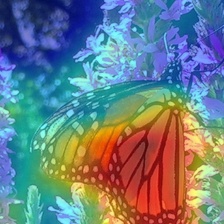}
\end{subfigure} \\ 
\begin{subfigure}[ht!]{0.018\linewidth}
\vspace{-14mm} \rotatebox{90}{\parbox{2cm}{\red{Ours}}}
\end{subfigure}
\begin{subfigure}[ht!]{0.105\linewidth}
\includegraphics[page=1, trim = 0mm 0mm 0mm 0mm, clip, width=1.0\linewidth]{Figure/source/IN_layerwise_visualization/ILSVRC2012_val_00011374.JPEG}
\end{subfigure} 
\begin{subfigure}[ht!]{0.105\linewidth}
\includegraphics[page=1, trim = 0mm 0mm 0mm 0mm, clip, width=1.0\linewidth]{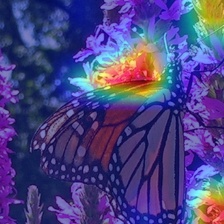}
\end{subfigure} 
\begin{subfigure}[ht!]{0.105\linewidth}
\includegraphics[page=1, trim = 0mm 0mm 0mm 0mm, clip, width=1.0\linewidth]{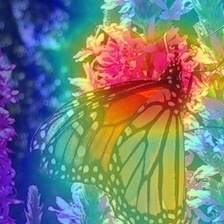}
\end{subfigure}
\begin{subfigure}[ht!]{0.105\linewidth}
\includegraphics[page=1, trim = 0mm 0mm 0mm 0mm, clip, width=1.0\linewidth]{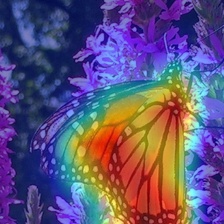}
\end{subfigure}
\begin{subfigure}[ht!]{0.105\linewidth}
\includegraphics[page=1, trim = 0mm 0mm 0mm 0mm, clip, width=1.0\linewidth]{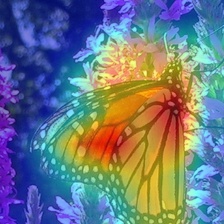}
\end{subfigure}
\begin{subfigure}[ht!]{0.105\linewidth}
\includegraphics[page=1, trim = 0mm 0mm 0mm 0mm, clip, width=1.0\linewidth]{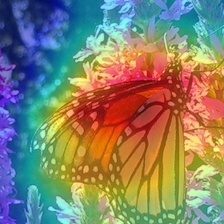}
\end{subfigure} 
\begin{subfigure}[ht!]{0.105\linewidth}
\includegraphics[page=1, trim = 0mm 0mm 0mm 0mm, clip, width=1.0\linewidth]{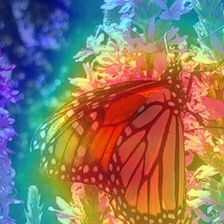}
\end{subfigure}\\ \vspace{1mm}
\begin{subfigure}[ht!]{0.018\linewidth}
\vspace{-10mm} \rotatebox{90}{\parbox{2cm}{\red{ResNet50}}} 
\end{subfigure}
\begin{subfigure}[ht!]{0.105\linewidth}
\includegraphics[page=1, trim = 0mm 0mm 0mm 0mm, clip, width=1.0\linewidth]{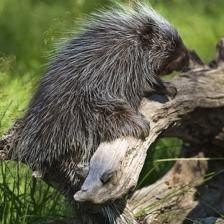}
\end{subfigure} 
\begin{subfigure}[ht!]{0.105\linewidth}
\includegraphics[page=1, trim = 0mm 0mm 0mm 0mm, clip, width=1.0\linewidth]{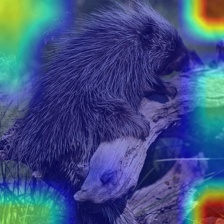}
\end{subfigure} 
\begin{subfigure}[ht!]{0.105\linewidth}
\includegraphics[page=1, trim = 0mm 0mm 0mm 0mm, clip, width=1.0\linewidth]{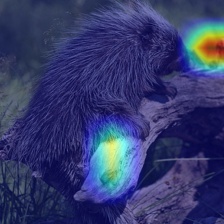}
\end{subfigure}
\begin{subfigure}[ht!]{0.105\linewidth}
\includegraphics[page=1, trim = 0mm 0mm 0mm 0mm, clip, width=1.0\linewidth]{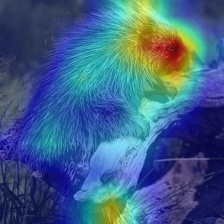}
\end{subfigure}
\begin{subfigure}[ht!]{0.105\linewidth}
\includegraphics[page=1, trim = 0mm 0mm 0mm 0mm, clip, width=1.0\linewidth]{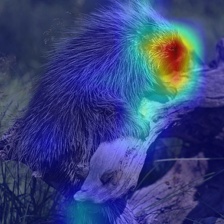}
\end{subfigure} 
\begin{subfigure}[ht!]{0.105\linewidth}
\includegraphics[page=1, trim = 0mm 0mm 0mm 0mm, clip, width=1.0\linewidth]{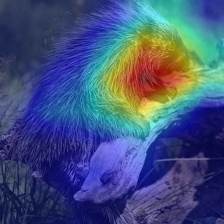}
\end{subfigure} 
\begin{subfigure}[ht!]{0.105\linewidth}
\includegraphics[page=1, trim = 0mm 0mm 0mm 0mm, clip, width=1.0\linewidth]{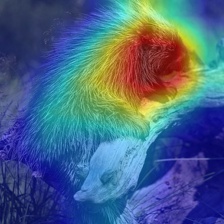}
\end{subfigure} \\
\hspace{0.1mm}
\begin{subfigure}[ht!]{0.018\linewidth}
\vspace{-14mm} \rotatebox{90}{\parbox{2cm}{\red{Ours}}}
\end{subfigure}
\begin{subfigure}[ht!]{0.105\linewidth}
\includegraphics[page=1, trim = 0mm 0mm 0mm 0mm, clip, width=1.0\linewidth]{Figure/source/IN_layerwise_visualization/ILSVRC2012_val_00047267.JPEG}
\end{subfigure} 
\begin{subfigure}[ht!]{0.105\linewidth}
\includegraphics[page=1, trim = 0mm 0mm 0mm 0mm, clip, width=1.0\linewidth]{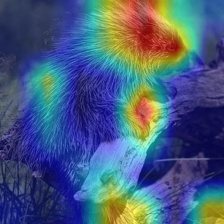}
\end{subfigure} 
\begin{subfigure}[ht!]{0.105\linewidth}
\includegraphics[page=1, trim = 0mm 0mm 0mm 0mm, clip, width=1.0\linewidth]{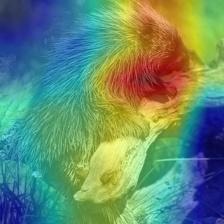}
\end{subfigure}
\begin{subfigure}[ht!]{0.105\linewidth}
\includegraphics[page=1, trim = 0mm 0mm 0mm 0mm, clip, width=1.0\linewidth]{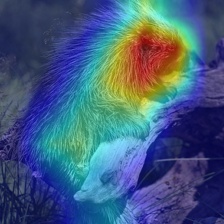}
\end{subfigure}
\begin{subfigure}[ht!]{0.105\linewidth}
\includegraphics[page=1, trim = 0mm 0mm 0mm 0mm, clip, width=1.0\linewidth]{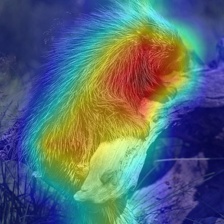}
\end{subfigure}
\begin{subfigure}[ht!]{0.105\linewidth}
\includegraphics[page=1, trim = 0mm 0mm 0mm 0mm, clip, width=1.0\linewidth]{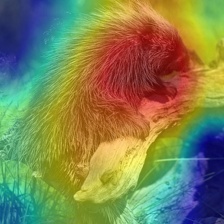}
\end{subfigure} 
\begin{subfigure}[ht!]{0.105\linewidth}
\includegraphics[page=1, trim = 0mm 0mm 0mm 0mm, clip, width=1.0\linewidth]{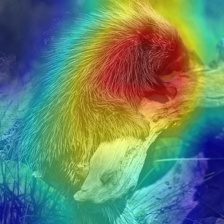}
\end{subfigure}
\vspace{-2mm}
\caption{\small \red{{\bf Grad-CAM visualization of intermediate features.} We visualize the highlighted features by Grad-CAM of the intermediate features produced by the six successive layers in the last two bottlenecks of the final stage in ResNet50 and ResNet50 ({\it hybrid}), respectively. Each visualized feature from left to right gets closer to the final output features of a model. Each feature of conv5\_x\_y denotes the output features of the y-th layer in the x-th bottleneck, where a bottleneck has the 1${\times}$1 convolutions at the 1st and 3rd layers) and 3x3 convolution/the max-pool operation at the 2nd layer in ResNet50/our ResNet50 ({\it hybrid)}. The input images are randomly picked from the ImageNet validation set.}}
\label{fig:fig_layerwise_feature_visualization}
\vspace{-3mm}
\end{figure*}

Since the spatial parameter-free operations have the receptive field~\citep{luo2016understanding} like convolutions, we conjecture such simple operations can capture striking pixels (or patches). We empirically support the conjecture by providing a visualization of the final output features of the ImageNet-pretrained ResNet50 and our ResNet50 ({\it hybrid}) in Fig.\ref{fig:fig_feature_visualization}. We observe that two models successfully localize the foreground object in common for all the images. A noticeable difference is that ResNet50 tends to focus more on specific crucial regions but ResNet50 ({\it hybrid}) on wider regions (a similar trend will be observed in Fig.\ref{fig:fig_layerwise_feature_visualization}). Note that the area of the highlighted region does not directly link to a model's accuracy (in fact, ResNet50 has higher accuracy). We believe that this indicates how much a model learns localizable features and can be utilized to understand the learning dynamics of various models. 

We further conjecture that a 1${\times}$1 convolution can complement the expressiveness of a parameter-free operation such as the max-pool operation or avg-pool operation by nonlinearly mixing the features, which is computed by the parameter-free operation. To make a ground for the conjecture, we would like to visualize the intermediate features extracted from 1) each of two 1${\times}$1 convolutions; 2) the spatial operations including the 3${\times}$3 convolution and the max-pool operation in a bottleneck block. We choose the last two bottlenecks in the final stage conv5 and visualize the output features of the aforementioned layers.

Fig.\ref{fig:fig_layerwise_feature_visualization} shows the highlighted features produced by the three different input images randomly picked from the ImageNet validation set. We use the identical models employed to visualize the final features shown in Fig.\ref{fig:fig_feature_visualization}. We let conv5\_x\_y denote the output features of the y-th layer in the x-th bottleneck of a specific model; for example, each visualized feature of conv5\_2\_2 in ResNet50 and our model (ResNet50 ({\it hybrid})) indicates the output features of the 3${\times}$3 convolution and the spatial max-pool operation, respectively. First, we observe the 1${\times}$1 convolutions refine the features to be more discriminative; when comparing the features in conv5\_x\_2 with those of conv5\_x\_3, the output features usually get more highlighted on the crucial region in the foreground objects. 
Furthermore, it seems that the 1${\times}$1 convolutions change more of the previous features in our model, which may come from the different output features by the 3${\times}$3 convolutions and the max-pool operations (compare the output features of ResNet50 with ours in conv5\_x\_2). 
Therefore, based on the observations, the 1${\times}$1 convolution performs to make features more discriminative, and they do more with the parameter-free operations like the max-pool operation.

\section{More Visualizations}
In this section, we further discuss the studied materials with the additional graphs and figures to provide more detailed information. \S\ref{sec_sub:extension_mbs} contains extended trade-off graphs of our study in \S\ref{sec_sub:empirical_studies}; \S\ref{sec_sub:deform_max_offsets} discusses predicted offsets by the proposed deformable max-pool operation; finally in \S\ref{sec_sub:trade_offs}, we discuss the performance trade-offs of the models on the ImageNet datasets in \S\ref{sec_sub:imagenet_exp}.

\subsection{Multiple Bottlenecks Study (cont'd)}
\label{sec_sub:extension_mbs}
\begin{figure*}[t]
\small
\centering
\hspace{-1mm}
\begin{subfigure}[ht!]{0.245\linewidth}
\includegraphics[page=1, trim = 0mm 0mm 0mm 0mm, clip, width=1.0\linewidth]{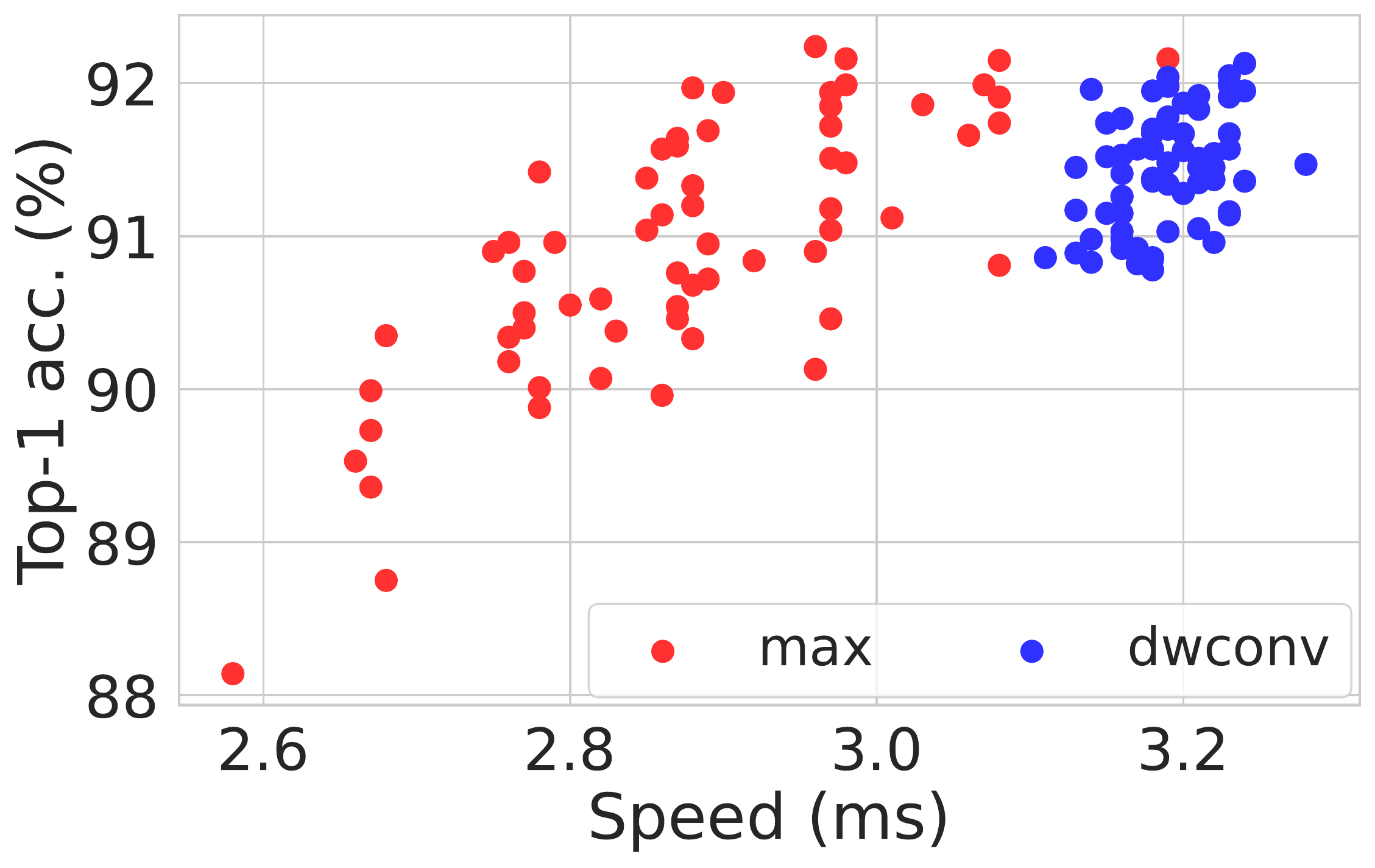}
\caption{\small Speed (1-image)}
\end{subfigure}  
\begin{subfigure}[ht!]{0.245\linewidth}
\includegraphics[page=1, trim = 0mm 0mm 0mm 0mm, clip, width=1.0\linewidth]{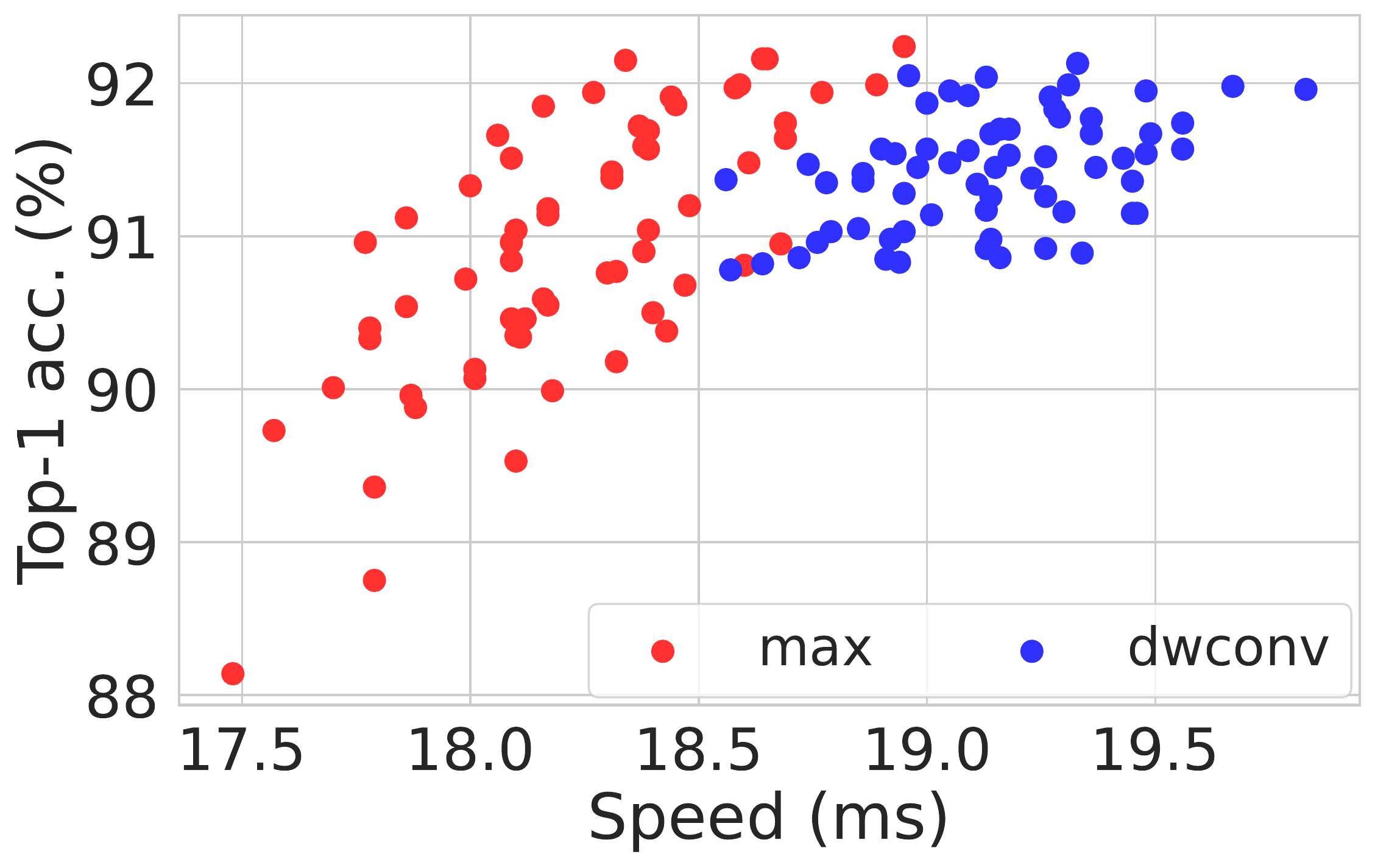}
\caption{\small Speed (256-images)}
\end{subfigure} 
\begin{subfigure}[ht!]{0.245\linewidth}
\includegraphics[page=1, trim = 0mm 0mm 0mm 0mm, clip, width=1.0\linewidth]{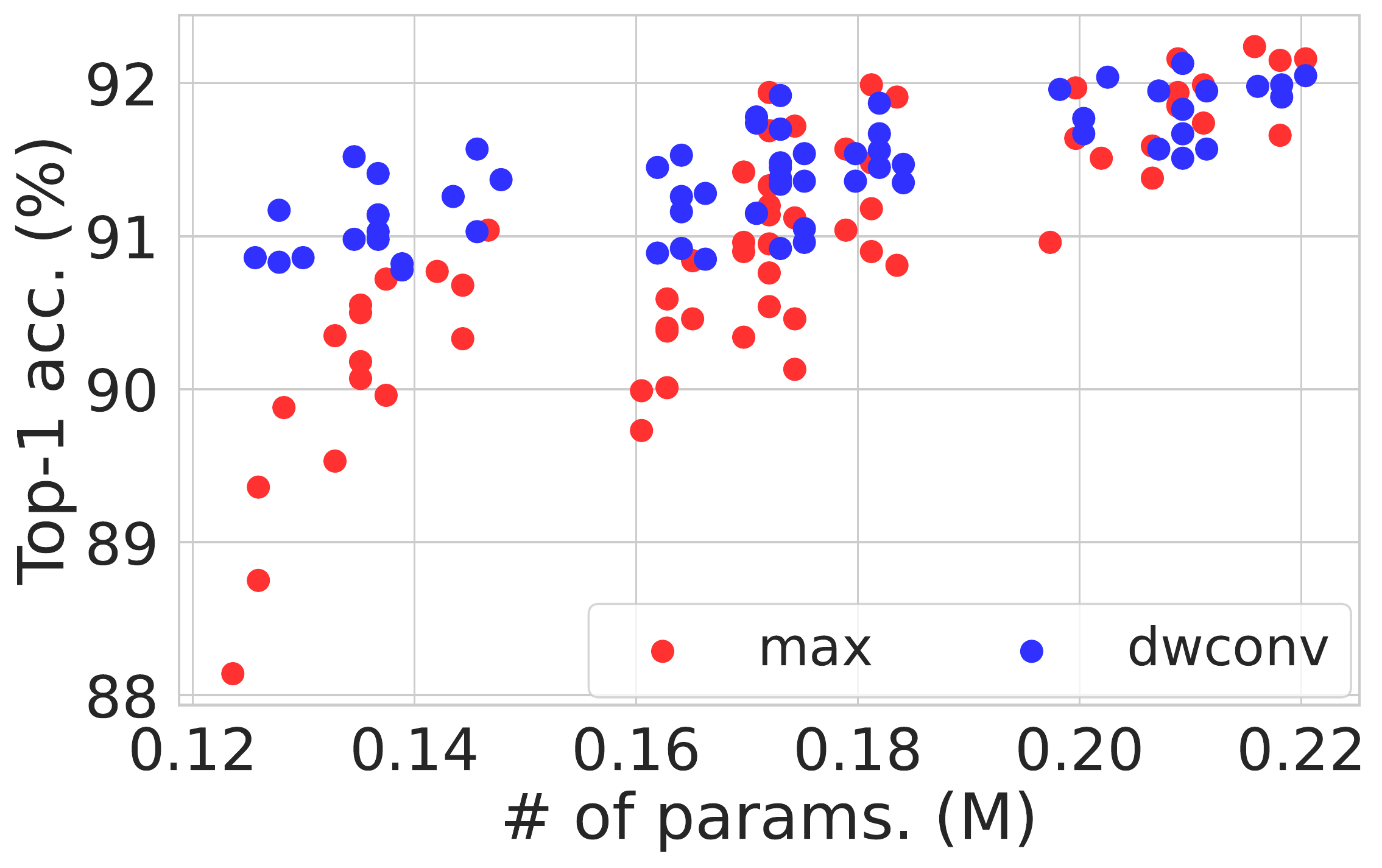}
\caption{\small \# parameters}
\end{subfigure} 
\begin{subfigure}[ht!]{0.245\linewidth}
\includegraphics[page=1, trim = 0mm 0mm 0mm 0mm, clip, width=1.0\linewidth]{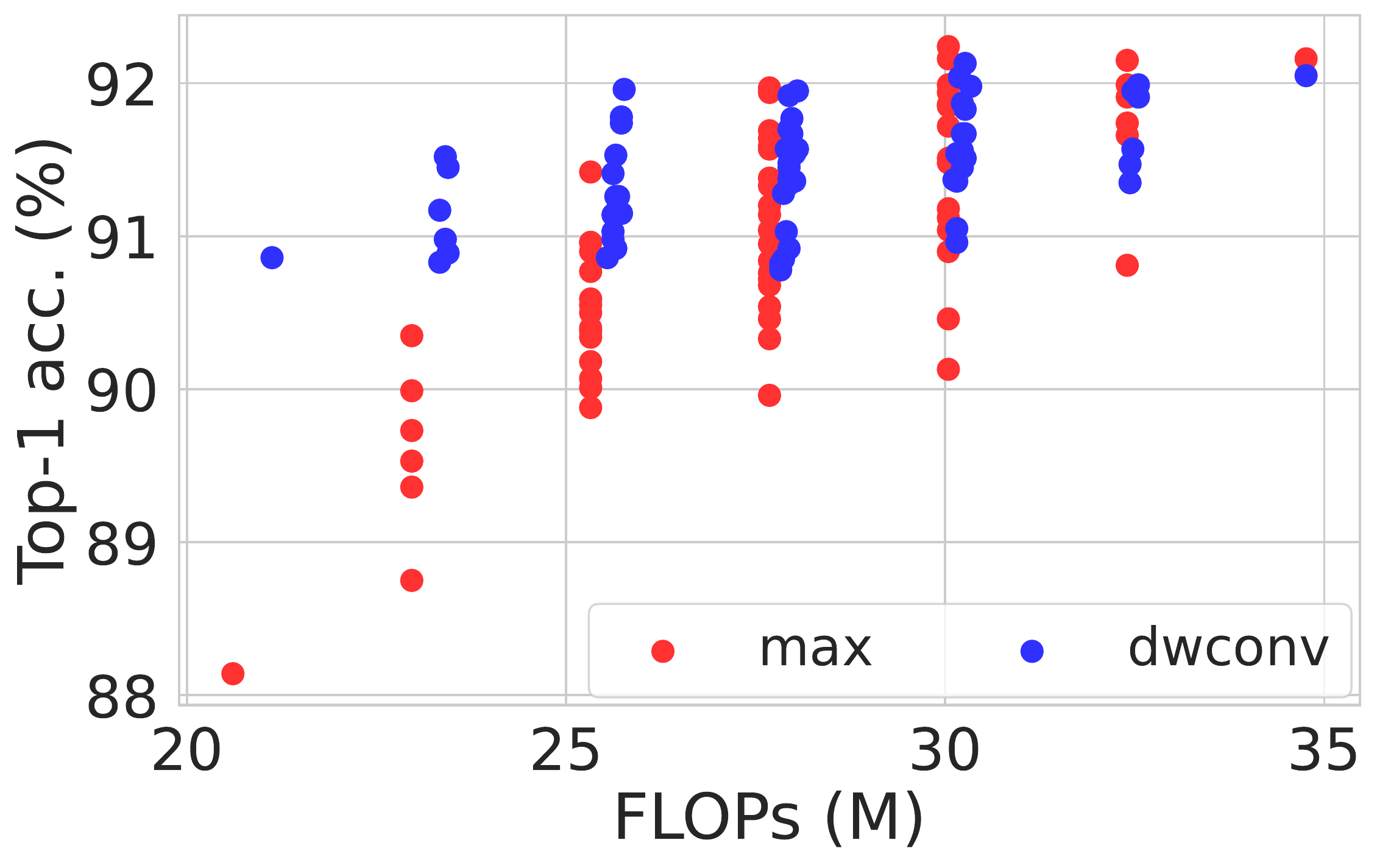}
\caption{\small FLOPs}
\end{subfigure} 
\vspace{-2mm}
\caption{\small \red{{\bf Multiple bottlenecks study (cont'd).} The entire models trained in the multiple bottlenecks study in \S\ref{sec_sub:empirical_studies} are visualized in the comparison graphs: (a) Accuracy vs. speed with the batch size of 1; (b) Accuracy vs. speed with the batch size of 256; (c) Accuracy vs. \# parameters; (d) Accuracy vs. FLOPs. The graphs includes the 20\% best-performing models plotted in Fig.\ref{fig:fig_multilayer}. The depthwise convolution ({\color{blue} blue dots}) shows efficiency again in \# parameters and FLOPs, but the max-pool (i.e., {\color{red} red dots}) has clear benefits in the speed measures. Note that the speed gap between the two operations gets larger when processing multiple images, as shown through (a) and (b). }}
\label{fig:fig_multilayer_full}
\vspace{-3mm}
\end{figure*}

Fig.\ref{fig:fig_multilayer_full} shows the performance of the entire models which are trained in \S\ref{sec_sub:empirical_studies}. The graphs includes the 20\% best-performing models shown in Fig.\ref{fig:fig_multilayer}. As expected, the models using many max-pool operations have degraded accuracies but show clear speed benefits; there is about a 1.5\% accuracy gap between the fastest models using each operation yet more than a 5ms speed gap between them. 

\subsection{Predicted Offsets by Deformable Max-pool Operation}
\label{sec_sub:deform_max_offsets}
\begin{figure*}[t]
\small
\centering
\hspace{-1mm}
\begin{subfigure}[ht!]{0.22\linewidth}
\includegraphics[page=1, trim = 0mm 0mm 0mm 0mm, clip, width=1.0\linewidth]{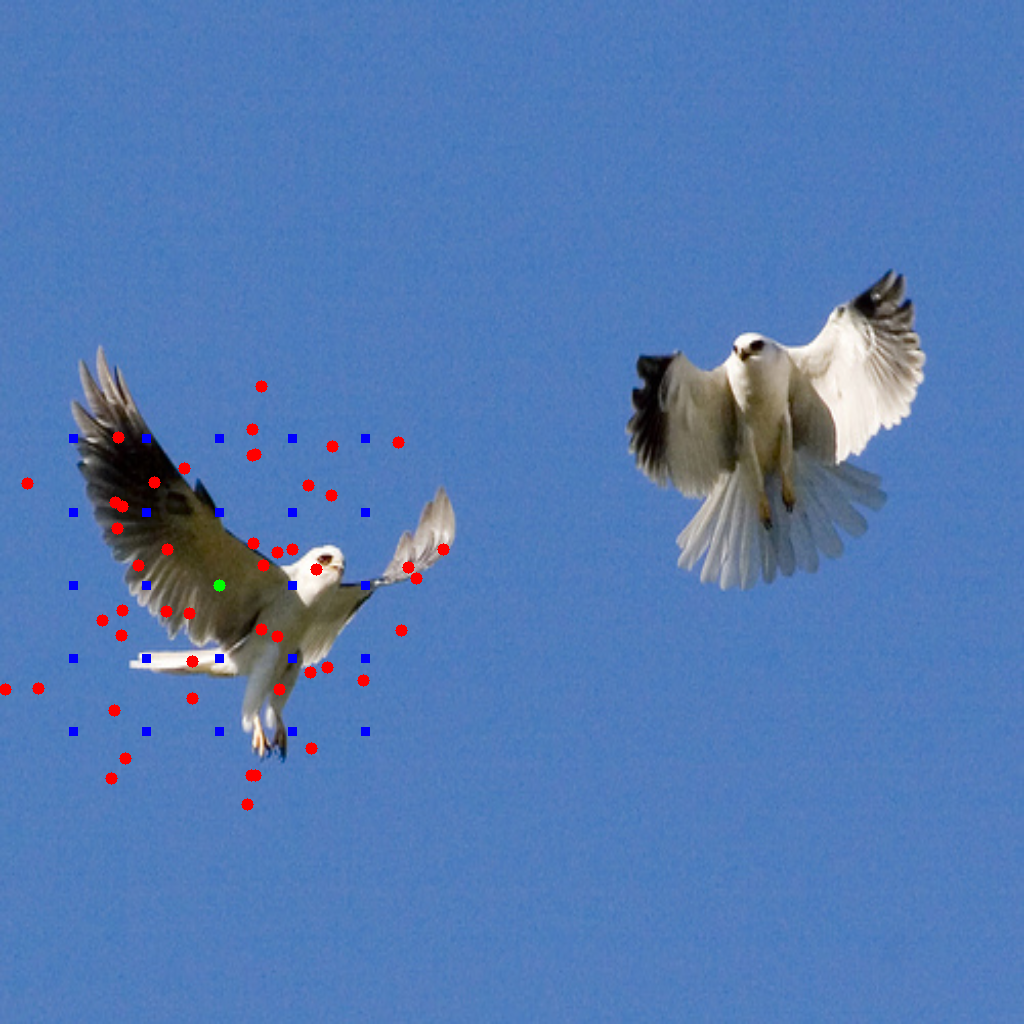} 
\end{subfigure}  \  \
\begin{subfigure}[ht!]{0.22\linewidth}
\includegraphics[page=1, trim = 0mm 0mm 0mm 0mm, clip, width=1.0\linewidth]{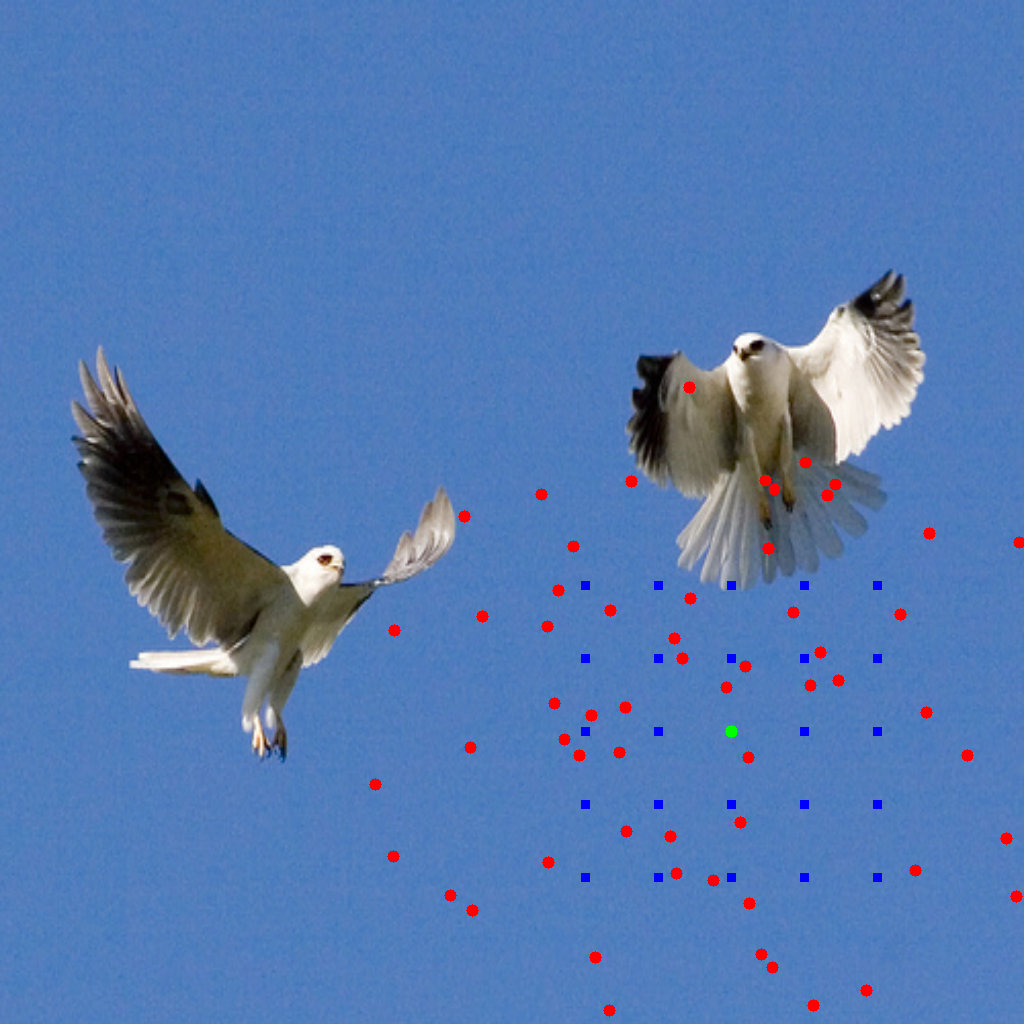}
\end{subfigure} \ \ \  \
\begin{subfigure}[ht!]{0.22\linewidth}
\includegraphics[page=1, trim = 0mm 0mm 0mm 0mm, clip, width=1.0\linewidth]{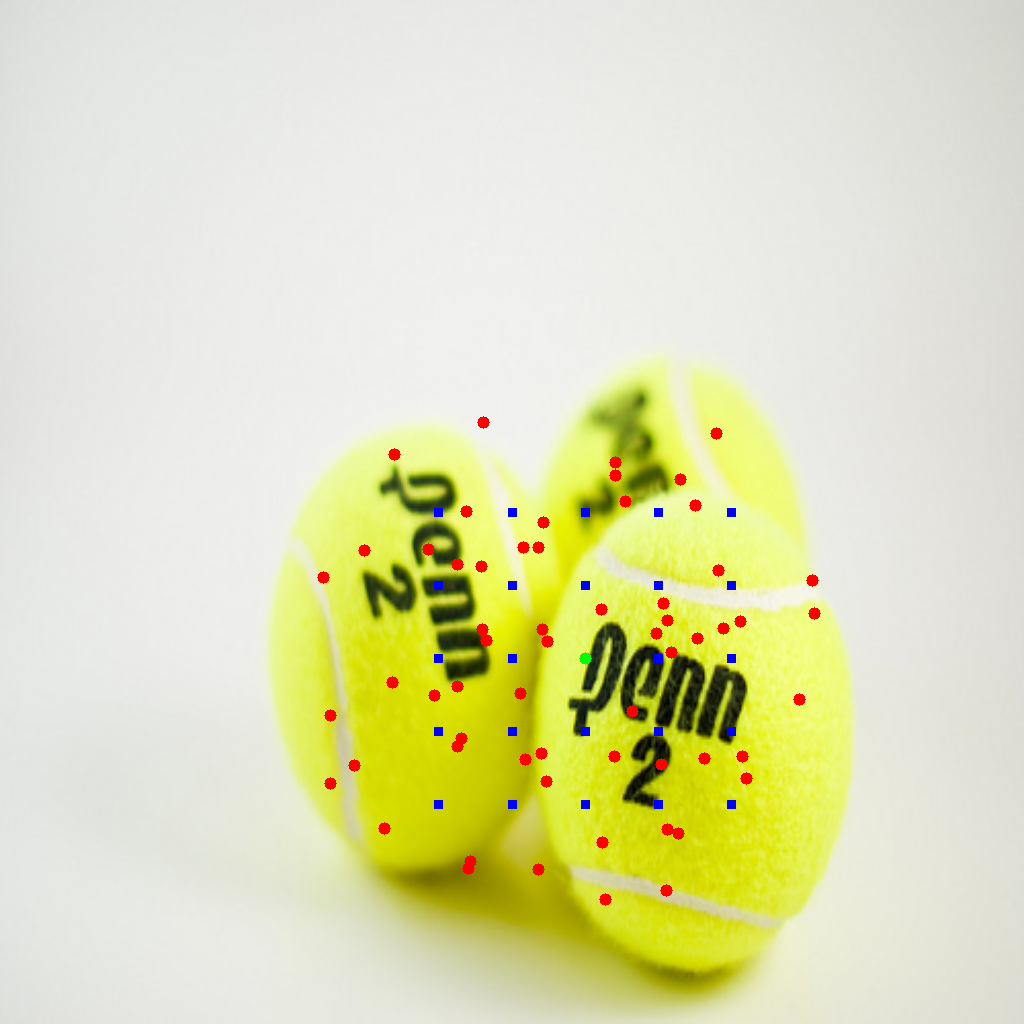}
\end{subfigure} \  \
\begin{subfigure}[ht!]{0.22\linewidth}
\includegraphics[page=1, trim = 0mm 0mm 0mm 0mm, clip, width=1.0\linewidth]{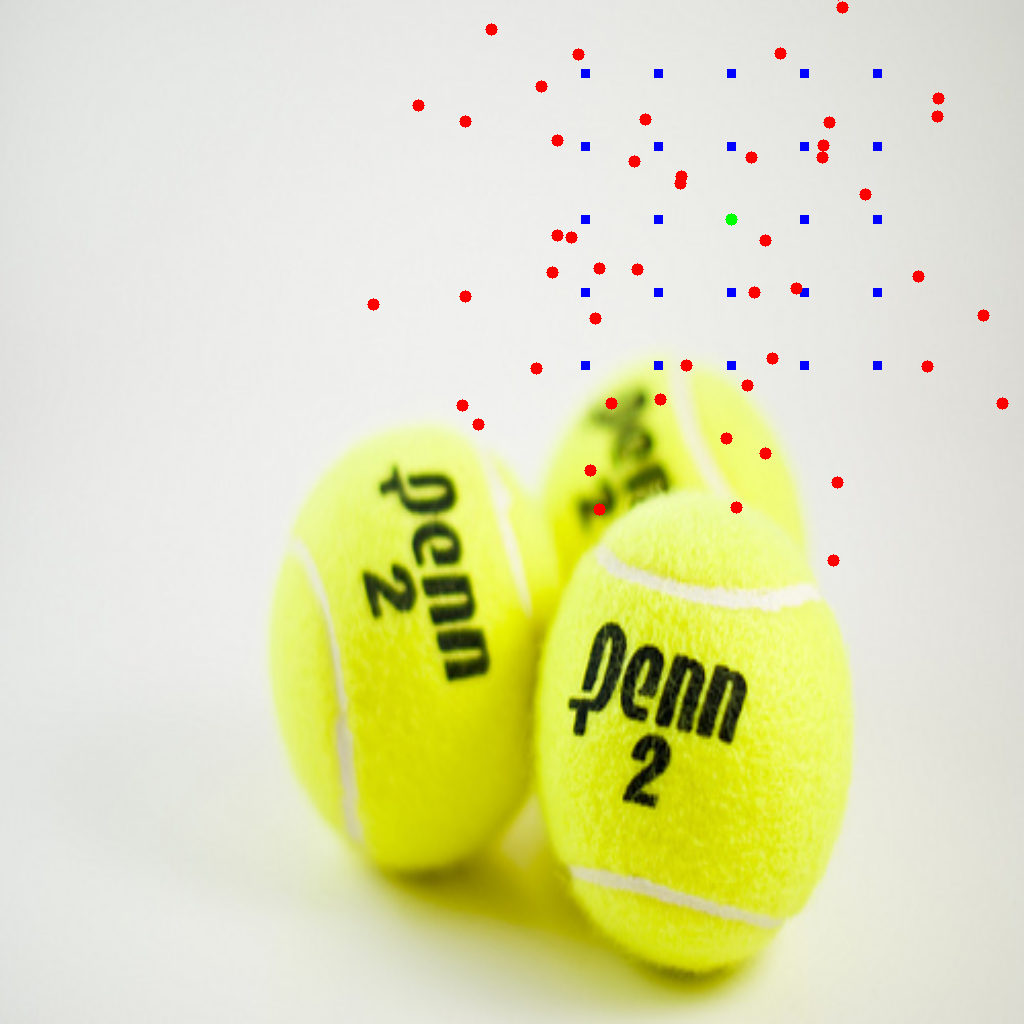}
\end{subfigure} 
\vspace{-2mm}
\caption{\small \red{{\bf Visualization of predicted offsets.} We illustrate the predicted offsets by our deformable max-pool operation from the images from the validation set in ImageNet. Each sampled image has two predicted offsets on a foreground object (left) and the background (right). We plot 1) the predicted offsets ({\color{red} red dots}); 2) the initial grid-like points ({\color{blue} blue dots}) with the center ({\color{green} green dots}) in each figure.}}
\label{fig:fig_deform_max_visualization}
\vspace{-3mm}
\end{figure*}

We visualize the predicted offsets by the proposed deformable max-pool operation (deform\_max) with the sample images in Fig.\ref{fig:fig_deform_max_visualization}. We plot the aggregated offsets of the last two deform\_maxs in the final stage of ResNet50 with the initial points, which look like the grid-like offset points of the two successive regular convolutions. We observe the offsets are concentrated on the objects when the center is on each foreground object. The offsets spread widely when the center is on the background, which is similarly observed in the deformable convolution paper~\citep{dai2017deformable,zhu2019deformable}.
It is surprising that our models only trained with the ImageNet's class labels without strong supervisions (i.e., detection boxes or segmentation masks) show the improved localization capability. Furthermore, albeit our model does not be trained with the background class, each shape of the predicted offsets on foreground and background looks quite different as shown in Fig.\ref{fig:fig_deform_max_visualization}.

We would like to stress the main differences of the experimental settings with those in the deformable convolution papers. 
First, our model is trained on ImageNet only with the class labels without any strong supervision so that the models may have a weaker localization capability than the models trained with more supervision, such as on detection and segmentation tasks. Second, our model incorporates much more deformable operations in the entire stages of a ResNet, so the dynamics of predicting offsets would be different from the deformable convolution where the only later stages have the deformable convolutions.

\subsection{Performance Trade-offs}
\label{sec_sub:trade_offs}
\begin{figure*}[h]
\small
\centering
\hspace{-1mm}
\begin{subfigure}[ht!]{0.3\linewidth}
\includegraphics[page=1, trim = 0mm 0mm 0mm 0mm, clip, width=1.0\linewidth]{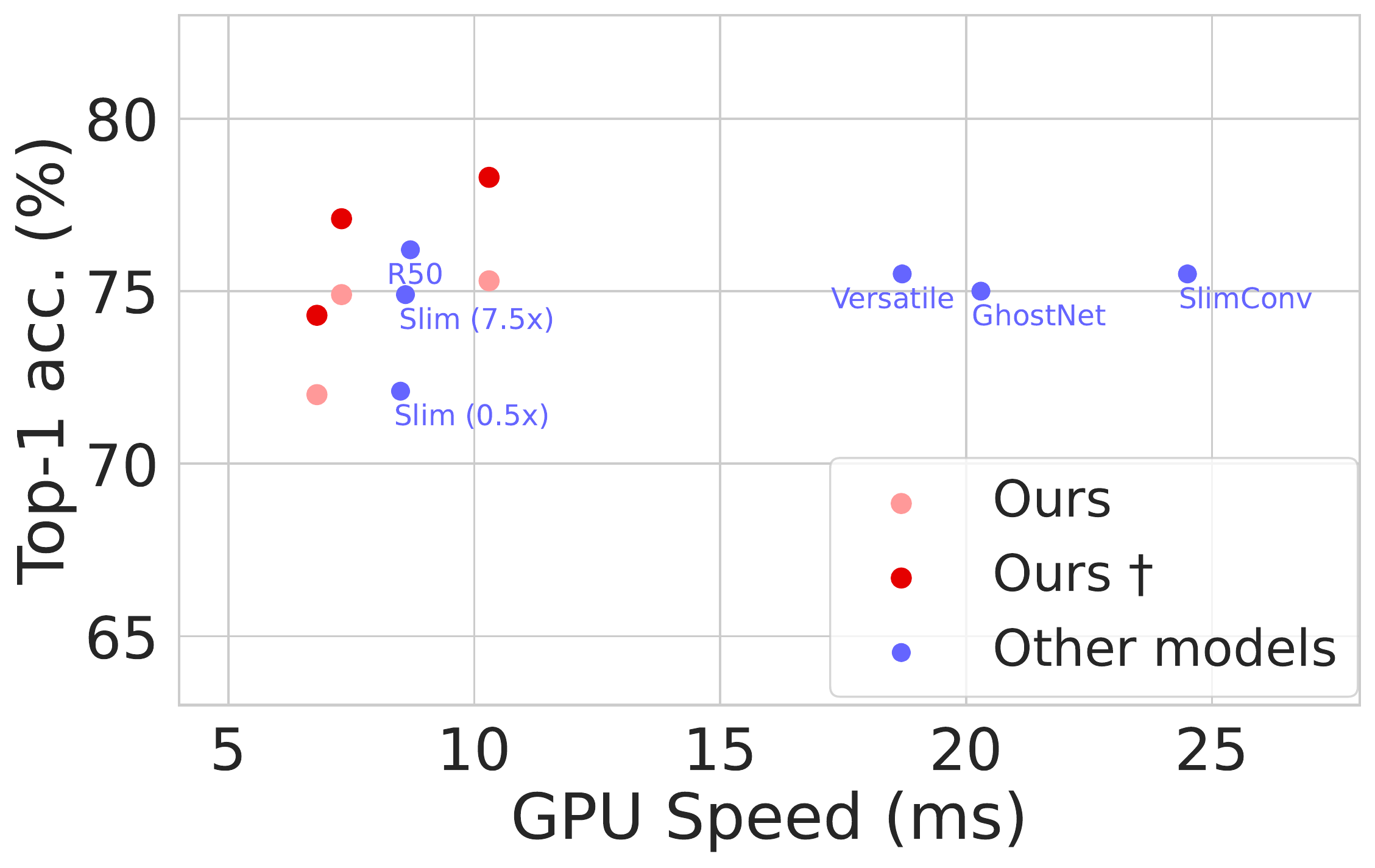}
\caption{\small GPU speed}
\end{subfigure}  \ \ 
\begin{subfigure}[ht!]{0.3\linewidth}
\includegraphics[page=1, trim = 0mm 0mm 0mm 0mm, clip, width=1.0\linewidth]{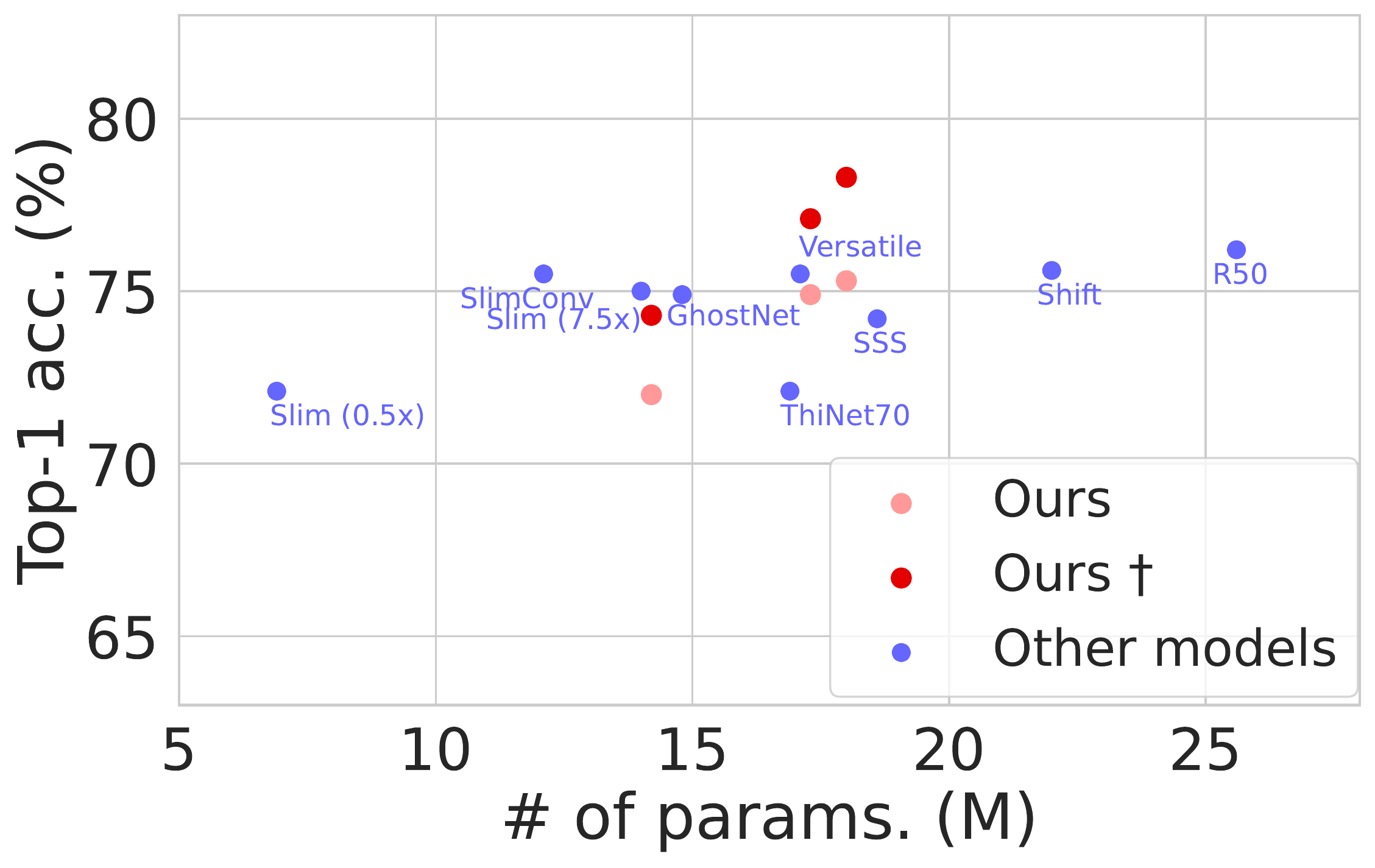}
\caption{\small \# parameters}
\end{subfigure} \ \
\begin{subfigure}[ht!]{0.3\linewidth}
\includegraphics[page=1, trim = 0mm 0mm 0mm 0mm, clip, width=1.0\linewidth]{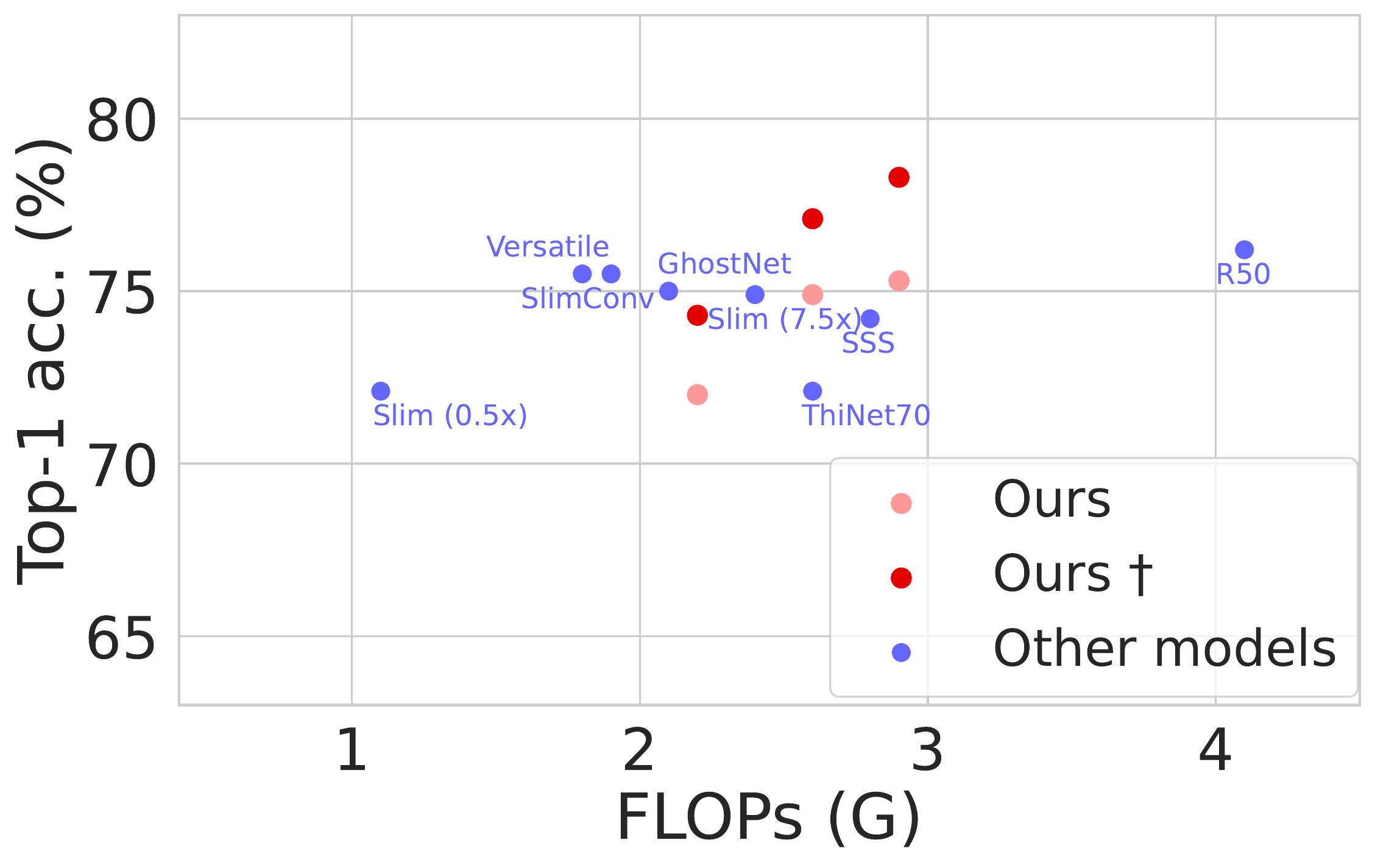}
\caption{\small FLOPs}
\end{subfigure} 
\vspace{-2mm}
\caption{\small \red{{\bf Performance Trade-offs of the models in Table~\ref{table:imagenet_comparison}.} We visualize the trade-offs between (a) accuracy and speed; (b) accuracy and \# parameters; (c) accuracy and FLOPs, respectively. Ours denote the performance of Ours-R50 (max), Ours-R50 ({\it hybrid}), and Ours-R50 (deform\_max); we also plot the models trained with the further training recipes.}}
\label{fig:fig_table3_visualization}
\vspace{-4.4mm}
\end{figure*}

This work presents the potential usefulness of parameter-free operations such as the max-pool as a building block rather than pushing the performance to the limit. We first visualize the ImageNet performance comparison with many efficient models shown in Table~\ref{table:imagenet_comparison} in Fig.\ref{fig:fig_table3_visualization}. Fig.\ref{fig:fig_table3_visualization} shows our models have 
competitive trade-offs between accuracy and speed. Since we do not modify ResNet50 but just replace the spatial operation into parameter-free operations such as the max-pool, our models do not have much benefits in the number of parameters and FLOPs. %

\begin{figure*}[h]
\small
\centering
\hspace{-1mm}
\begin{subfigure}[ht!]{0.3\linewidth}
\includegraphics[page=1, trim = 0mm 0mm 0mm 0mm, clip, width=1.0\linewidth]{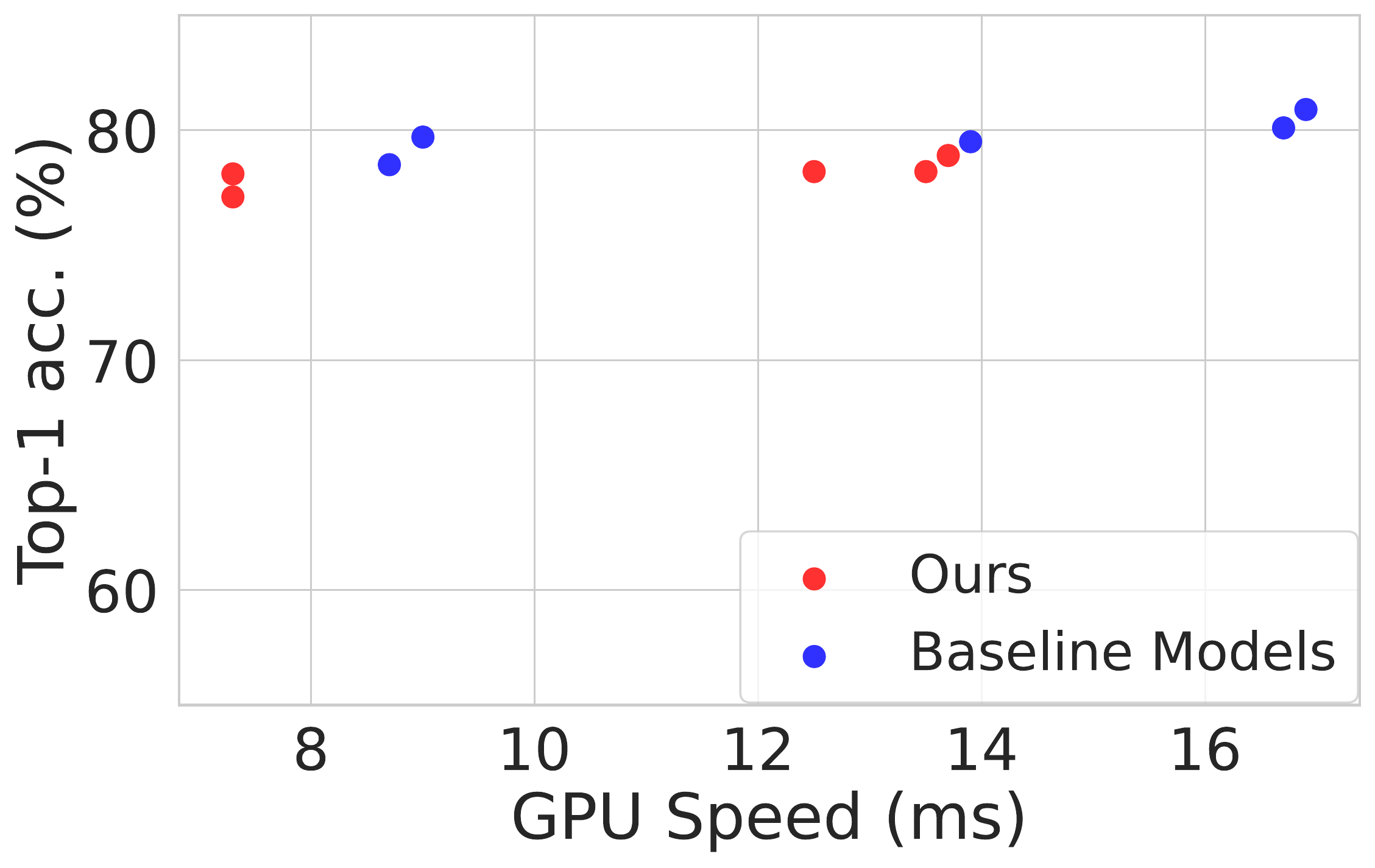}
\end{subfigure}  \ \
\begin{subfigure}[ht!]{0.3\linewidth}
\includegraphics[page=1, trim = 0mm 0mm 0mm 0mm, clip, width=1.0\linewidth]{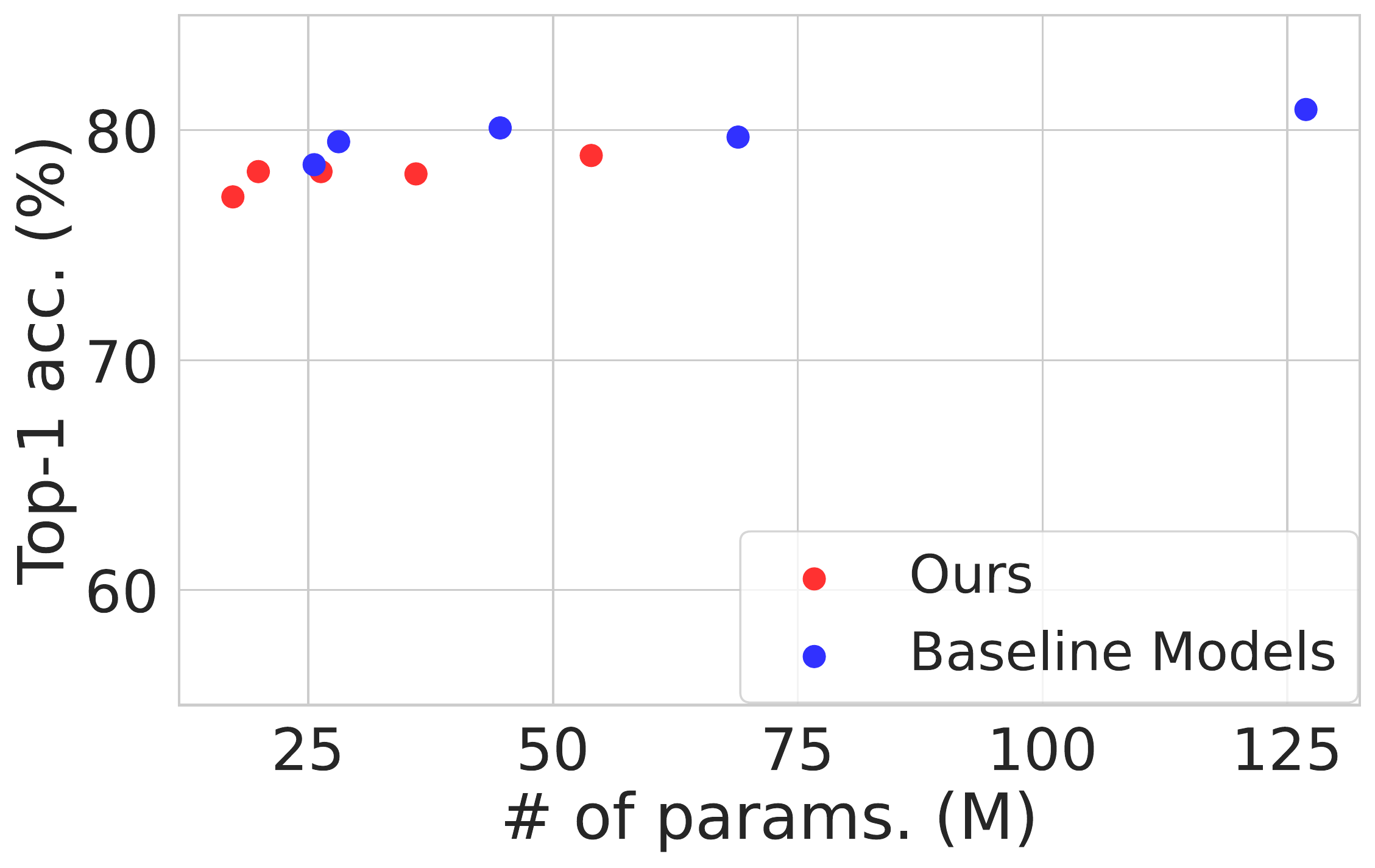}
\end{subfigure} \ \
\begin{subfigure}[ht!]{0.3\linewidth}
\includegraphics[page=1, trim = 0mm 0mm 0mm 0mm, clip, width=1.0\linewidth]{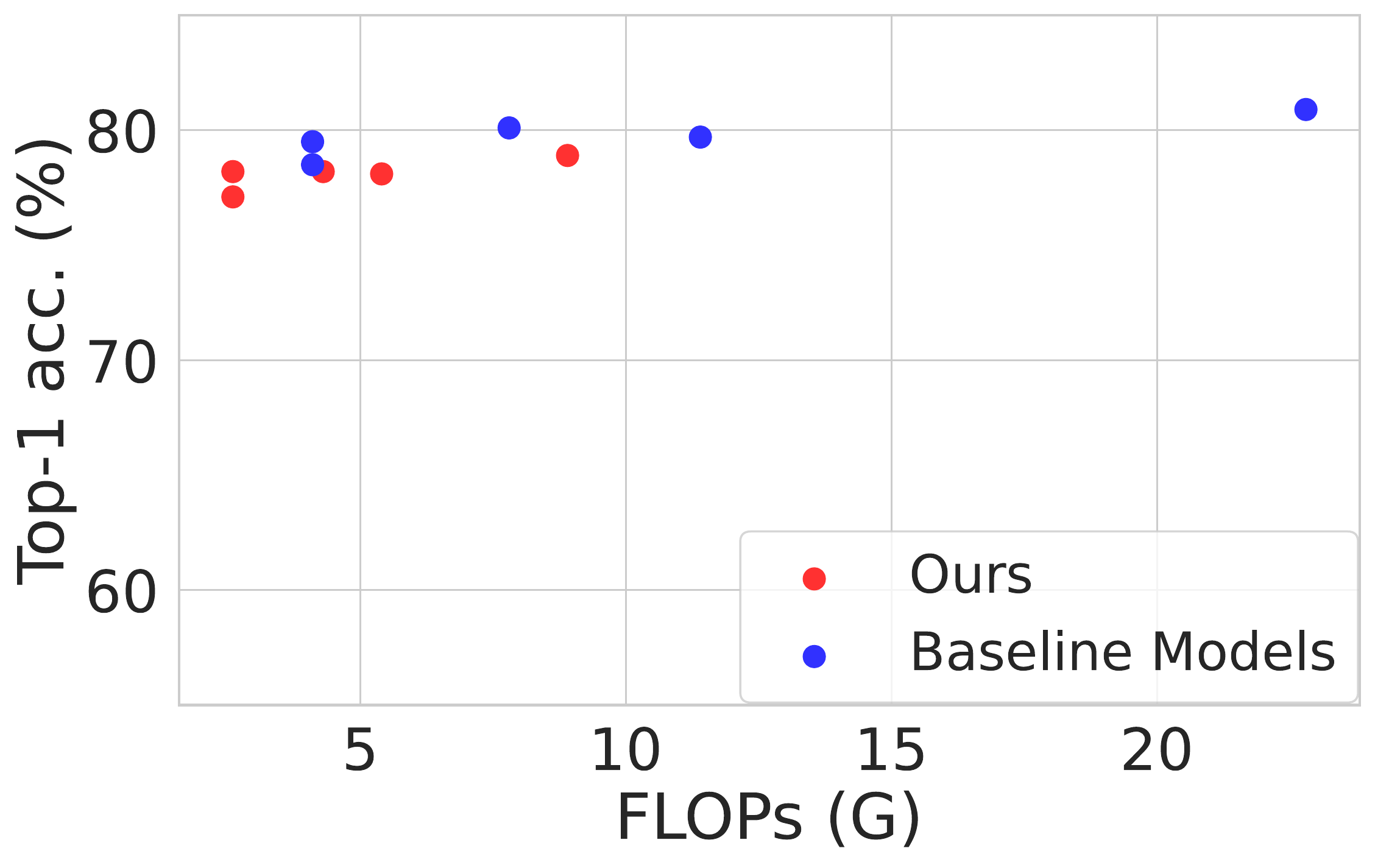}
\end{subfigure} \\
\hspace{-1mm}
\begin{subfigure}[ht!]{0.3\linewidth}
\includegraphics[page=1, trim = 0mm 0mm 0mm 0mm, clip, width=1.0\linewidth]{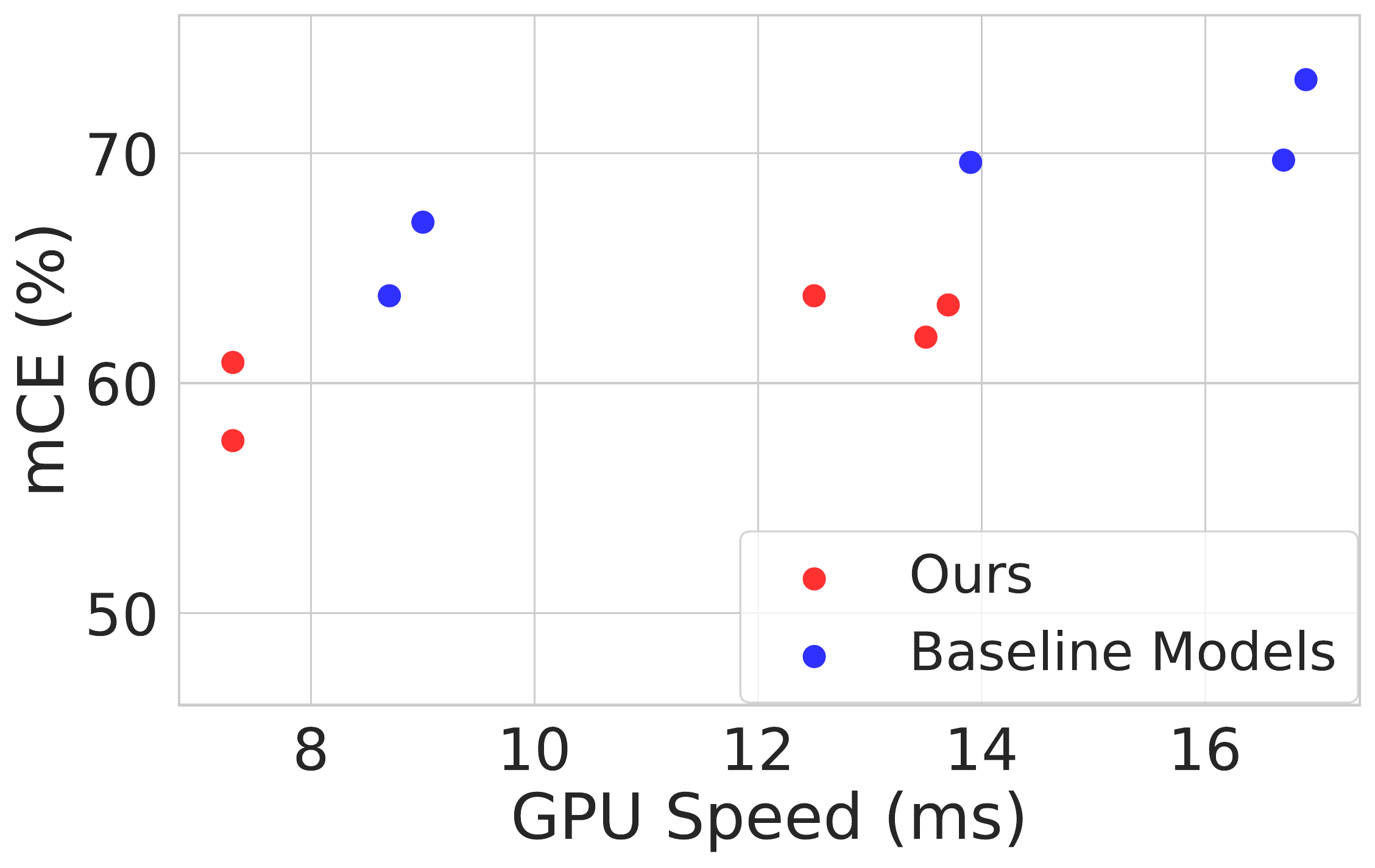}
\end{subfigure}  \ \
\begin{subfigure}[ht!]{0.3\linewidth}
\includegraphics[page=1, trim = 0mm 0mm 0mm 0mm, clip, width=1.0\linewidth]{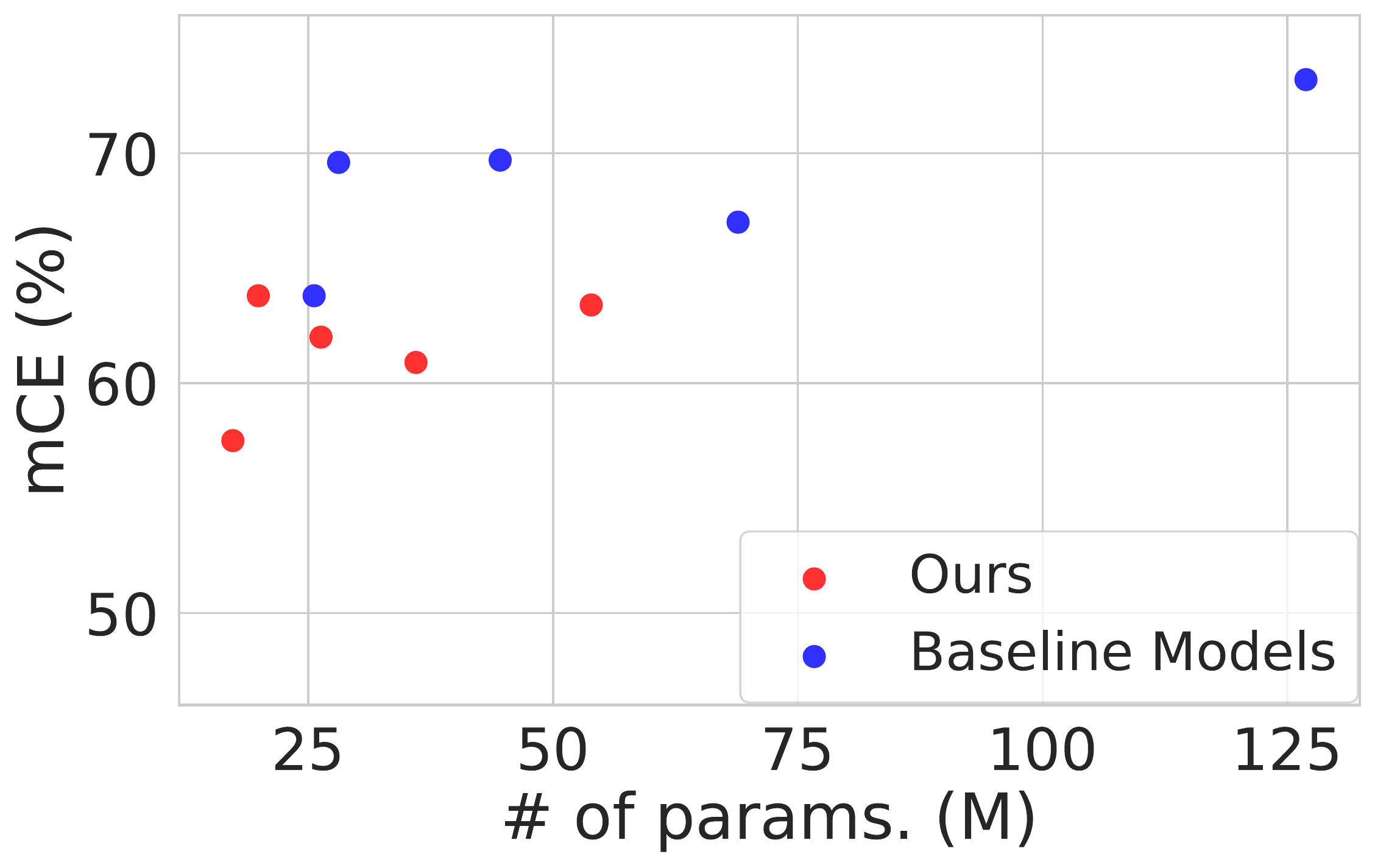}
\end{subfigure} \ \
\begin{subfigure}[ht!]{0.3\linewidth}
\includegraphics[page=1, trim = 0mm 0mm 0mm 0mm, clip, width=1.0\linewidth]{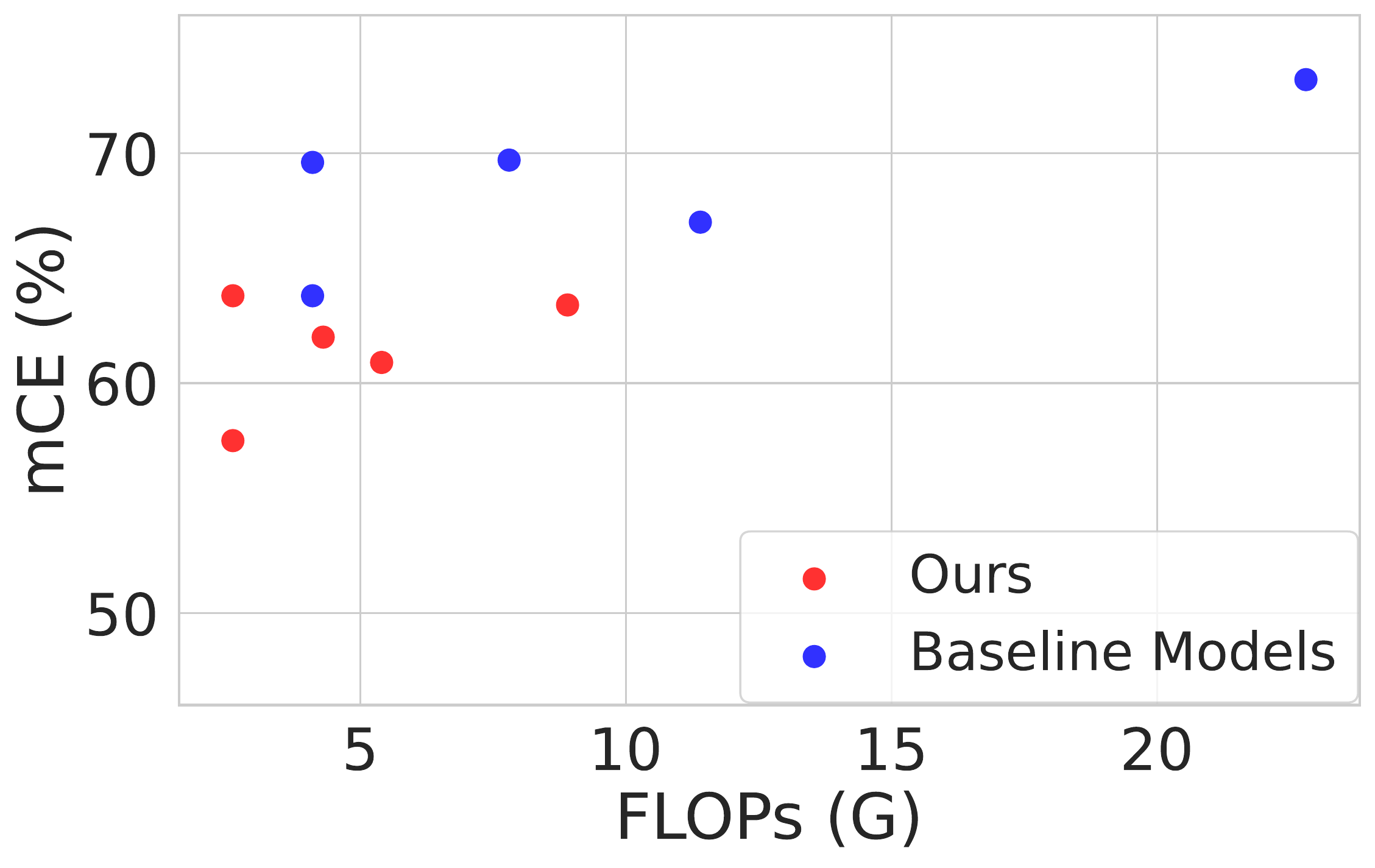}
\end{subfigure} \\
\hspace{-1mm}
\begin{subfigure}[ht!]{0.3\linewidth}
\includegraphics[page=1, trim = 0mm 0mm 0mm 0mm, clip, width=1.0\linewidth]{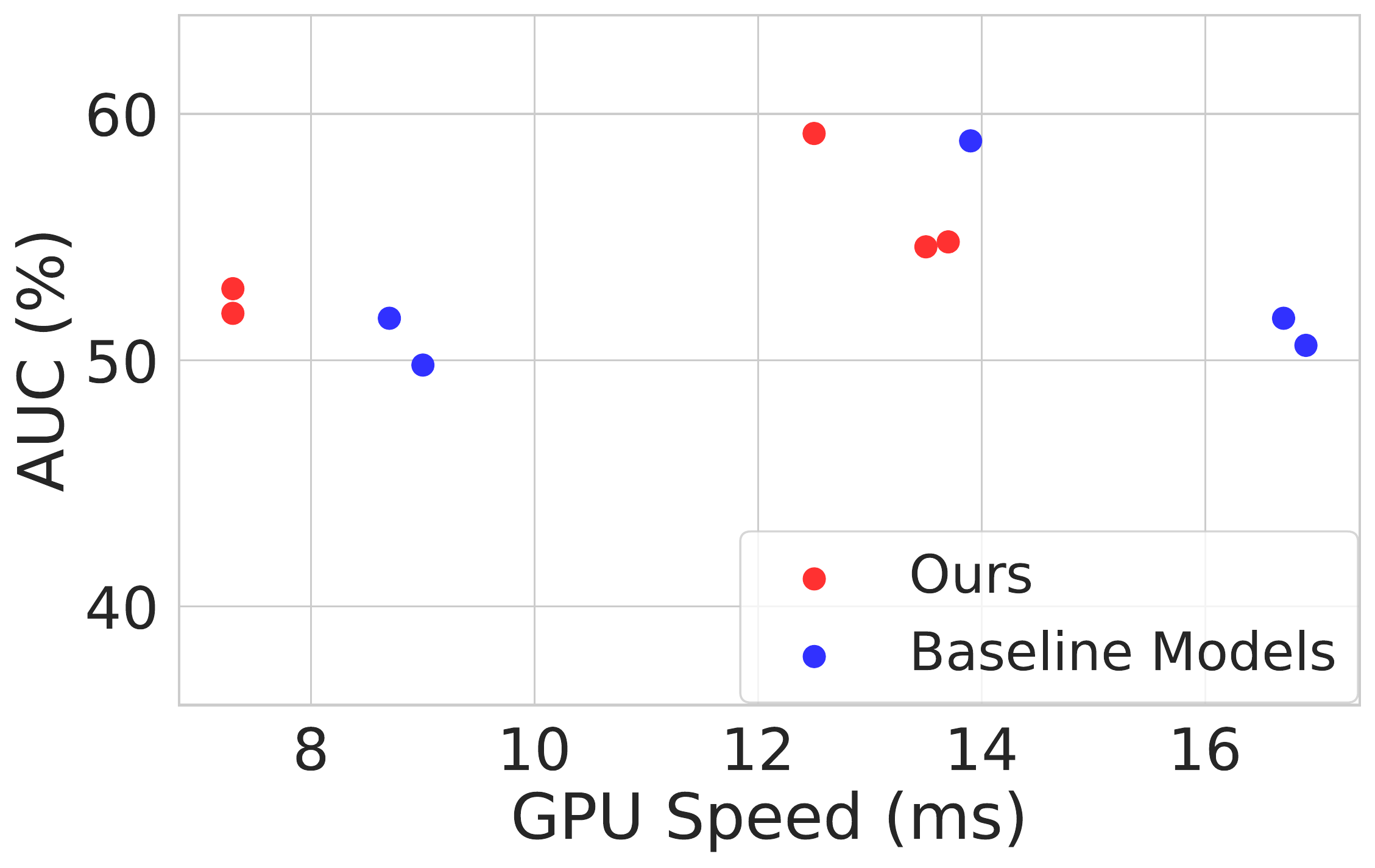}
\caption{\small GPU speed}
\end{subfigure}  \ \
\begin{subfigure}[ht!]{0.3\linewidth}
\includegraphics[page=1, trim = 0mm 0mm 0mm 0mm, clip, width=1.0\linewidth]{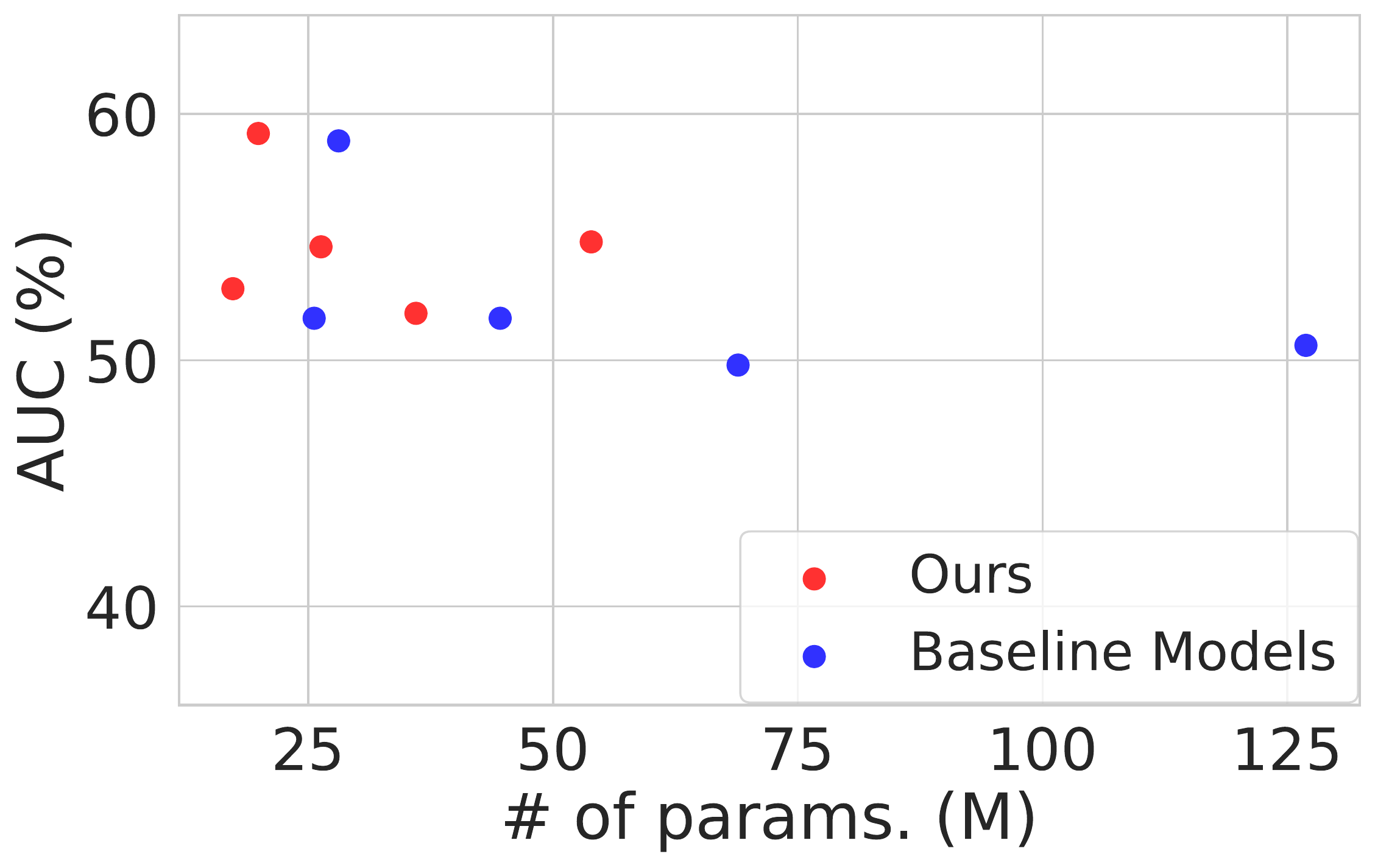}
\caption{\small \# parameters}
\end{subfigure} \ \
\begin{subfigure}[ht!]{0.3\linewidth}
\includegraphics[page=1, trim = 0mm 0mm 0mm 0mm, clip, width=1.0\linewidth]{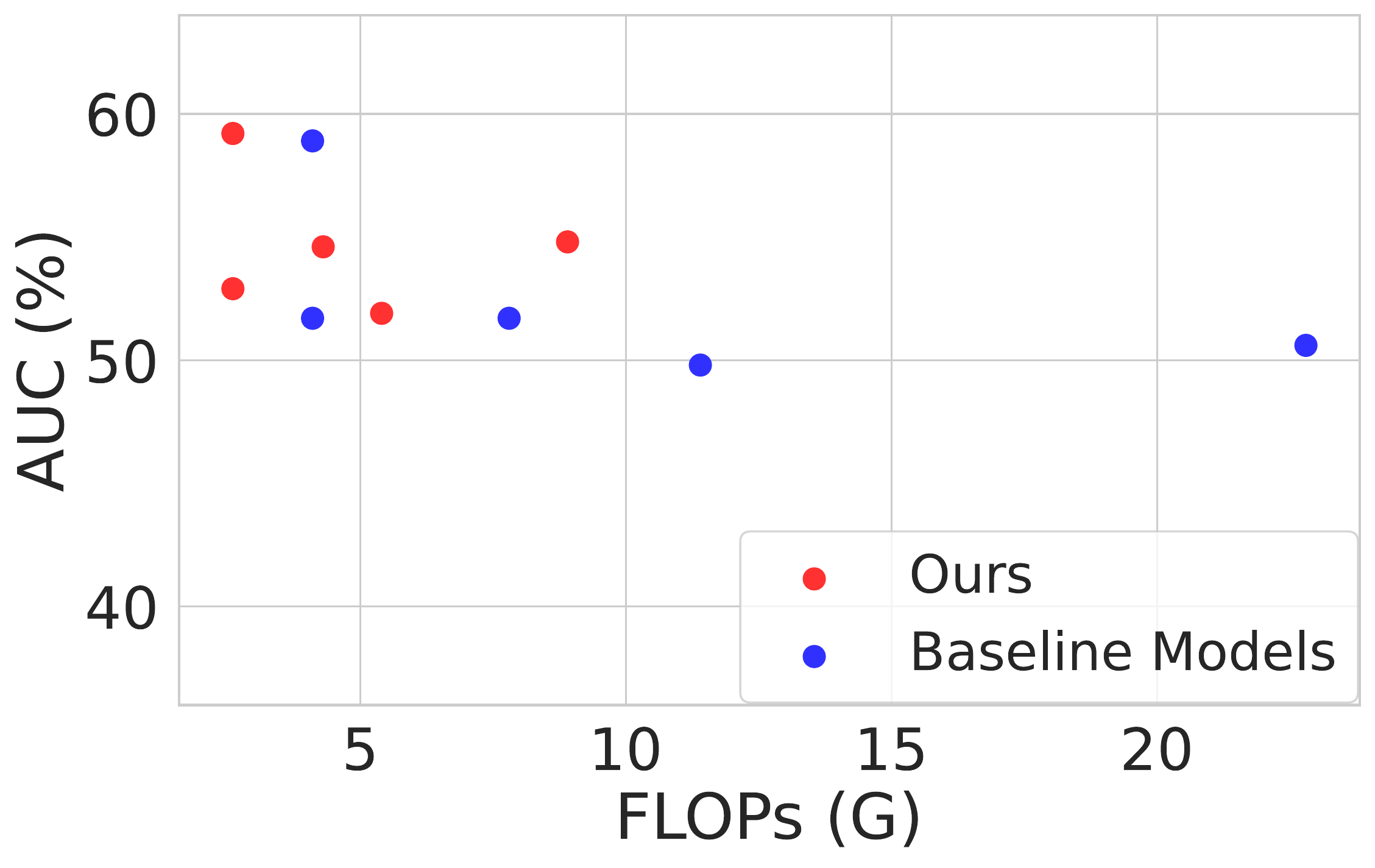}
\caption{\small FLOPs}
\end{subfigure}  
\vspace{-2mm}
\caption{\small \red{{\bf Performance Trade-offs of the models in Table~\ref{table:imagenet_bigger_models}.} We visualize the trade-offs between accuracy/error and (a) speed; (b) \# parameters; (c) FLOPs, respectively. Each row employs 1) Top-1 accuracy on ImageNet; 2) mean Corruption Error (mCE) on ImageNet-C; 3) Area Under the precision-recall Curve (AUC) on ImageNet-O as the performance measure, respectively.}}
\label{fig:fig_table4_visualization}
\vspace{-2mm}
\end{figure*}

Moreover, we visualize the ImageNet results including 1) top-1 accuracy on ImageNet; 2) mean Corruption Error (mCE) on ImageNet-C; 3) Area Under the precision-recall Curve (AUC) on ImageNet-O applying the efficient bottleneck into existing big CNNs, which is shown in Table~\ref{table:imagenet_bigger_models}, in Fig.\ref{fig:fig_table4_visualization}. Fig.\ref{fig:fig_table4_visualization} shows our models achieve large improvements on the entire computational costs but show similar trade-offs of the baseline models in top-1 accuracy. This experiment originally aims to investigate the redundancy of using standard building blocks inside heavy CNN models. However, a simple replacement of the operations in existing models achieves significant efficiency; namely, in terms of the model speeds, the number of parameters, and FLOPs, our models reduce meaningful amounts without much accuracy loss. Furthermore, our models significantly outperform the baseline models with efficiency in mCE and AUC measures. Improving the ImageNet top-1 accuracy while having the advantages of the parameters-free operations will be our future work.


\end{document}